\documentclass[sigconf]{acmart}

\copyrightyear{2023}
\acmYear{2023}
\acmConference[MM '23] {Proceedings of the 31st ACM International Conference on Multimedia}{October 29--November 3, 2023}{Ottawa, ON, Canada.}
\acmBooktitle{Proceedings of the 31st ACM International Conference on Multimedia (MM '23), October 29--November 3, 2023, Ottawa, ON, Canada}
\acmISBN{979-8-4007-0108-5/23/10}
\acmDOI{10.1145/3581783.3611859}
\AtBeginDocument{%
  }

\setcopyright{acmcopyright}
\copyrightyear{2023}
\acmYear{2023}
\acmDOI{10.1145/3581783.3611859}

\acmConference[MM '23] {Proceedings of the 31st ACM International Conference on Multimedia}{October 29--November 3, 2023}{Ottawa, ON, Canada.}
\acmPrice{15.00}
\acmISBN{979-8-4007-0108-5/23/10}

\usepackage{booktabs}
\usepackage{subfig}
\usepackage{hyperref}
\usepackage{multirow}
\usepackage{amsthm}
\usepackage{hyperref}
\usepackage{algorithm}
\usepackage{algorithmic}
\usepackage{url}

\usepackage{hyperref}
\usepackage{bm}

\usepackage{pifont}
\usepackage{algorithm}
\usepackage{algorithmic}
\usepackage{color}
\usepackage{colortbl}

\begin{document}

\title{Bilevel Generative Learning for Low-Light Vision}

\author{Yingchi Liu}
\affiliation{%
	\institution{Dalian University of Technology}\country{}}
\email{yingchi@mail.dlut.edu.cn}

\author{Zhu Liu}
\affiliation{%
	\institution{Dalian University of Technology} \country{}}
\email{liuzhu@mail.dlut.edu.cn}

\author{Long Ma}
\affiliation{%
	\institution{Dalian University of Technology}\country{}}
\email{malone94319@gmail.com}

\author{Jinyuan Liu}
\affiliation{%
	\institution{Dalian University of Technology}\country{}}
\email{atlantis918@hotmail.com}

\author{Xin Fan}
\affiliation{%
	\institution{Dalian University of Technology}\country{}}
\email{xin.fan@dlut.edu.cn}

\author{Zhongxuan Luo}
\affiliation{%
	\institution{Dalian University of Technology}\country{}}
\email{zxluo@dlut.edu.cn}

\author{Risheng Liu}
\authornote{Corresponding author: Risheng Liu}
\affiliation{%
	\institution{Dalian University of Technology\country{}}
	\institution{Peng Cheng Laboratory}\country{}}
\email{rsliu@dlut.edu.cn}
\renewcommand{\shortauthors}{Yingchi Liu et al.} 


\begin{abstract}
   Recently, there has been a growing interest in constructing deep learning schemes for Low-Light Vision (LLV). Existing techniques primarily focus on designing task-specific and data-dependent vision models on the standard RGB domain, which inherently contain latent data associations. In this study, we propose a generic low-light vision solution by introducing a generative block to convert data from the RAW to the RGB domain. This novel approach connects diverse vision problems by explicitly depicting data generation, which is the first in the field. To precisely characterize the latent correspondence between the generative procedure and the vision task, we establish a bilevel model with the parameters of the generative block defined as the upper level and the parameters of the vision task defined as the lower level. We further develop two types of learning strategies targeting different goals, namely low cost and high accuracy, to acquire a new bilevel generative learning paradigm. The generative blocks embrace a strong generalization ability in other low-light vision tasks through the bilevel optimization on enhancement tasks.
  Extensive experimental evaluations on three representative low-light vision tasks, namely enhancement, detection, and segmentation, fully demonstrate the superiority of our proposed approach. The code will be available at \url {https://github.com/Yingchi1998/BGL}.
\end{abstract}


\begin{teaserfigure}
  \includegraphics[width=\textwidth]{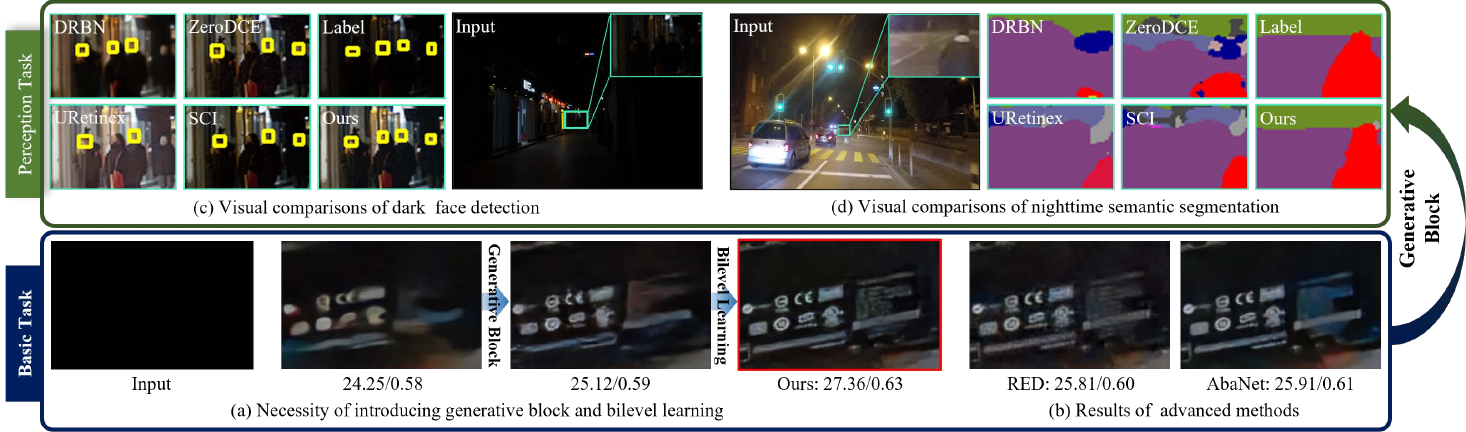}
  \caption{Illustration of the core unified framework for low-light vision tasks. We first demonstrate the necessity of introducing generative block from the RAW domain and further drastically improve the performance based on bilevel learning in (a) based on the basic image enhancement task. Then we compare with the advanced enhanced methods in (b). By plugging the pretrained generative block, significant promotion can be observed from the low-light face detection and nighttime semantic segmentation tasks in (c) and (d).}
  \Description{Enjoying the baseball game from the third-base
  seats. Ichiro Suzuki preparing to bat.}
  \label{fig:teaser}
\end{teaserfigure}


\maketitle

\section{Introduction}
Due to various forms of degradation, such as low illumination, color distortion, low contrast, and imaging noises, low-light vision tasks have become more challenging and have received widespread attention~\cite{huang2022deep, liu2022learning, jin2021bridging, liu2023task, huang2022exposure, zhang2022structure}. The inadequate representation of scenes with low-light observation reduces the perception of the human visual system~\cite{liu2023embracing, huang2022exposure, liu2023holoco, li2022learning} and severely hinders the feature utilization and performance of various scene perception tasks. Therefore, improving visual quality and investigating effective features for downstream scene understanding are significant in real-world applications, \textit{e.g.,} autonomous sensing~\cite{ignatious2022overview} and robotic operations~\cite{macenski2022robot}.

In recent years, significant advances have been made in learning-based schemes for low-light vision tasks. These approaches aim to highlight sufficient scene information under severe illumination conditions. For visual enhancement, Retinex-based decomposition principles that estimate illumination and recover normal-light reflection are widely utilized~\cite{liu2021retinex,zhang2019kindling, liu2022learning}. For scene perception~\cite{liu2022attention, liu2020bilevel, liu2021learning, liu2019knowledge}, approaches such as bi-directional high-low level adaptation~\cite{wang2022unsupervised} and recurrent exposure generation~\cite{liang2021recurrent} have been proposed. These schemes achieve excellent through task-specialized schematic design and data-driven optimization under  RGB domain.

However, the heuristic and independent design of current low-light vision tasks complicates unified modeling, making it challenging to realize diverse purposes of low-light vision under a versatile framework. Low-light images based on RGB formation  inevitably suffer from strong noise and color distortion, which limits their performance at its core.

Owing to the higher bit depth to capture features, many learning-based methods aim to investigate the low-light scene information from RAW domain for perception tasks~\cite{xu2022snr,ma2023bilevel, wang2023interactively, liu2023bilevel}. Despite the presence of large noise from hardware device, the original RAW formation also provides the sufficient scene information.
Theses methods  design the specialized solutions for specific 
task goals, which  cannot flexible generalize into other low-light vision tasks without modular-style design. Different from existing methods, we endow the adaptive data generation from RAW  to establish the connection of diverse  vision tasks, which is key to improving the performances for diverse perception tasks.

To partially mitigate the above issues, we propose  Bilevel Generative Learning ( BGL ) to uniformly solve various low-light vision tasks under  the RAW domain.  Within this formulation, we introduce the RAW-based generation  as the goal and the low-light vision is cast as a constraint. This bilevel generative learning can naturally provide a new perspective to describe the implicit relationship between the  image  generated from  RAW data and performance of low-light vision.  In addition, we construct two optimization strategies to provide
efficient solutions including truncated and implicit bilevel generative learning for real-world practices.   These training strategies are both with high efficiency compared with mainstream bilevel solutions and avoid any needs to fine-tune with downstream low-light vision networks. Note that the generative block trained by bilevel learning is versatile enough to  promote the performances of other low-light vision tasks.
As shown in Fig.~\ref{fig:teaser}, our scheme  can realize the significant improvement for diverse tasks (\textit{i.e.,} image enhancement, detection and segmentation). The main contributions of this work  can be summarized as:

\begin{itemize}
	\item To the best of our knowledge, we are the first to provide a versatile framework to  address low-light vision perception tasks under the support of generative procedure  from RAW space. It not only endows task guidance for generating data, but also boosts the performance of vision tasks.
	
	\item In order to exploit the latent connection between data generation and task-specific parameters, we  provide a bilevel generative learning paradigm to effectively model their inner  correspondence and acquire the data generation  with strong generalized capabilities for diverse vision tasks.
	
	\item To satisfy different demands in real-world practice, we develop two strategies for BGL, including a truncated BGL for alleviating computation costs and an implicit BGL for  high accuracy, which provide effective and precise responses for the data generation learning.
	
	\item Comprehensive experiments have elaborated the remarkable  improvements for various low-light visual perception tasks, achieving promising results against advanced methods. We also make extensive analysis in terms of our proposed method to verify our effectiveness.
\end{itemize}
\section{Related Works}
In this part, we first briefly review the current development of low-light vision based on RGB and RAW domains.

\textit{RGB-based LLV.}
Recently, existing low-light enhancement methods designed heuristic network structure based on data-driven for specific low-light scenes and yielded better results for image restoration. Among them, Retinex-theory has been the mainstream methodology for low-light enhancement. For instance,   image decomposition scheme is proposed to  simultaneously optimize the illumination and reflections. The work~\cite{zhang2019kindling} introduces the decomposition mechanisms to estimate the illumination and reflectance at feature domain.
As for scene perception, target-specific (e.g., face) low-light object detection is a typical task. ~\cite{wang2022unsupervised} designed a  dual-direction adaptation framework with a interactive bidirectional high-low level adaptation, which can extract effective information for high level from the low level module.  In summary,  learning-based schemes constructed independent architectures and training strategies to achieve promising results with high efficiency.
One major challenge is the difficulty to formulate these low-light vision tasks with a principled unified framework, due to the independent design of these schemes.  In our scheme, we aims to seek data  connection of these tasks to design a versatile framework with a suitable training strategy.

\textit{RAW-based LLV.}
Since RAW images contain comprehensive information with untreated scene representation. 
Several works have pioneered to simulate the RAW-to-RGB  mapping to address diverse vision tasks. Specifically, as for low-light vision enhancement, SIDfirstly proposed a landmark cornerstone with sufficient  RAW-RGB pairs under low-light scenes with different exposure time. Moreover, DID replaces the U-Net structure with the cascaded residual blocks, in order to preserve the texture details with the same resolution. Lastly, there are also some schemes to construct low-weight architectures, such as LLPackNet, RED and SGN, leveraging the low-resolution operating, multi-scale processing and self-guided
mechanisms respectively. 
 Based on the unprocessing technique, Chen~\textit{et.al} propose the low-light RAW-input instance segmentation algorithm~\cite{chen2023instance} based on a low-light RAW synthetic pipeline from RGB domain. Inspired by this work, we can further introduce the unprocessing to mitigate the shortage of  RAW datasets under low-light environment.
These learnable transformation schemes rely on data-driven, which achieve high superiority of performance.

\section{Proposed Method}

In this part, we first present the core motivation, introducing the data generation block from the RAW data to integrate the solution of diverse low-light vision tasks. Then we formulate the data generative block and downstream vision tasks based on the bilevel generative modeling.

\subsection{Low-light Vision with Generative Block}
The mainstream methodology to address low-light vision tasks is to design independent architectures for specific training data. In this way, due to the distinct principles to design architectures, it is difficult to introduce a unified scheme to take diverse low-light vision tasks under  comprehensive consideration.
From the intuitive viewpoint, we can observe that these learning-based methods actually share the similar data dependence, i.e., the low-light observations. Thus, how to investigate  connections among low-light tasks from  data generation perspective is a potential direction.

In this manuscript, we introduce a generative block from RAW space for vision task. It is noticeable that, in essence, generative block is task-agnostic, which only targets to establish the mapping between RAW and RGB images. In a word, data generation does not have the strong correlation with   specific  vision tasks. In this way, for diverse low-light vision tasks, we can introduce the learnable data generation to construct a common  dependence and further  formulate the unified framework to bridge these vision tasks. 
Furthermore, the original information from RAW domain is not corrupted. We can extract efficient features from RAW data to further improve the performance of vision tasks. 

In detail, we denote $\mathbf{z}$, $\mathbf{x}$ and $\mathbf{y}$ are  inputs from RAW, RGB space and task-specific outputs, respectively. The complete framework can be formulated as $\mathcal{N}:=\mathcal{F}\circ\mathcal{G}$, where $\mathcal{F}$ and $\mathcal{G}$ are the network of data generation and low-light vision tasks. The whole formation can be written as:
\begin{eqnarray}
	\left\{
	\begin{aligned}
		\mathbf{x} & =  \mathcal{F}(\mathbf{z};\bm{\omega}),\\
		\mathbf{y} & =  \mathcal{G}(\mathbf{x};\bm{\theta}), \\
	\end{aligned}
	\right.
\end{eqnarray}
where $\bm{\omega}$ and $\bm{\theta}$ are the corresponding parameters. 
These details about the network  implementation  are reported at Section.3.5.

In this part, we construct a cascade framework to formulate their forward data flow.  The direct heuristic optimization strategy is to perform the end-to-end learning. However, we argue that the straightforward optimization cannot model the complicated relationship between  data generation and vision tasks. The straightforward training only consider their goal independently, without the formulation of their coupled influences.
Thus, how to depict their latent correspondence becomes another challenging issue.

\subsection{Modeling with Bilevel Optimization}
In this part, we demonstrate how the bilevel modeling relates to the formulation between generative learning and  parameter learning of low-light vision. Bilevel modeling is a technique to involve hierarchical optimization tasks, where the lower-level task is nested in the upper-level optimization. Analogy with effective practices on generative adversarial networks~\cite{goodfellow2020generative,liu2022revisiting} and adversarial attacks~\cite{zhang2022revisiting}, we can naturally discover the core relationship between data generation and  task-specific parameter learning.

Intuitively, the data generation from RAW domain can be considered as one ``Generator'',
to provide inputs satisfied for downstream low-light vision tasks. On the other hand, the role of network for low-light vision can be viewed as the ``Discriminator'', to provide the task guidance and feedback in order to adjust the learning procedure of generative block. Importantly, it supplies a more powerful understanding compared with  naive end-to-end learning and new perspective to model the correspondence between training data and network parameters.

To make the above intuition clear, the bilevel modeling between data generation and parameter learning can be formulated as:
\begin{align}
	&\min\limits_{\bm{\omega}}\mathcal{L}_{\mathcal{F}}(\mathcal{F}(\mathbf{z};\bm{\omega})\circ\mathcal{G}(\mathbf{x};\bm{\theta}^{*})),\label{eq:main}\\
	&\mbox{s.t.}\quad\bm{\theta}^{*} = \arg\min\limits_{\bm{\theta}} \mathcal{L}_{\mathcal{G}}(\mathcal{G}(\mathbf{x};\bm{\theta}(\bm{\omega}))),
	\label{eq:constraint}
\end{align}
where $\mathbf{x} = \mathcal{F}(\mathbf{z};\bm{\omega}^{*})$. We define the parameters of generative block $\bm{\omega}$ as the related hyper-parameters to learn the scene-agnostic feature representation.
$\mathcal{L}_{\mathcal{F}}$ denotes the task-specific loss on the validation dataset to optimize the generative block. $\mathcal{L}_{\mathcal{G}}$  represents the optimization objective for the visual task module. 
We introduce the vision task as constraints to utilize more task-related responses.  Furthermore, this formulation is flexible enough to replace effective learnable modules for $\mathcal{F}$ and $\mathcal{G}$ with strong versatility. Different from other works that focus on network design, we put more attention on characterizing the correspondence between informative data generation and low-light vision tasks.

\subsection{Bilevel Generative Learning}

In literature, amounts of computation strategies  are proposed to address these  constrained modeling. However, due to the complicated nested formulation of Eq.~\eqref{eq:main} and Eq.~\eqref{eq:constraint},  parameters of generative block $\bm{\omega}$ is actually embedded into the optimization of vision tasks. The solving procedure of current bilevel optimization always  needs massive computation resource, limiting their applications on the simple tasks. Specifically, the bottleneck of optimization is to introduce the task-related response based on $\mathcal{G}$, when computing the gradient of $\mathcal{F}$, which can be formulated as  
\begin{equation}\label{eq:bilevel}
	\mathbf{G}_\mathcal{F}:=\nabla_{\bm{\omega}}\mathcal{L}_\mathcal{F}+\underbrace{\nabla_{\bm{\theta}}\mathcal{L}_\mathcal{F}\nabla_{\bm{\omega}}\bm{\theta}(\bm{\omega})}_{\text{Second-order gradient}}.
\end{equation}
The first term represents the direct gradient for generative block. The second term reveals the task-specific responses of $\mathcal{F}$ for downstream vision tasks. Thus, this provides us with a more intuitive view to understand the complicated connection between $\bm{\omega}$ and $\bm{\theta}$. 

Existing methods cannot satisfy the demand of low cost or high accuracy to solve our model. In this part, we propose two effective  strategies to approximate the second term of Eq.~\eqref{eq:bilevel}, which provide efficient solutions for different demands (\textit{i.e.}, low computation cost and high performance). The practical solutions for different demands are summarized at Alg.~\ref{alg:framework} and Alg.~\ref{alg:framework2}.
\begin{algorithm}[htb] 
	\caption{ Truncated Bilevel Generative Learning (TBGL).}\label{alg:framework}
	\begin{algorithmic}[1] 
		\REQUIRE Inputs from RAW domain $\mathbf{z}$, loss function $\mathcal{L}_\mathcal{F}$ and $\mathcal{L}_\mathcal{G}$ and other necessary hyper-parameters.
		
		\STATE RGB output $\mathbf{x}$ and $\mathbf{y}$.
		
		\STATE  Pretrain the $\mathcal{F}(\mathbf{z};\bm{\omega})\circ\mathcal{G}(\mathbf{x};\bm{\theta})$ for initializing $\bm{\theta}$ and $\bm{\omega}$. 
		\STATE Joint optimizing $\mathcal{F}$ and $\mathcal{G}$ utilizing $\mathcal{L}_\mathcal{G}$.
		
		\STATE \% Solving the bilevel modeling.
		\WHILE {not converged}
		\STATE \% Updating the network of vision task $\mathcal{G}$ (with one step).
		\STATE $\bm{\theta}\leftarrow\bm{\theta}-\nabla_{\bm{\theta}} \mathcal{L}_{\mathcal{G}}(\mathcal{G}(\mathbf{x};\bm{\theta}(\bm{\omega})))$.
		\STATE  \% Updating the network of generative learning $\mathcal{F}$.
		\STATE Calculating gradient $\mathbf{G}_\mathcal{F}$ with  TBGL (Eq.~\eqref{eq:fda}) .
		
		\STATE $\bm{\omega}\leftarrow\bm{\omega}-\mathbf{G}_\mathcal{F}$.
		\ENDWHILE
		\RETURN $\bm{\omega}^{*}$ and $\bm{\theta}^{*}$.
	\end{algorithmic}
\end{algorithm}
\subsubsection{Truncated BGL for Lower Cost}
We first introduce a truncated BGL to reduce the computation cost.
In detail, this strategy is based on the one-step forward model~\cite{liu2022learning}, which is truncated with recurrent updates of $\mathcal{G}$ ($\mathbf{k}=1$ in the Alg.~1). Denoted that $\bm{\theta}^{'} = \bm{\theta}-\eta\nabla_{\bm{\theta}}\mathcal{L}_\mathcal{G}$, the gradient of $\mathcal{F}$ can be written as $\nabla_{\bm{\omega}}\mathcal{L}_\mathcal{F}-\eta\nabla_{\bm{\omega}\bm{\theta}}^{2}\mathcal{L}_\mathcal{G}\nabla_{\bm{\theta}^{'}}\mathcal{L}_\mathcal{F}$. First-order based approximation is widely used, which can significantly circumvent the huge memory utilization for expensive hessian matrix and improve computation efficiency by approximating the second-order gradient. By utilizing the finite difference approximation, the second term can be computed as:
\begin{equation}\label{eq:fda}
	\nabla_{\bm{\omega},\bm{\theta}}^{2}\mathcal{L}_\mathcal{G}\nabla_{\bm{\theta}^{'}}\mathcal{L}_\mathcal{F} \approx  \frac{\nabla_{\omega}\mathcal{L}_\mathcal{G}(\bm{\omega};\bm{\theta}^{+})-\nabla_{\omega}\mathcal{L}_\mathcal{G}(\bm{\omega};\bm{\theta}^{-})}{2\delta},
\end{equation}
where $\delta$ is a constant and $\bm{\theta}^{\pm} = \bm{\theta}\pm\delta \nabla_{\bm{\theta}^{'}}\mathcal{L}_\mathcal{F}$. In this way, the solution of Eq.~\eqref{eq:bilevel} actually contains three first-order gradient computation for $\bm{\omega}$. Thus the simplification only perform one-step updation, which drastically reduce the computation cost. However, though highly efficiency obtained, this strategy is restricted to the one-step back-propagation for sufficiently learning the conversion $\mathcal{F}$. In other words,  Truncated BGL (TBGL) cannot guarantee the performance stably, without  reasonable responses from well-trained low-light vision tasks. Thus, we introduce another strategy to provide higher performance to solve the Eq.~\eqref{eq:bilevel}.
In the next part, we will explore the concrete computation learning strategy to address this bilevel optimization objective.

\subsubsection{Implicit BGL for Higher Accuracy}
Introducing the Gaussian-Newton method to approximate the second-order matrix for addressing Generative Adversarial Networks
(GANs) and continuous learning is introduced in~\cite{liu2022revisiting}. This strategy is based on the implicit gradient approximation, where the implicit function theory is employed to fastly estimate the coupled second-order term (denoted as $\mathcal{G}_\mathtt{C}$). Based on the implicit function assumption, $\nabla_{\bm{\theta}}\mathcal{L}_\mathcal{G}=0$, we can have
$\nabla_{\bm{\omega}}\bm{\theta}(\bm{\omega}) = -\nabla_{\bm{\theta},\bm{\theta} }^{2}\mathcal{L}_\mathcal{G}^{-1}  \nabla_{\bm{\omega},\bm{\theta}}^{2}\mathcal{L}_\mathcal{G}$. 
Assuming that $\nabla_{\bm{\omega},\bm{\theta}}^{2}\mathcal{L}_\mathcal{G}$ is transposeable, the coupling gradient $\mathcal{G}_\mathtt{C}$ can be writen as follows:
\begin{equation}\label{eq:out-product_0}
	\mathcal{G}_\mathtt{C} = - \nabla_{\bm{\omega},\bm{\theta} }^{2}\mathcal{L}_\mathcal{G}(\bm{\omega},\bm{\theta})^{\top} \nabla_{\bm{\theta},\bm{\theta}}^{2}\mathcal{L}_\mathcal{G}(\bm{\omega};\bm{\theta})^{-1}\nabla_{\bm{\theta}}\mathcal{L}_\mathcal{F}.
\end{equation}
This formulation actually have two second-order derivatives (Hessian and its inverse). Utilizing the inspiration of Gauss-Newton method,  which utilize the outer-product-based approximations. This term can be computed as:
\begin{equation}\label{eq:outer-product_1}
	\nabla_{\bm{\theta},\bm{\theta}}^{2}\mathcal{L}_\mathcal{G}(\bm{\omega};\bm{\theta}) \approx \nabla_{\bm{\theta}}\mathcal{L}_\mathcal{G}(\bm{\omega};\bm{\theta})\nabla_{\bm{\theta}}\mathcal{L}_\mathcal{G}(\bm{\omega};\bm{\theta})^{\top}.
\end{equation}
\begin{equation}\label{eq:outer-product_2}
	\nabla_{\bm{\omega},\bm{\theta}}^{2}\mathcal{L}_\mathcal{G}(\bm{\omega};\bm{\theta}) \approx \nabla_{\bm{\theta}}\mathcal{L}_\mathcal{G}(\bm{\omega};\bm{\theta})\nabla_{\bm{\omega}}\mathcal{L}_\mathcal{G}(\bm{\omega};\bm{\theta})^{\top}.
\end{equation}
This way of gradient approximation converts the two second-order Hessian matrix into  a simple product operation of  the first-order derivation, which can  greatly improve the computation efficiency. The coupling gradient $\mathcal{G}_\mathtt{C}$ can be obtained as follows:

\begin{equation}\label{eq:ida}
	\mathcal{G}_\mathtt{C} \approx - \frac{\nabla_{\bm{\theta}}\mathcal{L}_\mathcal{F}(\bm{\omega};\bm{\theta})^{\top} \nabla_{\bm{\theta}}\mathcal{L}_\mathcal{G}{\bm{\omega}}(\bm{\omega};\bm{\theta})}{\nabla_{\bm{\theta}}\mathcal{L}_\mathcal{G}(\bm{\omega};\bm{\theta})^{\top} \nabla_{\bm{\theta}}\mathcal{L}_\mathcal{G}(\bm{\omega};\bm{\theta})}
	\nabla_{\bm{\omega}}\mathcal{L}_\mathcal{G}(\bm{\omega};\bm{\theta}).
\end{equation}
Different from the above truncated BGL, which only updates one step of the lower-level objective, we set more training steps to update $\mathcal{F}$ for the convergence and provide more accurate responses for the generative learning.
In this way, we can avoid the limitation of one-step updating. Similarly, we only need to compute several first-order gradients rather than  Hessian products. Network of LLV can be trained to approximate the optimal, which can offer more precise response. Thus, IBGL can guarantee the high performance of the solution. But this strategy cannot maintain the efficiency and needs the amounts of updating for $\mathcal{G}$ compared with TBGL.
\begin{algorithm}[htb] 
	\caption{Implicit Bilevel Generative Learning (IBGL).}\label{alg:framework2}
	\begin{algorithmic}[1] 
		\REQUIRE Same as the Algorithm 1.
		
		\STATE \% Warm start phase. 
		\STATE Joint optimizing $\mathcal{F}$ and $\mathcal{G}$ utilizing $\mathcal{L}_\mathcal{G}$.
		\WHILE {not converged}
		\STATE \% Updating the network of vision task $\mathcal{G}$ with $\mathbf{k}$ steps.
		\FOR{$i=1$ to $\mathbf{k}$ steps}
		\STATE 	$\bm{\theta}\leftarrow\bm{\theta}-\nabla_{\bm{\theta}} \mathcal{L}_{\mathcal{G}}(\mathcal{G}(\mathbf{x};\bm{\theta}(\bm{\omega})))$;
		\ENDFOR
		
		\STATE  \% Updating the network of generative learning $\mathcal{F}$.
		\STATE Calculating gradient $\mathbf{G}_\mathcal{F}$ with IBGL(Eq.~\eqref{eq:ida}).
		
		\STATE $\bm{\omega}\leftarrow\bm{\omega}-\mathbf{G}_\mathcal{F}$.
		\ENDWHILE
		\RETURN $\bm{\omega}^{*}$ and $\bm{\theta}^{*}$.
	\end{algorithmic}
\end{algorithm}

\subsection{Discussion}
It is worth noting that this is the first work to model the RAW-based data generation with vision tasks from a generic bilevel optimization perspective. Instead of introducing a unique design or modules in the RAW processing, we are investigating the data generation from RAW space as a hyper-parameter optimization and considering the constraints from low-light vision to assist the learning of data generation. 
\begin{table*}[htb]
	\centering
	
	\caption{  Quantitative results of low-light enhancement on the SID dataset.}~\label{tab:SID_tab}
	\renewcommand{\arraystretch}{1.1}
	
	\setlength{\tabcolsep}{2.598mm}{
		\begin{tabular}{|c|c|c|c|c|c|c|c|c|}
			\hline
			\multirow{2}{*}{Method} & \multicolumn{1}{c|}{SID}                                                                                                                         & \multicolumn{1}{c|}{
				SGN }                                                                                                                                               & \multicolumn{1}{c|}{DID}&\multicolumn{1}{c|}{LLpack}&\multicolumn{1}{c|}{RED}&\multicolumn{1}{c|}{RawFormer} &\multicolumn{1}{c|}{AbaNet}&\multicolumn{1}{c|}{Ours}    \\                                                                                                                 
			& \multicolumn{1}{c|}{\textit{CVPR'2018}} & \multicolumn{1}{c|}{\textit{ICCV'2019}} & \multicolumn{1}{c|}{\textit{ICME'2019}} & \multicolumn{1}{c|}{\textit{BMVC'2020} } & \multicolumn{1}{c|}{\textit{CVPR'2021} } & \multicolumn{1}{c|}{\textit{SPL'2022} } & \multicolumn{1}{c|}{\textit{CVPR'2022}} & \multicolumn{1}{c|}{--}    \\ \hline
			
			PSNR$\uparrow$	& \multicolumn{1}{c|}{28.262}    & \multicolumn{1}{c|}{26.883}       & \multicolumn{1}{c|}{28.814}     & \multicolumn{1}{c|}{27.944}     & \multicolumn{1}{c|}{28.811}     & \multicolumn{1}{c|}{28.483}      
			& \multicolumn{1}{c|}{\textcolor{blue}{\textbf{29.011}}}    & \multicolumn{1}{c|}{\textcolor{red}{\textbf{29.122}}}                                       
			\\ \hline
			
			SSIM$\uparrow$	& \multicolumn{1}{c|}{0.763}    & \multicolumn{1}{c|}{0.753}       & \multicolumn{1}{c|}{0.767}     & \multicolumn{1}{c|}{0.753}     & \multicolumn{1}{c|}{0.744}     & \multicolumn{1}{c|}{0.765}      
			& \multicolumn{1}{c|}{\textcolor{blue}{\textbf{0.768} }}   & \multicolumn{1}{c|}{\textcolor{red}{\textbf{0.769}}}                          
			\\ \hline
			
			LPIPS$\downarrow$& \multicolumn{1}{c|}{\textcolor{blue}{\textbf{0.439}}}    & \multicolumn{1}{c|}{0.469}       & \multicolumn{1}{c|}{0.440}     & \multicolumn{1}{c|}{0.768}     & \multicolumn{1}{c|}{0.497}     &     \multicolumn{1}{c|}{0.441}  
			& \multicolumn{1}{c|}{0.444}    & \multicolumn{1}{c|}{\textcolor{red}{\textbf{0.432}}}                                    
			\\ \hline
			VIF$\uparrow$	& \multicolumn{1}{c|}{0.917}    & \multicolumn{1}{c|}{0.927}       & \multicolumn{1}{c|}{0.909}     & \multicolumn{1}{c|}{0.388}     & \multicolumn{1}{c|}{0.890 }    &\multicolumn{1}{c|}{\textcolor{blue}{\textbf{0.951}}}& \textcolor{red}{\textbf{0.960}}&\multicolumn{1}{c|}{0.923} 
			\\ \hline
			FSIM$\uparrow$	& \multicolumn{1}{c|}{0.930}    & \multicolumn{1}{c|}{0.923}       & \multicolumn{1}{c|}{0.930}   & \multicolumn{1}{c|}{0.680}     & \multicolumn{1}{c|}{0.921}    &\multicolumn{1}{c|}{0.844}&\textcolor{red}{\textbf{0.937 }}&\multicolumn{1}{c|}{\textcolor{blue}{\textbf{0.931}}}
			\\ \hline
			BRISQUE$\downarrow$	& \multicolumn{1}{c|}{28.726}    & \multicolumn{1}{c|}{29.429}       & \multicolumn{1}{c|}{29.427}    & \multicolumn{1}{c|}{35.544}     & \multicolumn{1}{c|}{27.653 }    &\multicolumn{1}{c|}{27.706}&\textcolor{blue}{\textbf{27.702}}& \multicolumn{1}{c|}{\textcolor{red}{\textbf{27.603}}}
			\\ \hline
		\end{tabular}	
	}
\end{table*}

\begin{figure*}[t]
	
	\centering
	\begin{tabular}{c@{\extracolsep{0.25em}}c@{\extracolsep{0.25em}}c@{\extracolsep{0.25em}}c@{\extracolsep{0.25em}}c@{\extracolsep{0.25em}}c@{\extracolsep{0.25em}}c} 
		\includegraphics[width=0.133\textwidth]{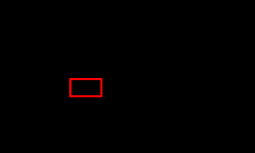}&
		\includegraphics[width=0.133\textwidth]{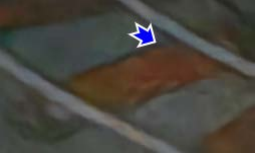}&
		\includegraphics[width=0.133\textwidth]{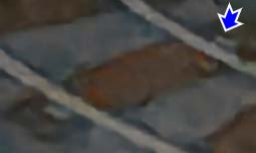}&
		\includegraphics[width=0.133\textwidth]{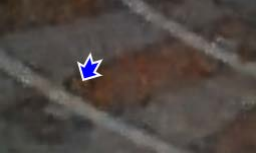}&
		\includegraphics[width=0.133\textwidth]{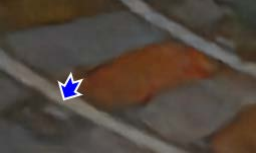}&
		\includegraphics[width=0.133\textwidth]{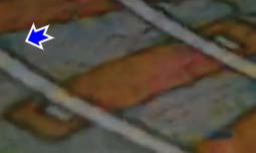}&
		\includegraphics[width=0.133\textwidth]{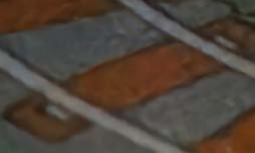}\\
		
		\includegraphics[width=0.133\textwidth]{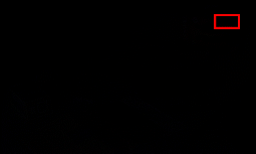}&
		\includegraphics[width=0.133\textwidth]{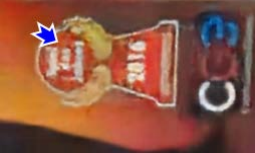}&
		\includegraphics[width=0.133\textwidth]{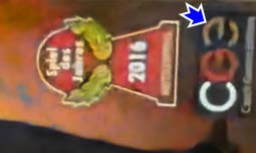}&
		\includegraphics[width=0.133\textwidth]{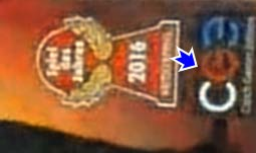}&
		\includegraphics[width=0.133\textwidth]{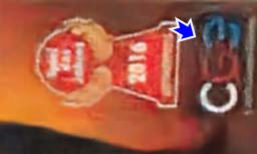}&
		\includegraphics[width=0.133\textwidth]{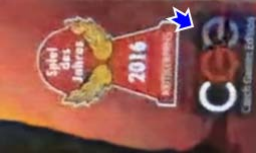}&
		\includegraphics[width=0.133\textwidth]{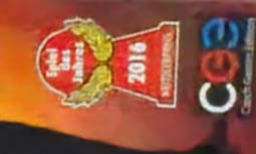}\\
		
		\includegraphics[width=0.133\textwidth]{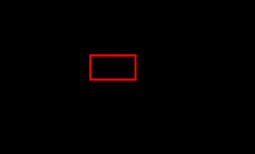}&
		\includegraphics[width=0.133\textwidth]{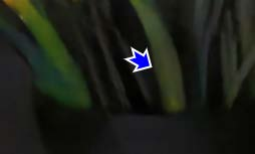}&
		\includegraphics[width=0.133\textwidth]{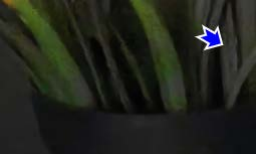}&
		\includegraphics[width=0.133\textwidth]{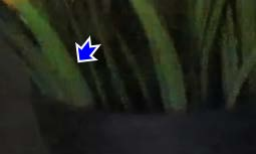}&
		\includegraphics[width=0.133\textwidth]{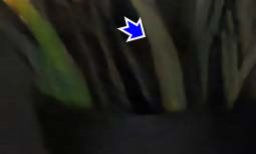}&
		\includegraphics[width=0.133\textwidth]{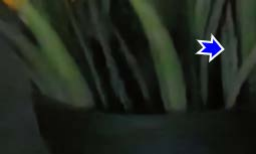}&
		\includegraphics[width=0.133\textwidth]{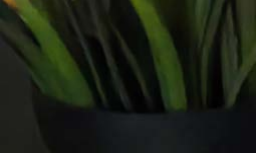}\\
		Input & SID & SGN& RED& RawFormer& AbaNet&  Ours\\
	\end{tabular}
	
	\caption{Visual comparison  against state-of-the-art low-light image enhancement approaches on the SID dataset. }
	\label{fig:SID_fig}
	
\end{figure*}

\begin{figure*}[t]
	\centering
	\begin{tabular}{c@{\extracolsep{0.25em}}c@{\extracolsep{0.25em}}c@{\extracolsep{0.25em}}c@{\extracolsep{0.25em}}c@{\extracolsep{0.25em}}c@{\extracolsep{0.25em}}c@{\extracolsep{0.25em}}c@{\extracolsep{0.25em}}c@{\extracolsep{0.25em}}c} 

		\includegraphics[width=0.090\textwidth]{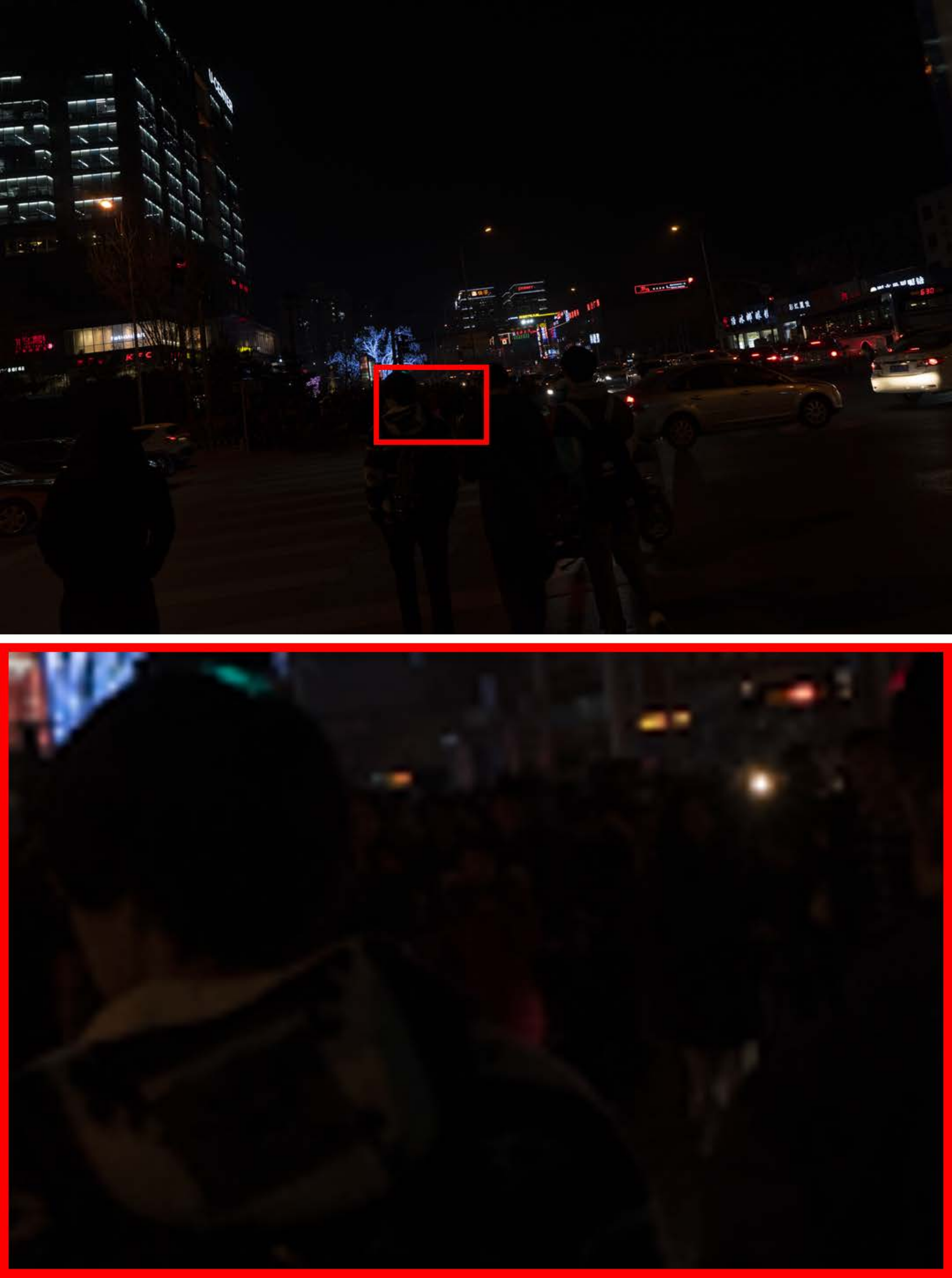}&
		\includegraphics[width=0.090\textwidth]{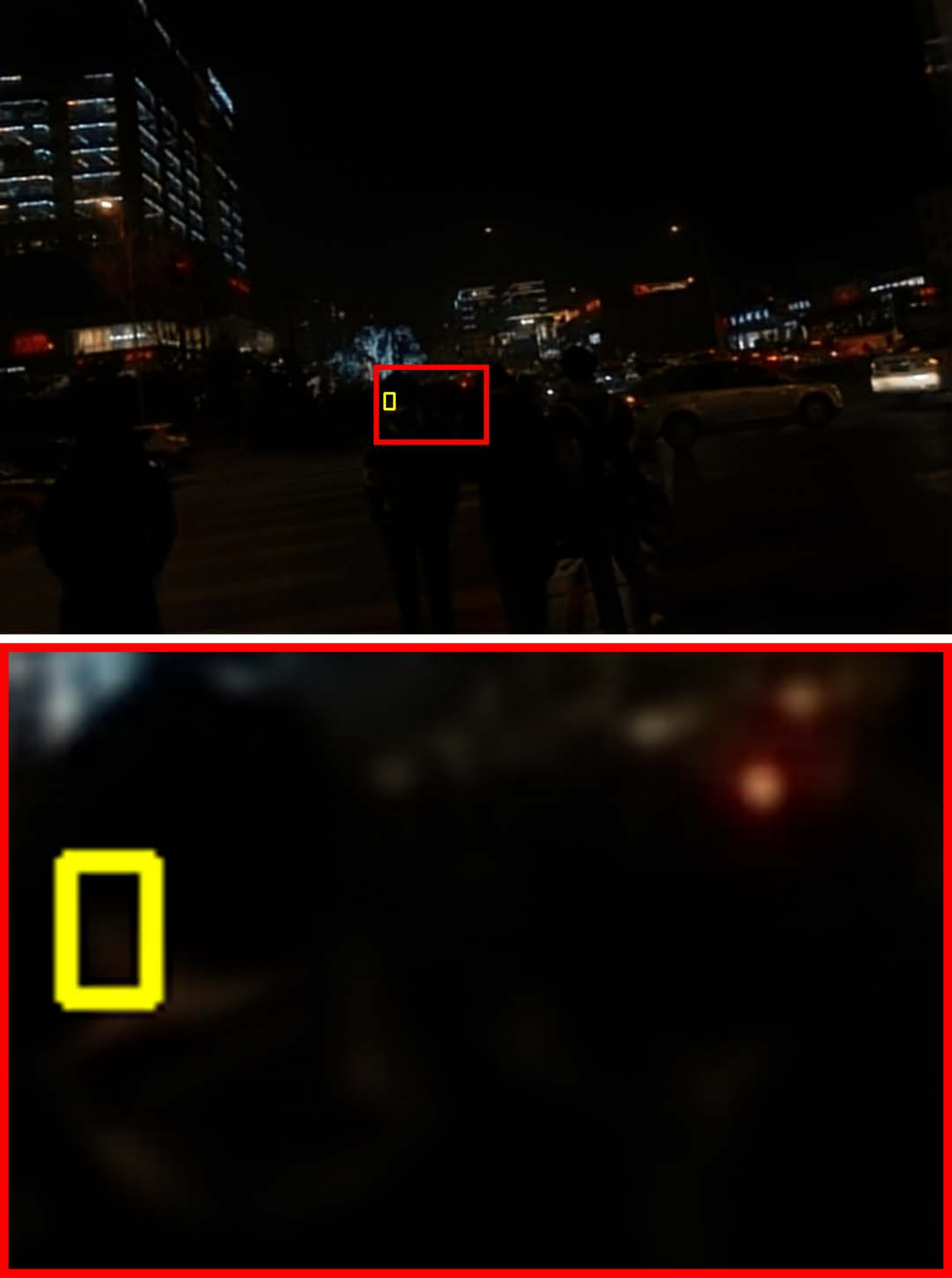}&
		\includegraphics[width=0.090\textwidth]{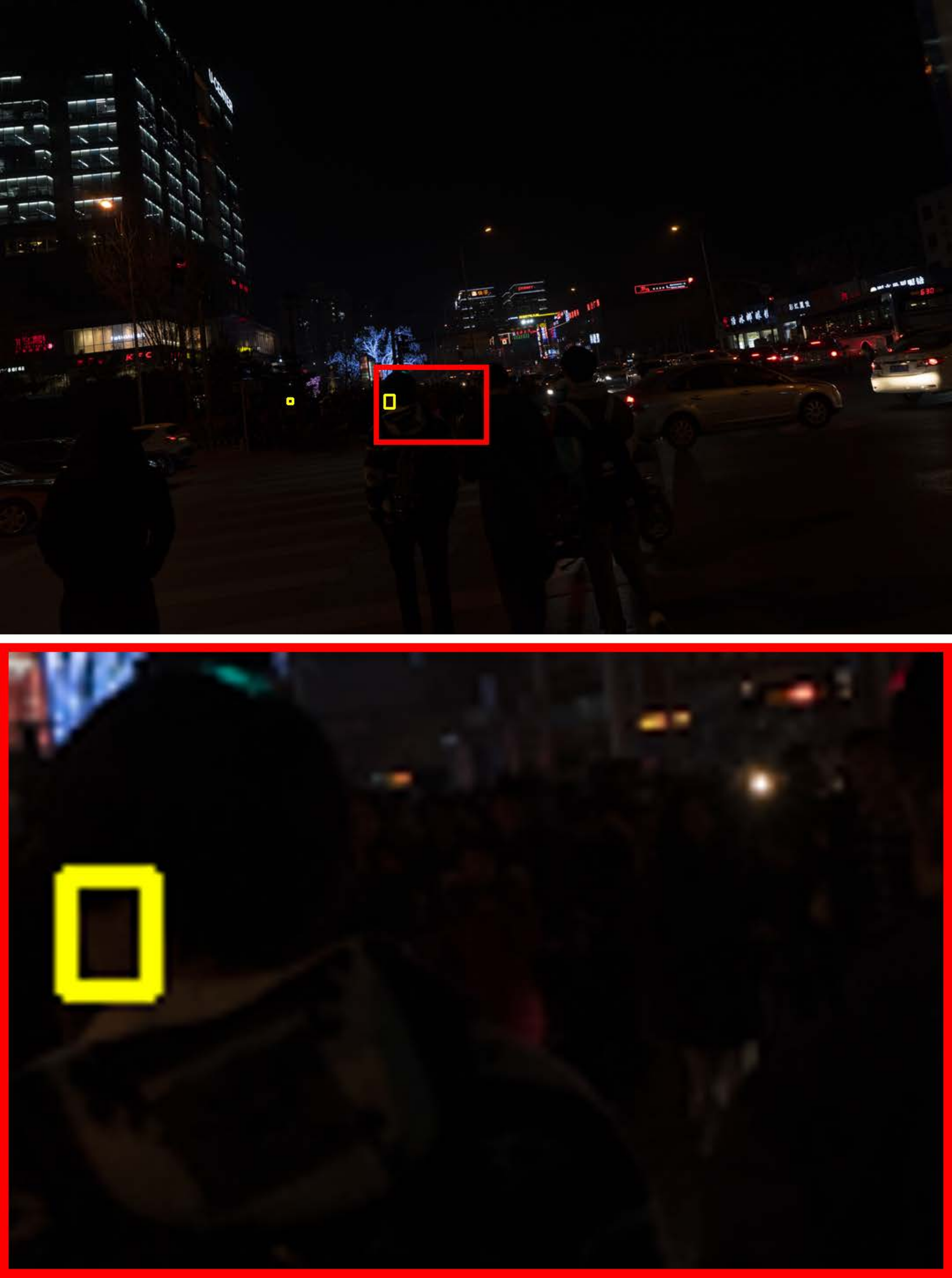}&
		\includegraphics[width=0.090\textwidth]{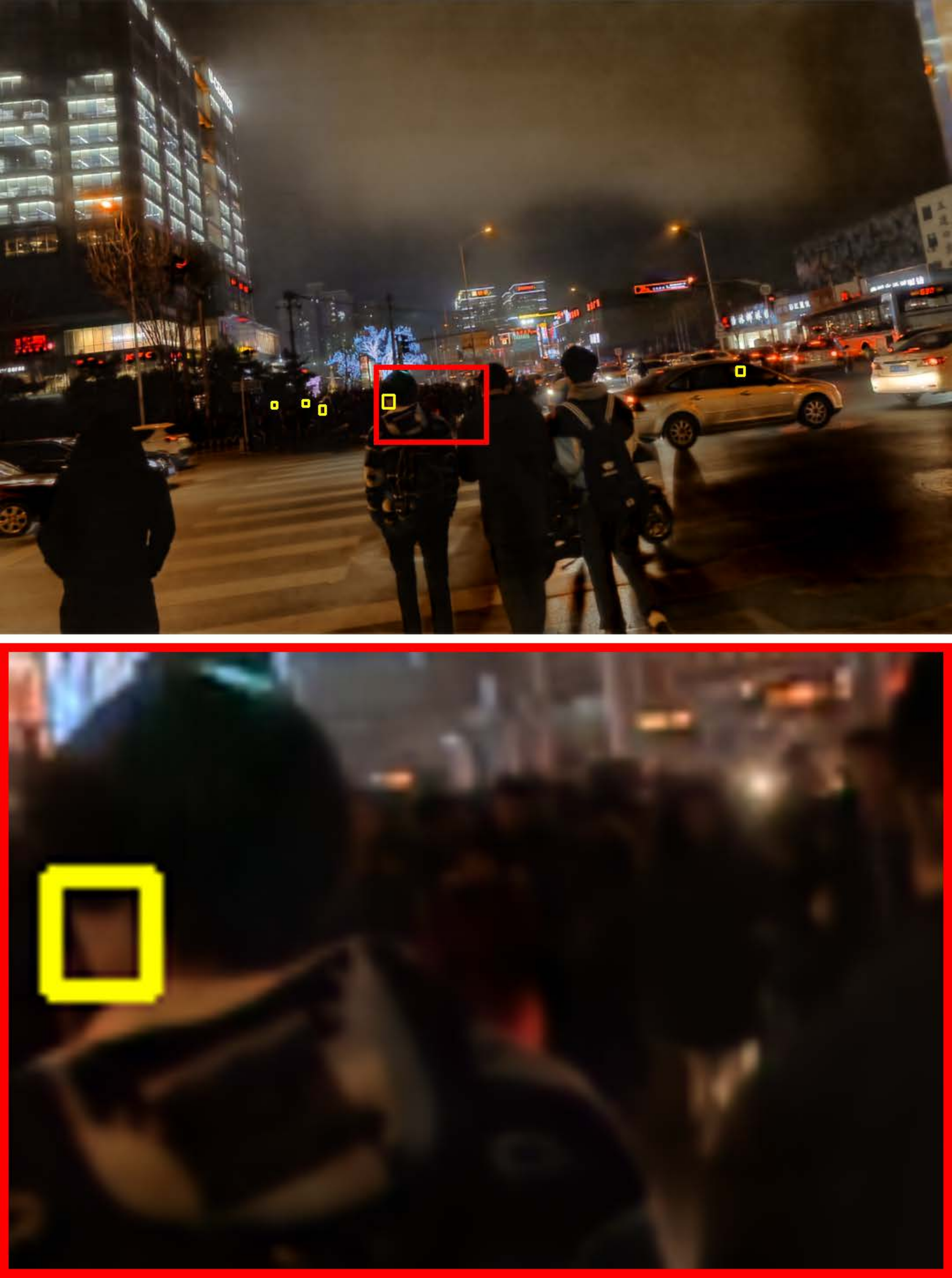}&
		\includegraphics[width=0.090\textwidth]{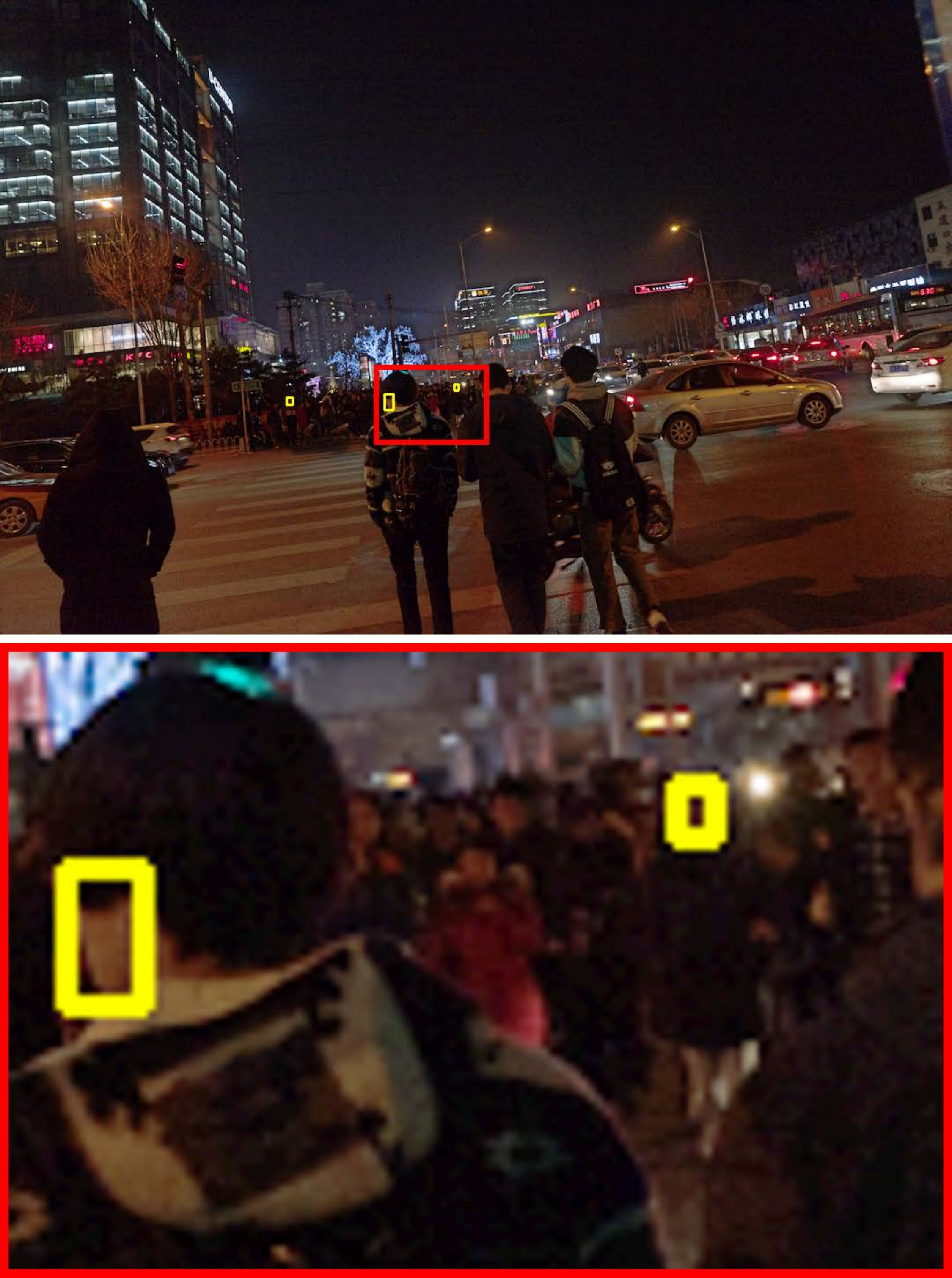}&
		\includegraphics[width=0.090\textwidth]{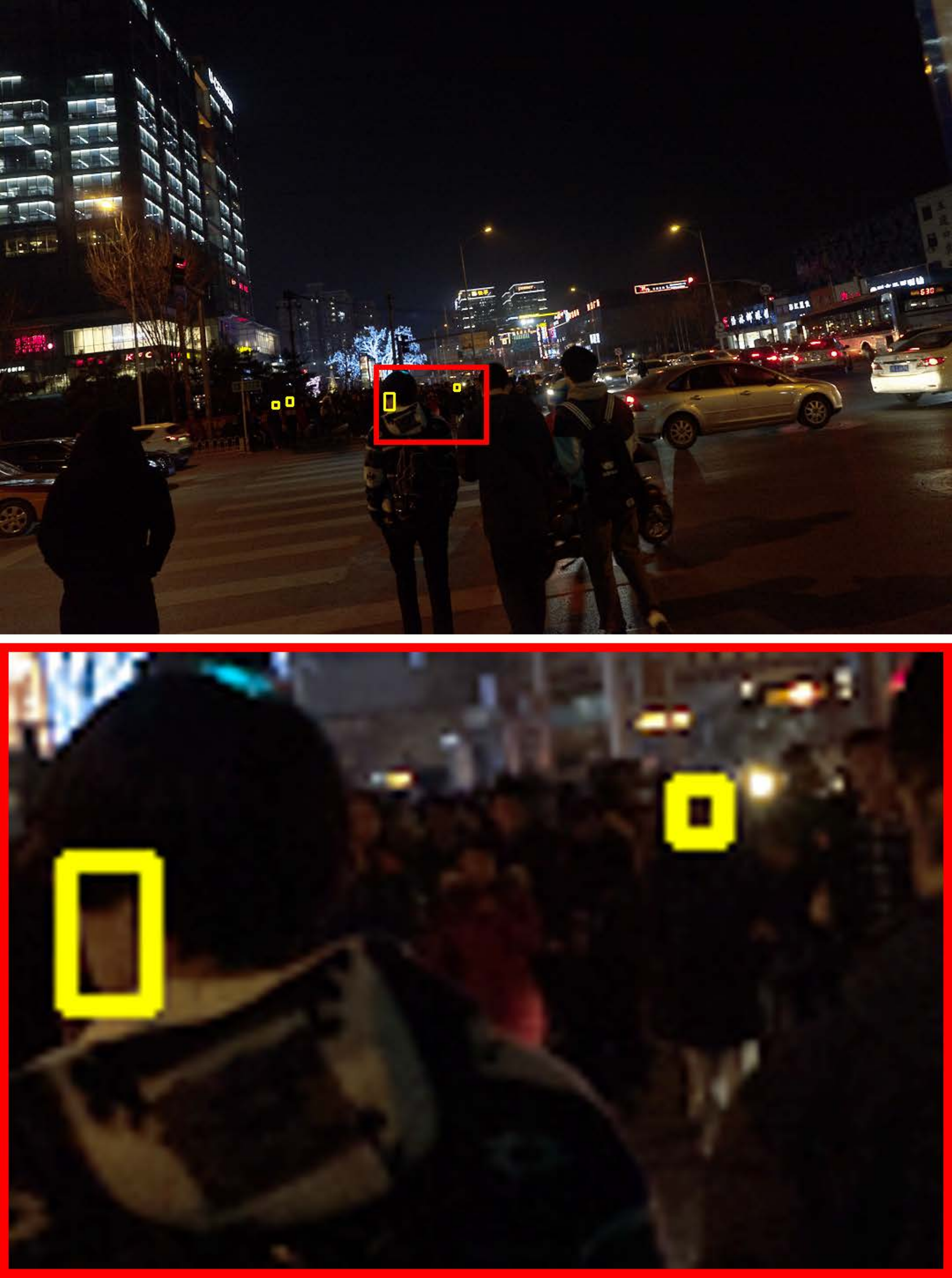}&
		\includegraphics[width=0.090\textwidth]{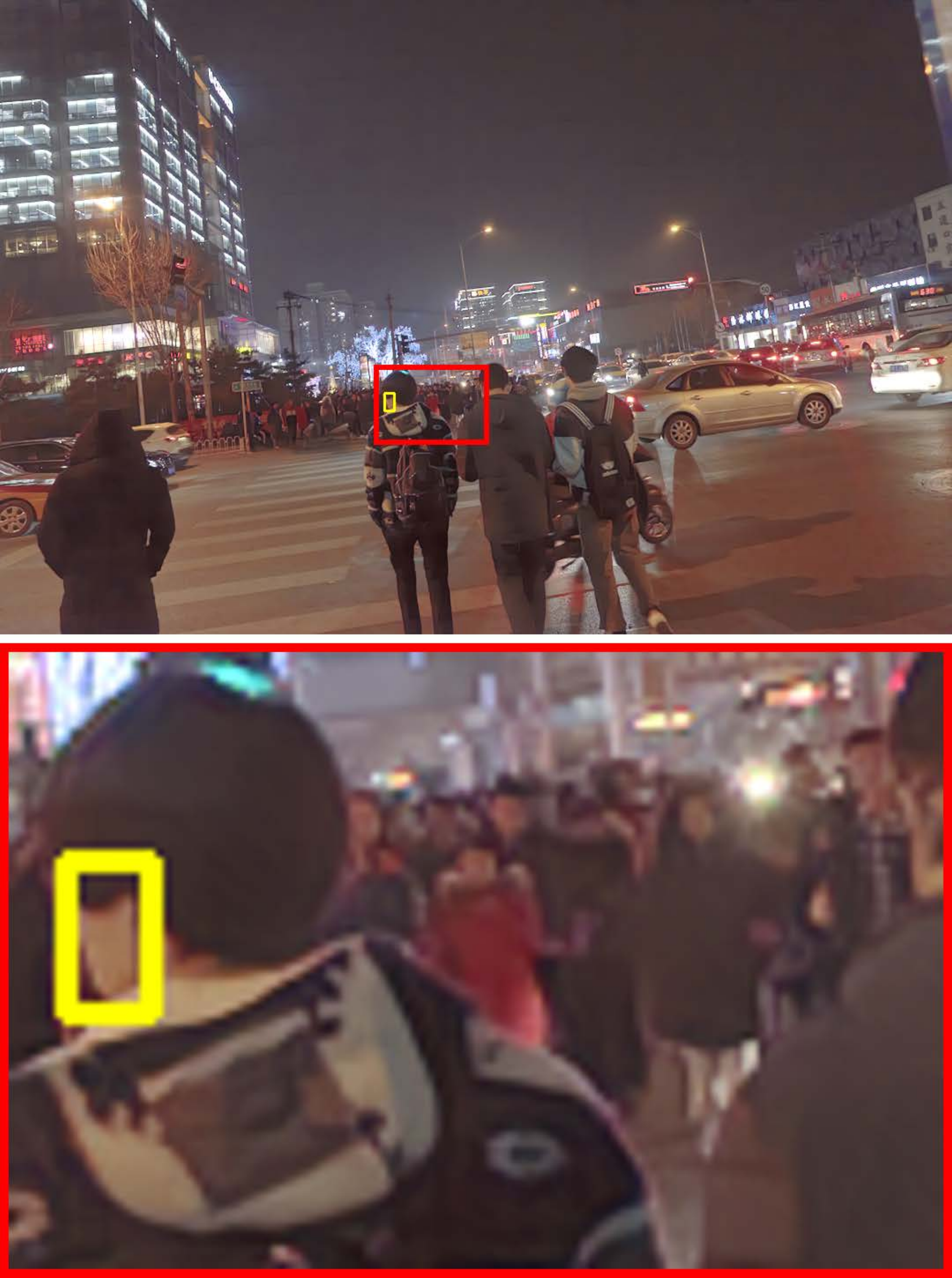}&
		\includegraphics[width=0.090\textwidth]{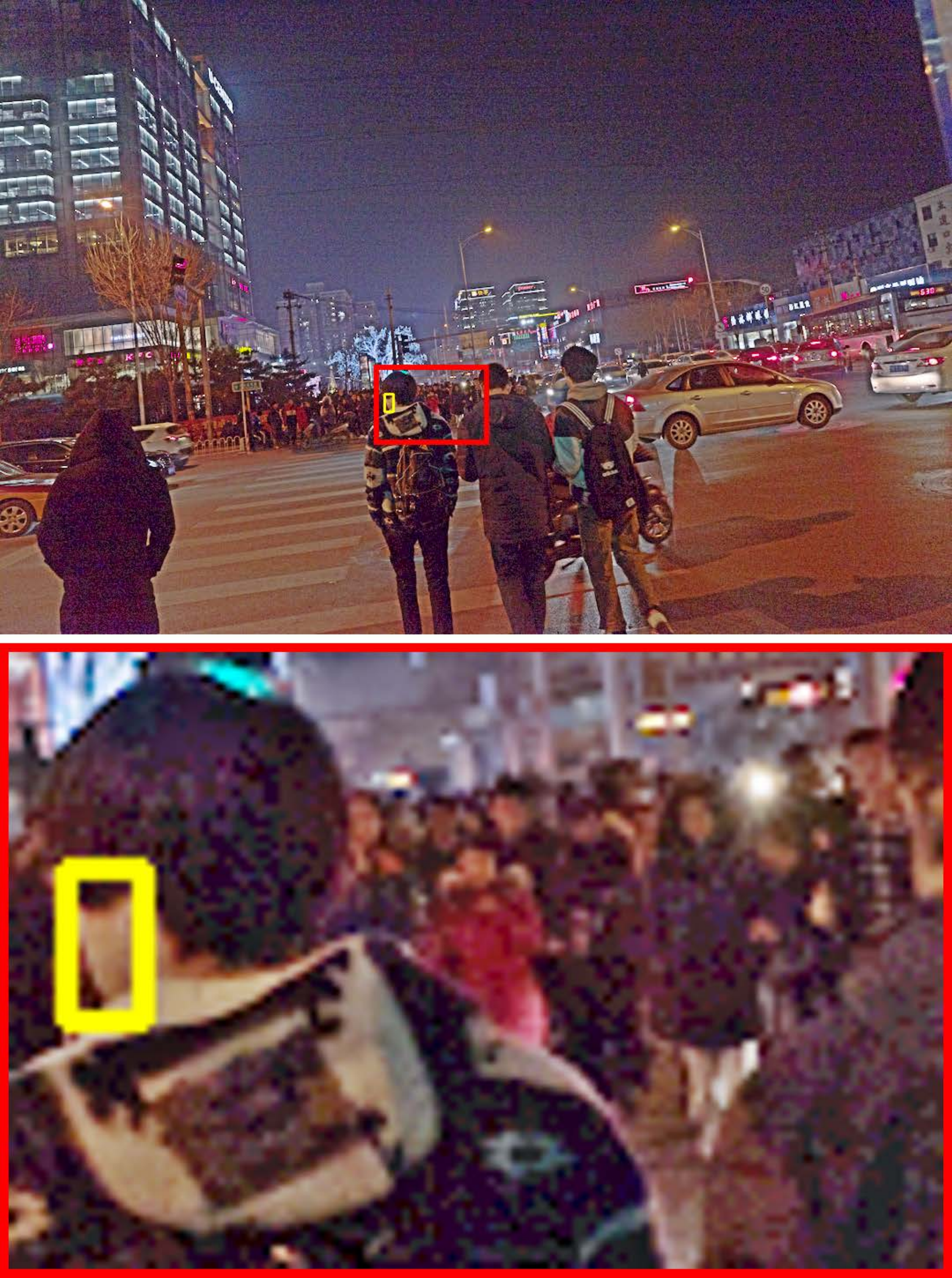}&
		\includegraphics[width=0.090\textwidth]{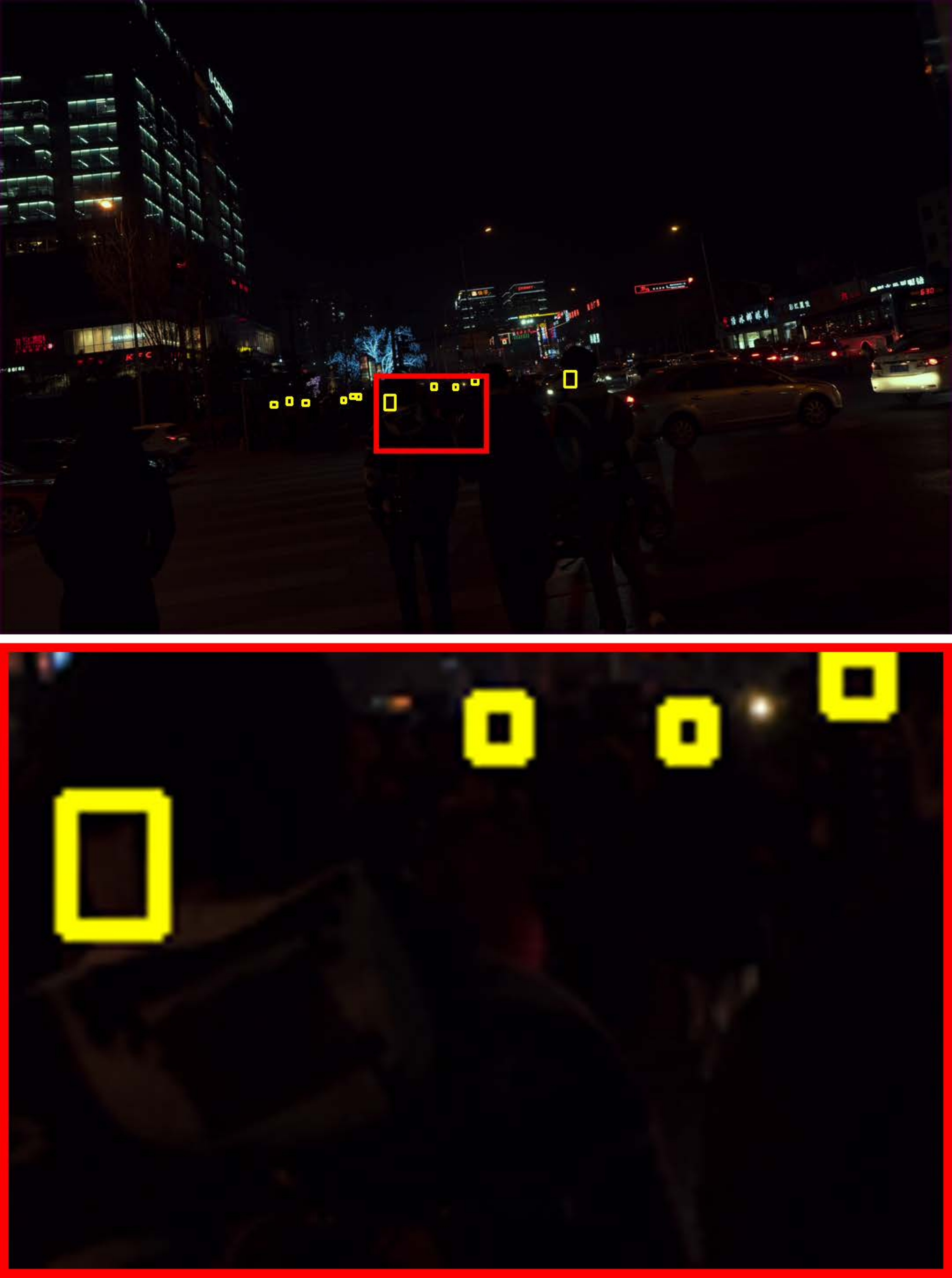}&
		\includegraphics[width=0.090\textwidth]{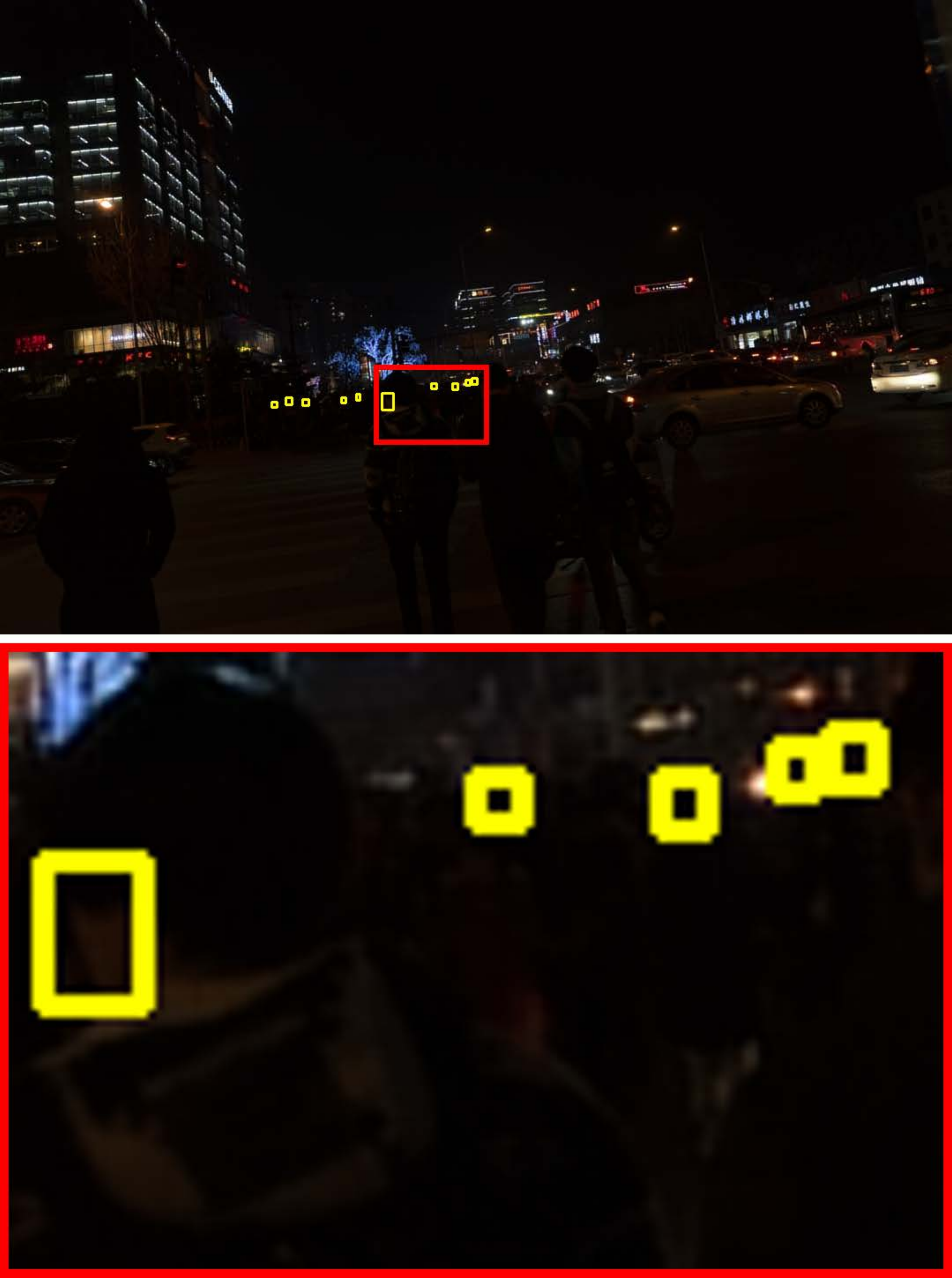}\\
		
		\includegraphics[width=0.090\textwidth]{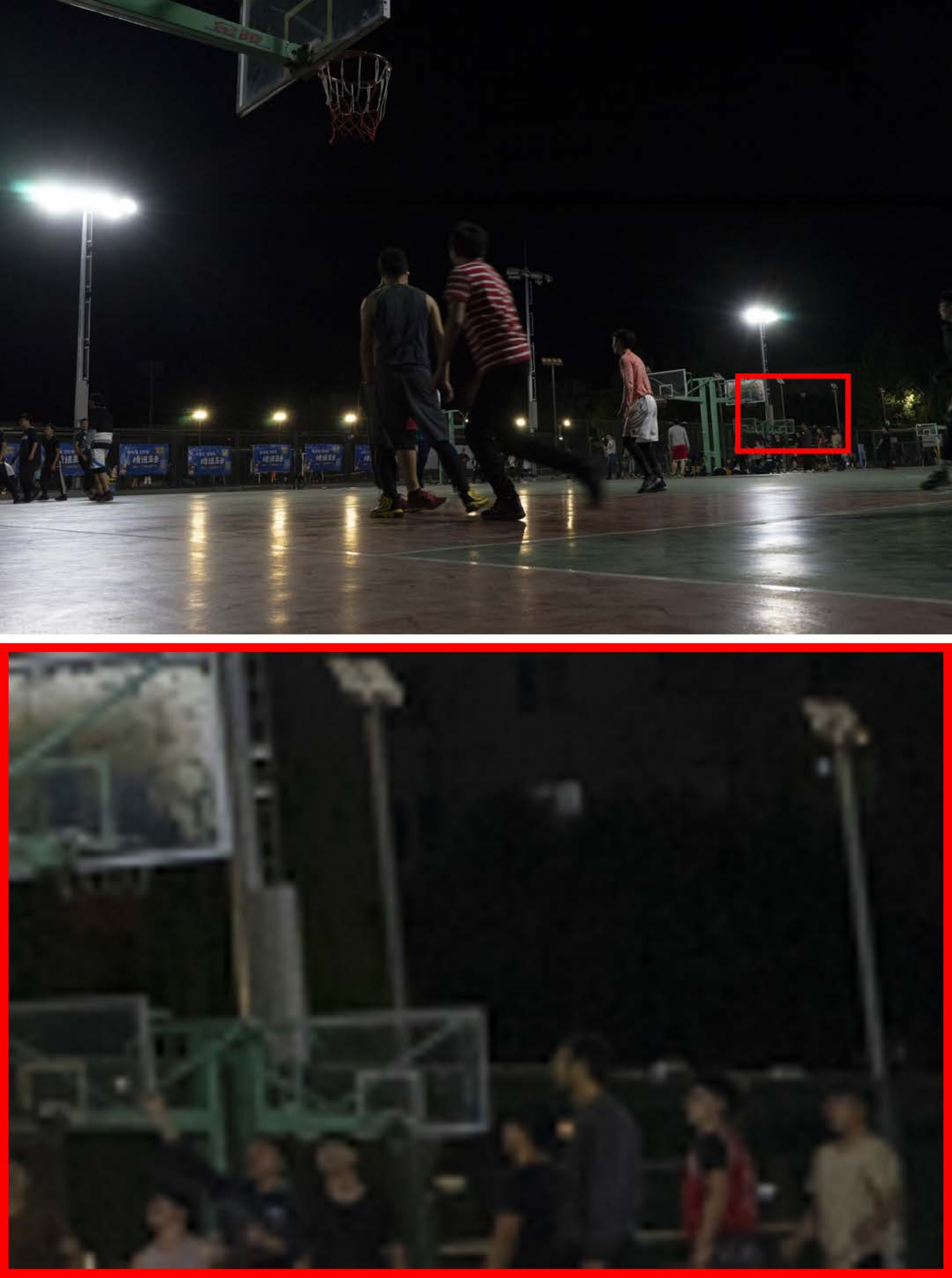}&
		\includegraphics[width=0.090\textwidth]{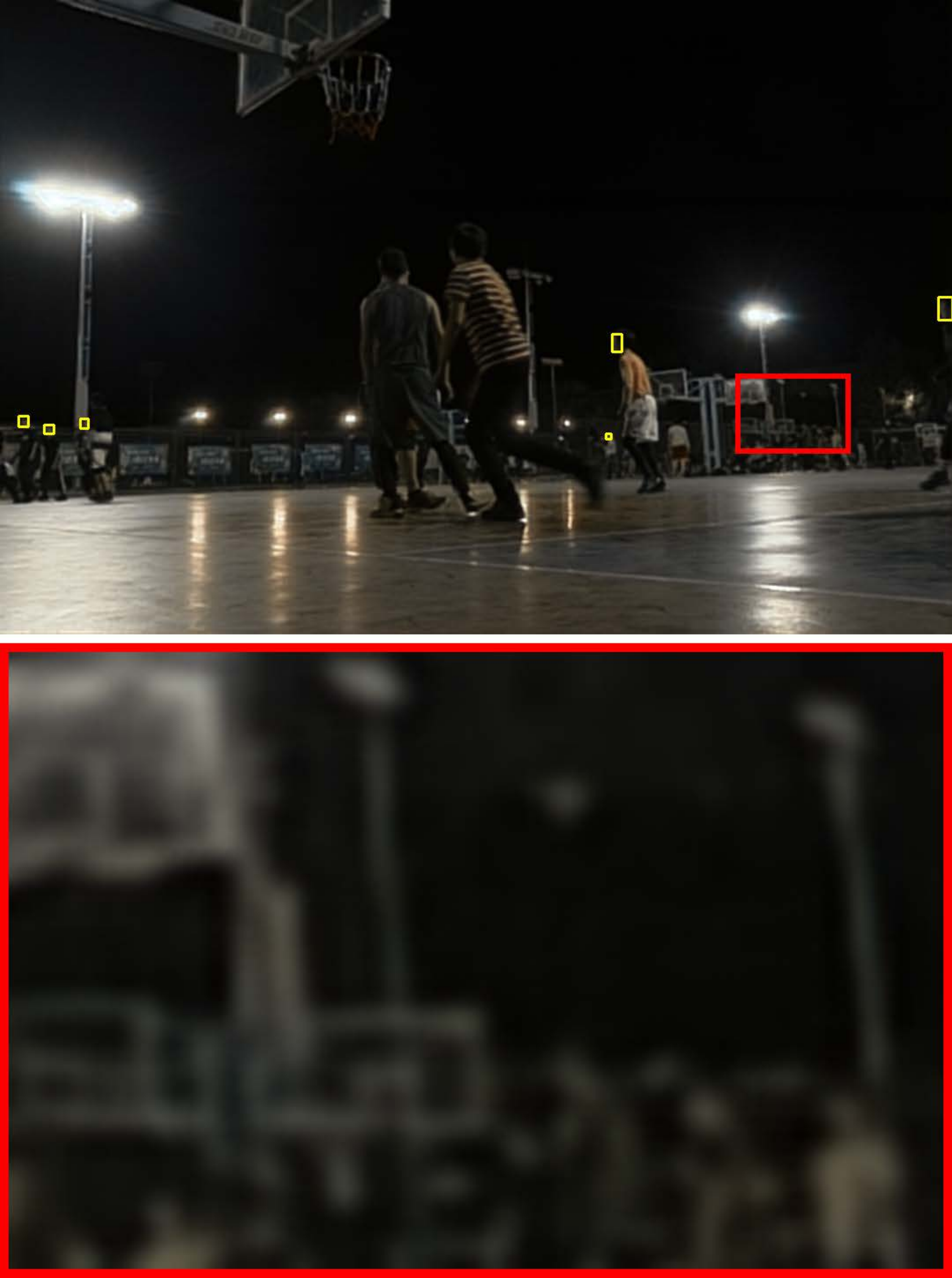}&
		\includegraphics[width=0.090\textwidth]{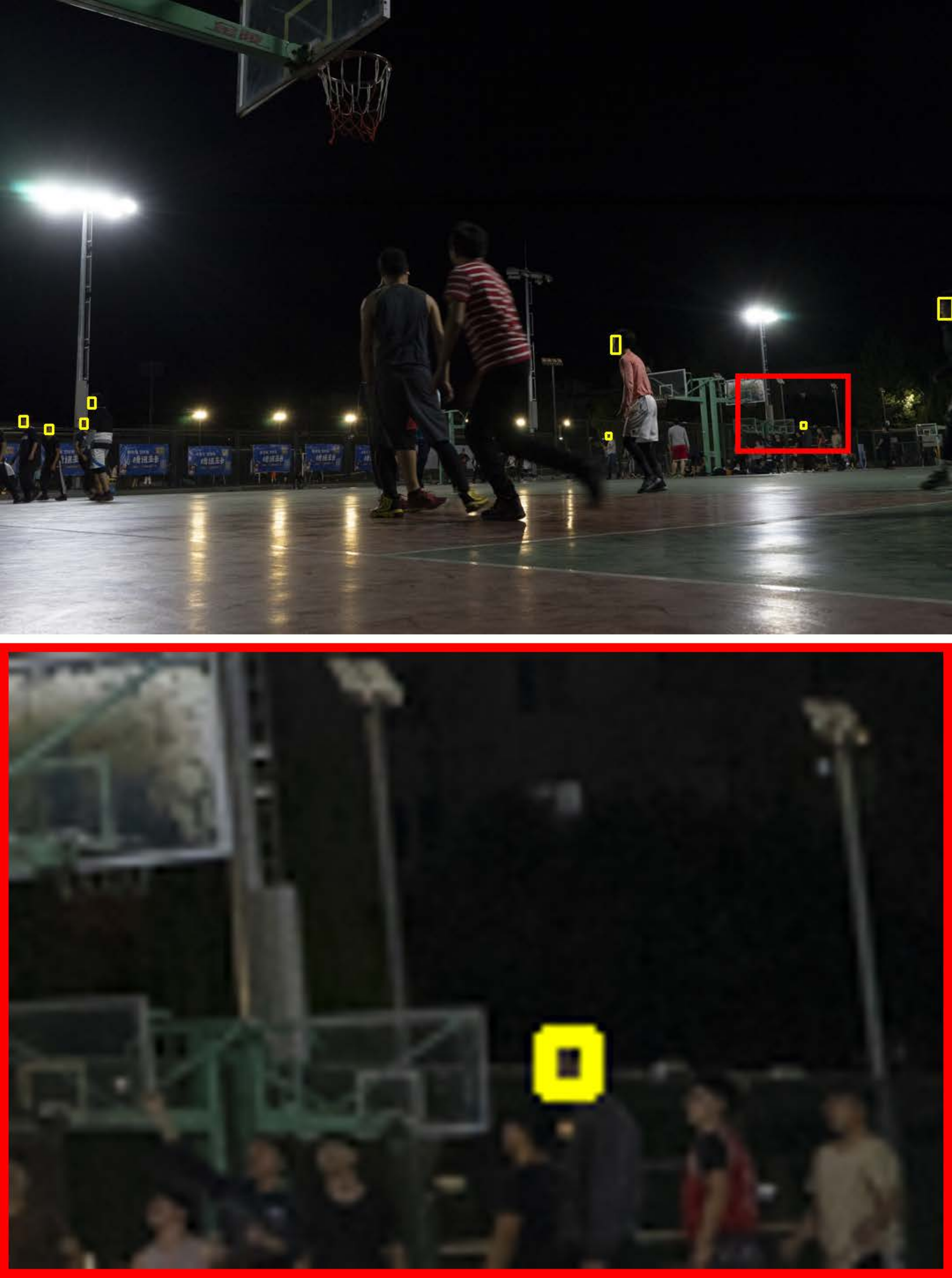}&
		\includegraphics[width=0.090\textwidth]{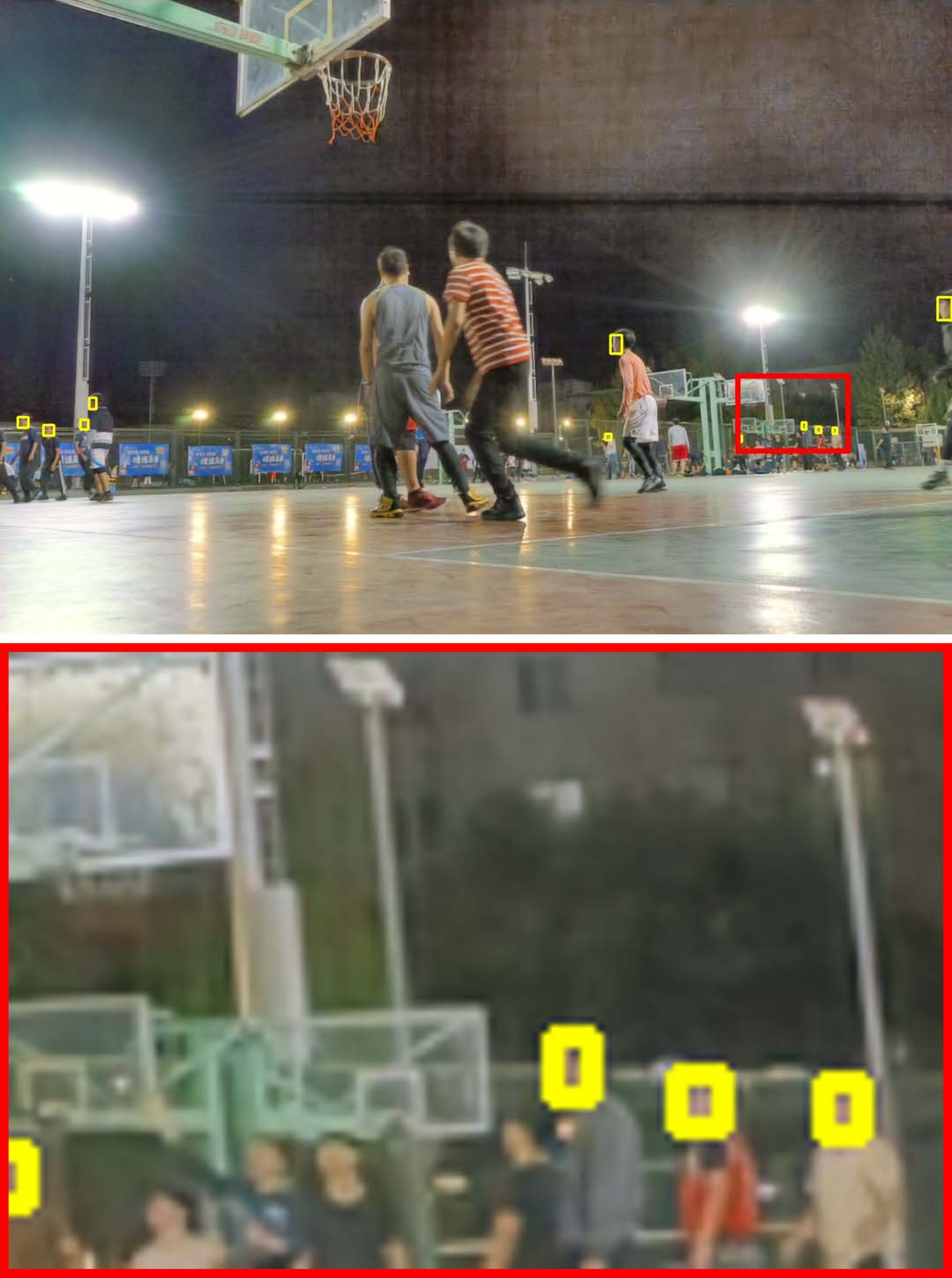}&
		\includegraphics[width=0.090\textwidth]{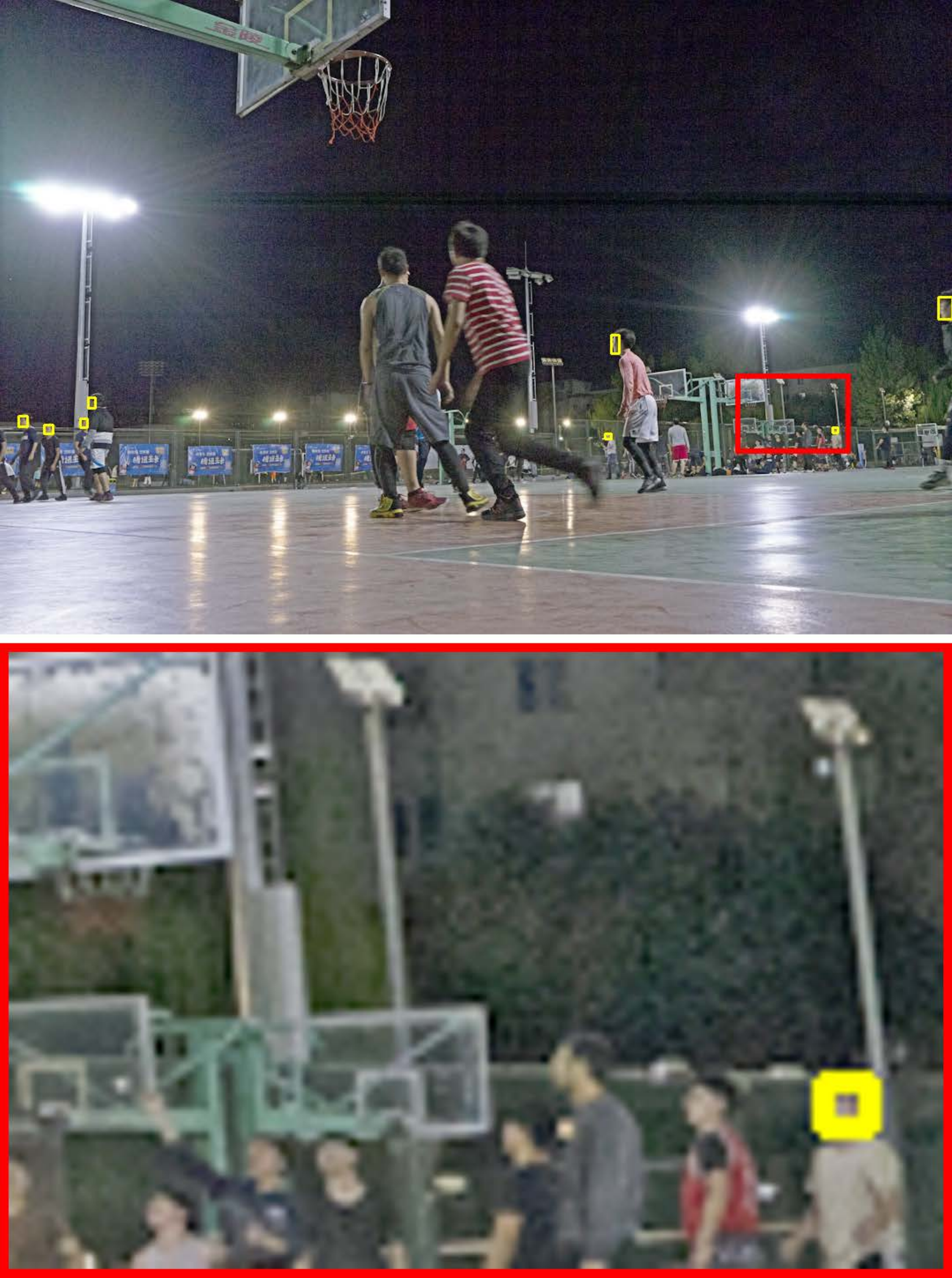}&
		\includegraphics[width=0.090\textwidth]{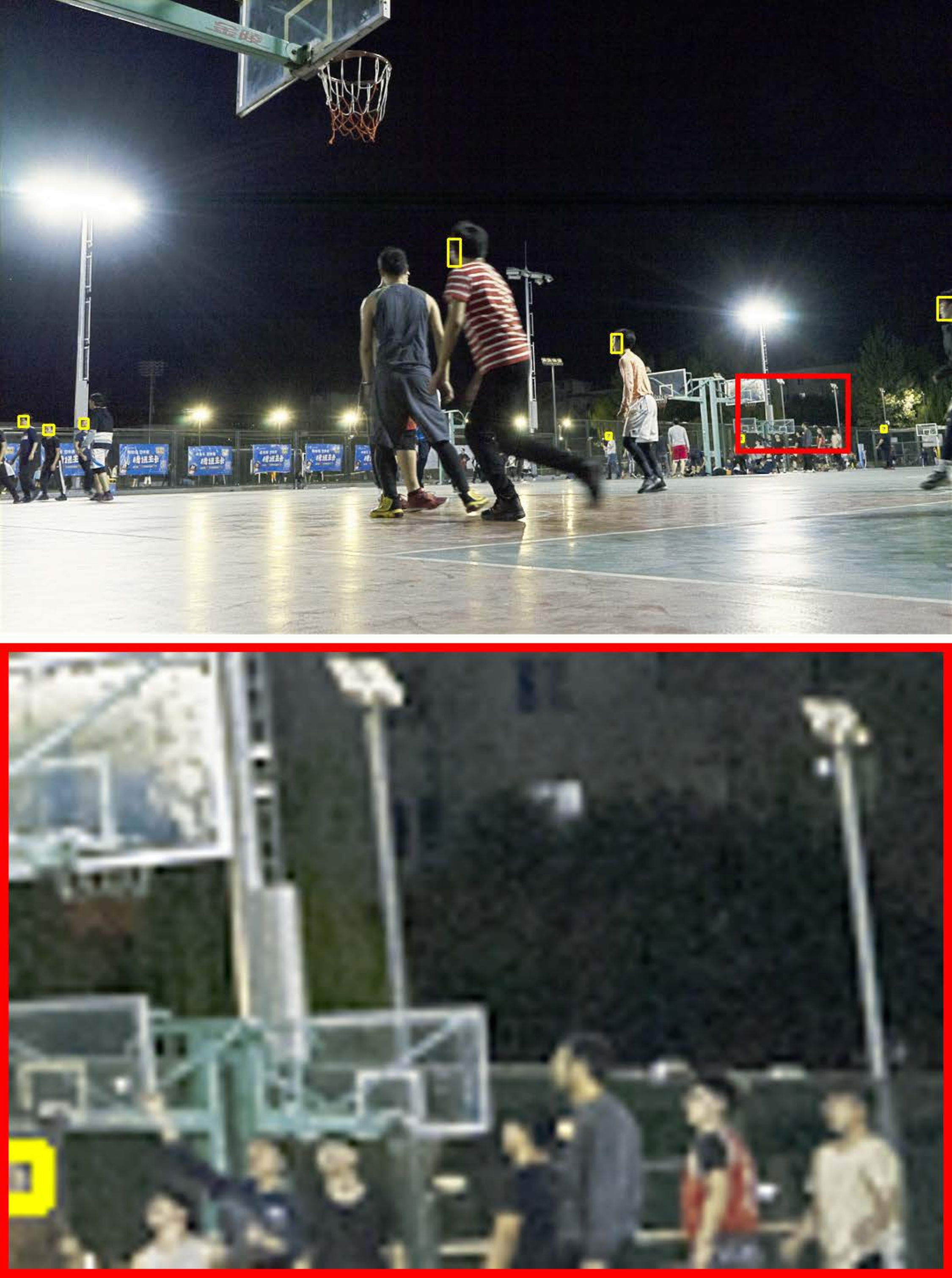}&
		\includegraphics[width=0.090\textwidth]{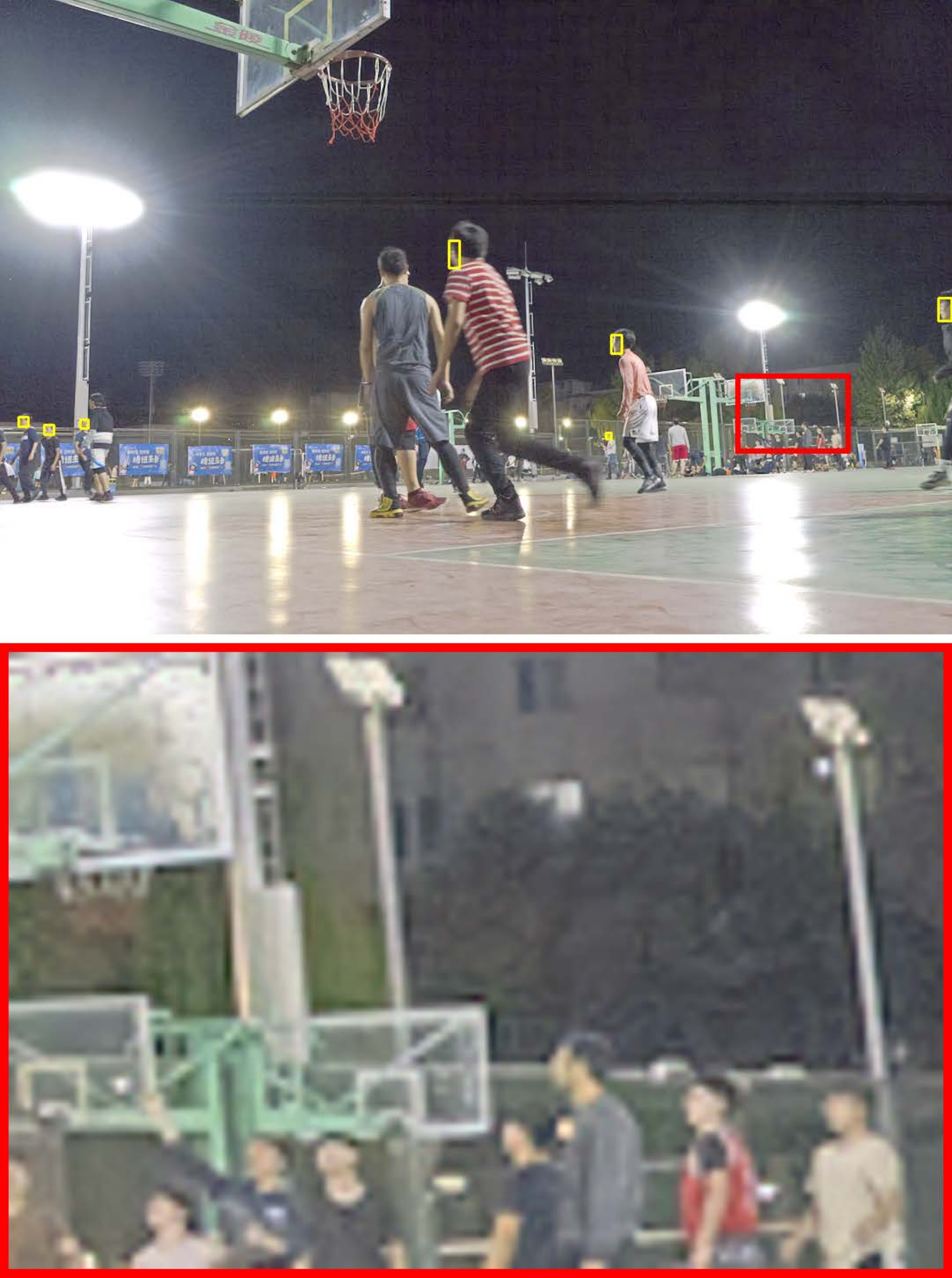}&
		\includegraphics[width=0.090\textwidth]{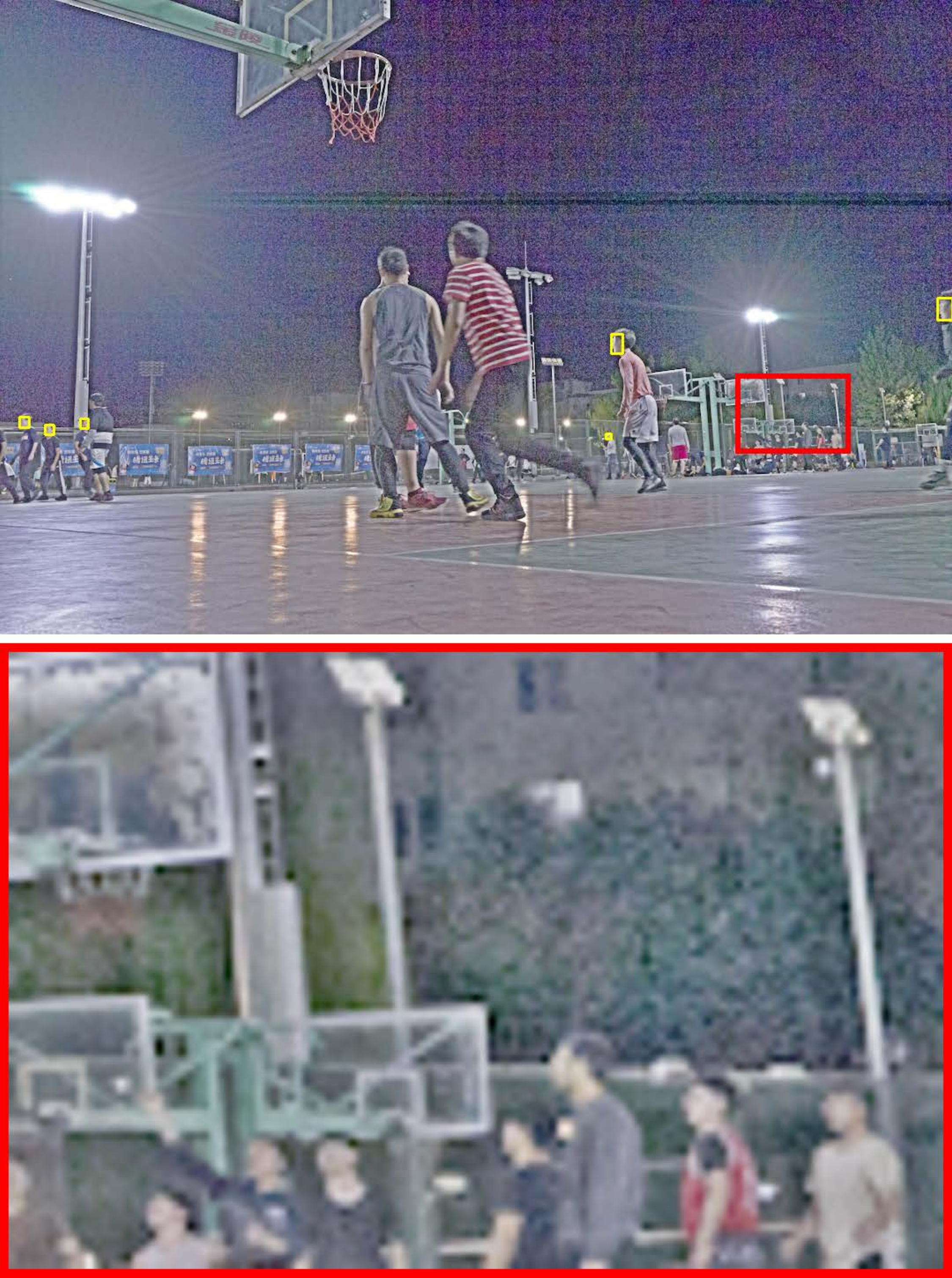}&
		\includegraphics[width=0.090\textwidth]{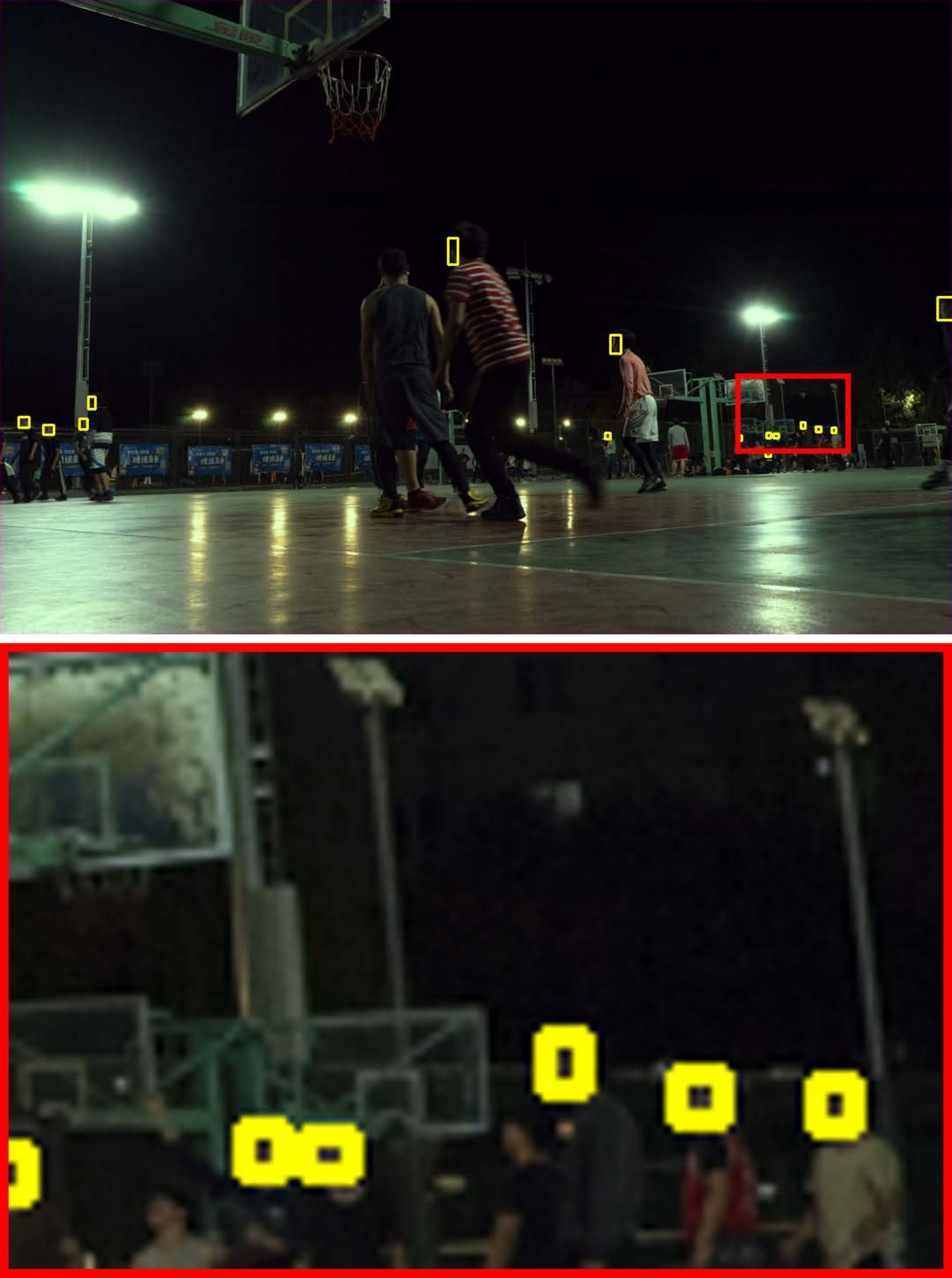}&
		\includegraphics[width=0.090\textwidth]{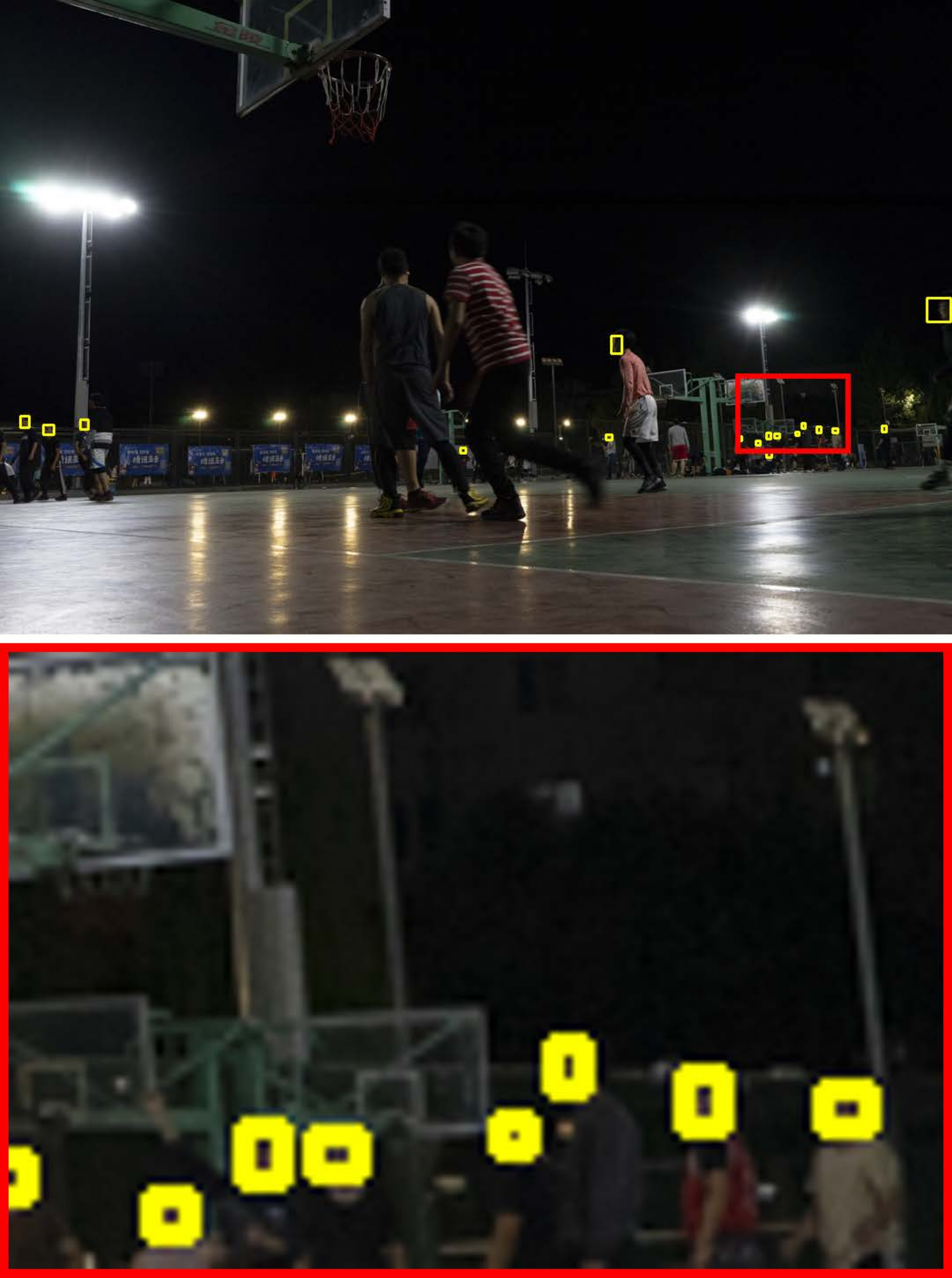}\\
		Input & SGN & REG& DRBN& ZeroDCE& SCI& URetinex& HLA& Ours & \small{Ground Truth}\\
	\end{tabular}
	
	\caption{Visual comparison  results against state-of-the-art low-light image enhancement approaches on the DARK FACE. }
	\label{fig:darkface_fig}
\end{figure*}

\subsection{Architectures and Loss Functions}
In detail, we introduce a U-Net type encoder-decoder to construct the neural architecture of generative block, which contains four layers of convolutions to extract the multi-scale features and introduce the skip connections to fuse features from encoder and decoder. We consider two classical low-light vision tasks, \textit{i.e.}, visual enhancement and object detection. As for the  visual enhancement, designating the enhancement module as $\mathcal{G}$. we introduce the Retinex theory~\cite{wei2018deep} to estimate the normal illumination with a light-weight architecture design. We establish three cascaded convolution layers to generate the  four three-channel illumination, which couples with one single  skip connection.  Visual enhancement module can be formulated as $\mathbf{y} =  \mathbf{x} \oslash\mathcal{G}(\mathbf{x},\bm{\theta})$. As for the dark face detection, we introduce a well-known  face detector DSFD~\cite{li2019dsfd} as $\mathcal{G}$.
Noting that, our framework also can realize other low-light scene perception tasks with strong generalization. The L1 loss is defined as  $\mathcal{L}_\mathcal{F}$ in the generative block.
We utilize the self-supervised illumination learning loss to define $\mathcal{L}_\mathcal{G}$ on enhancement.  Multibox objective learning loss is utilized to define $\mathcal{L}_\mathcal{G}$ on detection.

\section{Experiments}

\subsection{Implementation Details}

\textit{Benchmarks description and metrics.} 
We utilized three datasets to conduct our experiments, including SID~\cite{chen2018learning}, DARK FACE ~\cite{yang2020advancing} and ACDC~\cite{sakaridis2021acdc}. For the SID dataset, we sampled 2697 short-exposure images and 231 long-exposure reference images based on the Sony $\alpha$7S II sensor for training, and 598 image pairs for testing. We randomly selected 400 pairs of data for training and 106 pairs for testing on the ACDC dataset. For the DARK FACE dataset, we used 5000 images for training and 1000 images for testing in the context of object detection. It is worth noting that due to the unavailability of raw data on the DARK FACE dataset, we obtained raw data by means of an unprocess~\cite{chen2023instance} approach. For the image enhancement, we used eight metrics including PSNR, SSIM, LPIPS~\cite{zhang2018unreasonable}, VIF~\cite{sheikh2006image}, FSIM~\cite{zhang2011fsim}, BRISQUE~\cite{mittal2012no}, NIQE~\cite{mittal2012making} and LOE~\cite{wang2013naturalness}. The  metrics of mAP and mIOU are used for dark face detection and low-light semantic segmentation respectively.

\textit{Parameter settings.} 
During the training phase for the enhancement task, we applied data augmentation methods throughout the training stages and cropped the images to a size of 512 $\times$ 512 for the SID dataset. We set the initial learning rate to $10^{-4}$ during the warm start phase. In the bilevel training phase, we used the IBGL solver to achieve better performance. The learning rate of the upper model $\mathcal{F}$ was set to $3\times 10^{-3} $ and decayed to $5\times 10^{-6} $, while the learning rate of the lower model $\mathcal{G}$ was set to $3\times 10^{-3} $ and decayed to $3\times 10^{-5} $. With a batch size of 2, we set $\mathbf{k}$ = 80 iterations for optimizing the generative block and updating the parameters of the underlying model at each iteration. 

For the generative block trained on detection task, we use SGD optimizers with an initial learning rate of $3\times 10^{-3} $ and $5\times 10^{-4} $ for generative block and detection model with  5e-4 weight decay and 0.9 momentum for both optimizers. 
The input images were cropped to a size of 640 $\times$ 640 and the batch size was set to 8. In the bilevel training phase, we set $\mathbf{k}$ = 30. While training the segmentation task, the batch size is changed to 5 and the initial learning rate of the optimizer is set to 0.1. In the process of training the generation block on the enhancement task, the ADAM optimizer with parameters ( 0.9, 0.999 ) and initial learning rate of $3\times 10^{-3} $ is used for enhancement model. The following parameter configurations remain unchanged from the previous section. All  experiments were performed on an NVIDIA RTX 3090 GPU using PyTorch framework.

\subsection{Low-Light Image Enhancement}
\begin{table*}[htb]
	\centering
	\renewcommand{\arraystretch}{1.05}
	\caption{ Qualitative results compared with state-of-the-art methods on ACDC dataset. The best result is in \textcolor{red}{\textbf{red}} whereas the second one is in \textcolor{blue}{\textbf{blue}}. The symbol set \{RO, SI, BU, WA, FE, PO, TL, TS, VE, TE, SK, PE, RI, CA, TR, BI \} represents \{road, sidewalk, building, wall, fence, pole, traffic light, traffic sign, vegetation, terrain, sky, person, rider, car,  train, bicycle \}.}
	\setlength{\tabcolsep}{0.86mm}{
		\begin{tabular}{|c|cccccccccccccccc|c|}
			\hline
			Methods & RO   &  SI &  BU  & WA & FE & PO & TL& TS&VE&TE&SK&PE&RI&CA&TR&BI&mIOU$\uparrow$\\\hline
			RetineNet &89.43&61.00&74.21&32.86&28.13&42.43&49.82&25.71&65.87&8.63&77.37&21.55&\textcolor{red}{\textbf{13.88}}&54.88&67.44&8.20&41.96  \\
			\hline
			DRBN &90.51&61.51&72.81&30.16&32.54&44.56&47.63&27.22&65.74&10.20&76.56&24.28&13.24&55.47&71.12&11.94&43.38\\
			\hline
			FIDE &90.00&60.74&72.81&32.48&34.13&43.32&47.91&26.10&67.04&13.79&78.00&26.57&5.80&57.10&71.04&12.48&43.47 \\
			\hline
			KinD &90.00&61.02&73.25&31.96&32.89&43.58&47.91&27.76&65.55&13.33&77.47&22.83&8.10&55.16&74.57&11.56&43.00\\
			\hline
			EnGAN &89.74&58.96&73.56&32.84&31.82&\textcolor{blue}{\textbf{44.74}}&42.75&26.28&\textcolor{red}{\textbf{67.33}}&\textcolor{blue}{\textbf{14.28}}&77.88&25.00&10.69&59.00&71.28&7.80&43.85\\
			\hline
			SSIENet  &89.66&59.38&72.58&29.90&31.72&\textcolor{red}{\textbf{45.46}}&43.90&24.50&66.75&10.68&78.39&22.82&0.22&52.67&71.12&5.48&41.47\\
			\hline
			ZeroDCE&89.63&59.93&73.94&32.65&31.76&44.35&46.27&25.88&67.20&\textcolor{red}{\textbf{14.67}}&79.10&24.73&7.76&13.29&66.83&13.99&43.42\\
			\hline
			SCI&92.29&59.93&69.22&39.78&\textcolor{blue}{\textbf{39.56}}&37.18&44.78&33.72&39.82&5.03&67.78&26.03&30.03&59.44&70.91&7.89&44.06\\
			\hline
			SUC&\textcolor{blue}{\textbf{92.40}}&\textcolor{blue}{\textbf{69.55}}&\textcolor{blue}{\textbf{76.00}}&39.57&34.70&44.70&\textcolor{red}{\textbf{54.15}}&\textcolor{blue}{\textbf{39.10}}&67.07&11.22&\textcolor{blue}{\textbf{79.32}}&25.24&10.44&\textcolor{red}{\textbf{69.77}}&\textcolor{blue}{\textbf{84.76}}&\textcolor{blue}{\textbf{27.43}}&\textcolor{blue}{\textbf{44.34}}\\
			\hline
			URetinex&90.38&65.59&75.81&\textcolor{blue}{\textbf{40.10}}&35.95&43.59&53.51&\textcolor{red}{\textbf{40.79}}&65.07&8.30&78.81&\textcolor{blue}{\textbf{26.95}}&\textcolor{blue}{\textbf{13.48}}&\textcolor{blue}{\textbf{67.31}}&80.90&25.58&43.82\\
			\hline
			Ours &\textcolor{red}{\textbf{93.06}}&\textcolor{red}{\textbf{70.27}}&\textcolor{red}{\textbf{76.31}}&\textcolor{red}{\textbf{41.76}}&\textcolor{red}{\textbf{40.74}}&44.08&\textcolor{blue}{\textbf{53.61}}&34.40&\textcolor{blue}{\textbf{68.53}}&9.40&\textcolor{red}{\textbf{79.38}}&\textcolor{red}{\textbf{27.19}}&8.01&66.90&\textcolor{red}{\textbf{85.05}}&\textcolor{red}{\textbf{45.76}}&\textcolor{red}{\textbf{44.91}}\\
			\hline
		\end{tabular}	
	}
	\label{tab:ACDC_tab}
\end{table*}

\textit{Qualitative comparison.} Meanwhile, we also make comparisons among these methods in Fig.~\ref{fig:SID_fig}. There are three discriminative advantages. Firstly, our approach efficiently preserves the texture details present in the image,  as shown in the zoom-out region of the second row, the results generated by AbaNet and SGN exhibit obvious texture detail distortion and our findings demonstrate the ability to depict a greater level of detail.   Secondly, Our method preserves the natural colour distribution, \textit{e.g.}, in the case of glass and trademark. Although RED recovers the clear background with vivid colours, the texture details  severely corrupted. Thirdly, our approach can effectively reduce a significant amount of noise from the image, as shown in the zoom-out region of the first row, in comparison to the outcomes produced by RED, our results exhibit a higher level of clarity.  Additionally, the results of our method exhibit abundant textural details.

\textit{Quantitative comparison.} We performed a comprehensive evaluation of various methods on the SID, including  SID~\cite{chen2018learning}, SGN~\cite{gu2019self},  DID~\cite{maharjan2019improving}, LLPack~\cite{lamba2020towards},  RED~\cite{lamba2021restoring}, RawFormer~\cite{xu2022rawformer} and AbaNet~\cite{dong2022abandoning}. 
We observed that our approach achieves significant improvements over RED method, with a 0.31dB  increase in PSNR  and the lowest perceptual errors based on LPIPS. A higher PSNR value indicates that the restored image is more similar to the ground truth image on pixel intensity. LPIPS  measures the perceptual similarity incline with human visual perception.

\subsection{Dark Face Detection}

\textit{Qualitative comparison.} Fig.~\ref{fig:darkface_fig} showcases visual comparisons with advanced methods. For the enhanced images, although they exhibit noticeable enhancement effects, they may still fail to detect relatively small faces. In contrast, our method can effectively detect smaller faces and significantly improve precision, particularly in challenging scenes, for exmaple, crowded  in the distance. In the last row of images, overexposed images such as HLA can be observed, and these can lead to a reduction in detection accuracy. Furthermore, compared to  the ``SID +detection" method, our final results retain more texture detail information.

\begin{figure}[htb]   
	\centering
	\begin{tabular}{c} 
		\includegraphics[width=0.48\textwidth]{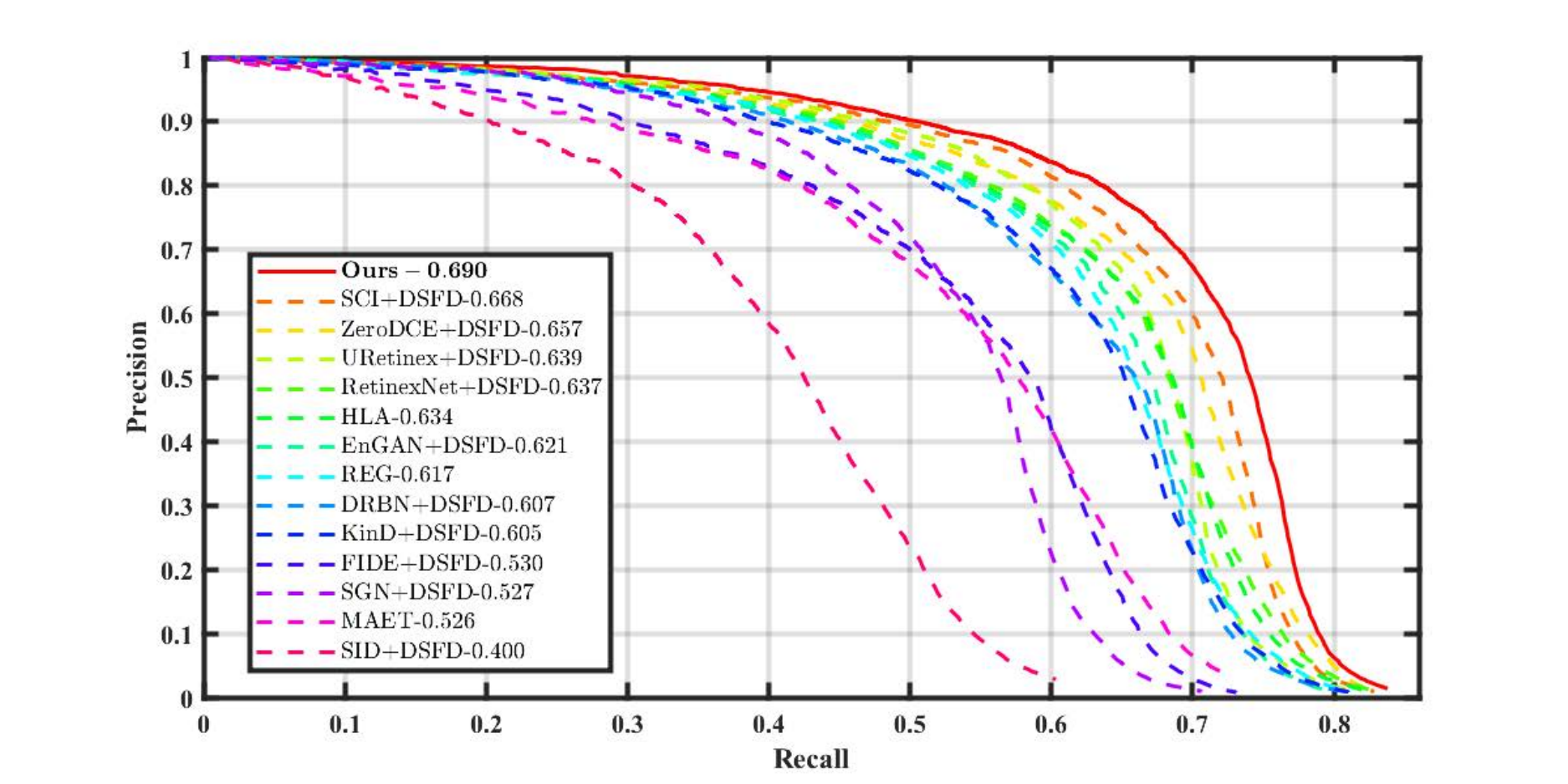}\\
	\end{tabular}
	\caption{Precision-Recall curve of detection results on the DARK FACE dataset.}
	\label{fig:PRcurve}
\end{figure}
\textit{Quantitative comparison.} We first plotted the Precision-Recall curve in Fig.~\ref{fig:PRcurve} to demonstrate the superiority of our method. In comparison to various  methods, including two ``raw+detection" methods ( SID, SGN ),  ``enhancement+detection" methods ( REG, FIDE~\cite{xu2020learning}, KinD~\cite{zhang2021beyond}, EnGAN~\cite{jiang2021enlightengan}, RetinexNet~\cite{wei2018deep},  DRBN~\cite{yang2020fidelity}, ZeroDCE~\cite{guo2020zero}, SCI~\cite{ma2022toward} and  URetinex~\cite{wu2022uretinex} ), all based on the same detector DSFD~\cite{li2019dsfd}, as well as three face detection methods, MAET~\cite{cui2021multitask}, REG~\cite{liang2021recurrent} and HLA~\cite{wang2022unsupervised}. Our method improves mAP by 3.29\% and 7.98\% compared to ``SCI+detection" and ``URetinex+detection". In a word, our method achieves the best detection precision.

\subsection{Nighttime Semantic Segmentation }
\begin{figure*}[htb]
	\centering
	\begin{tabular}{c@{\extracolsep{0.25em}}c@{\extracolsep{0.25em}}c@{\extracolsep{0.25em}}c@{\extracolsep{0.25em}}c@{\extracolsep{0.25em}}c@{\extracolsep{0.25em}}c@{\extracolsep{0.25em}}c@{\extracolsep{0.25em}}c} 
		\includegraphics[width=0.102\textwidth]{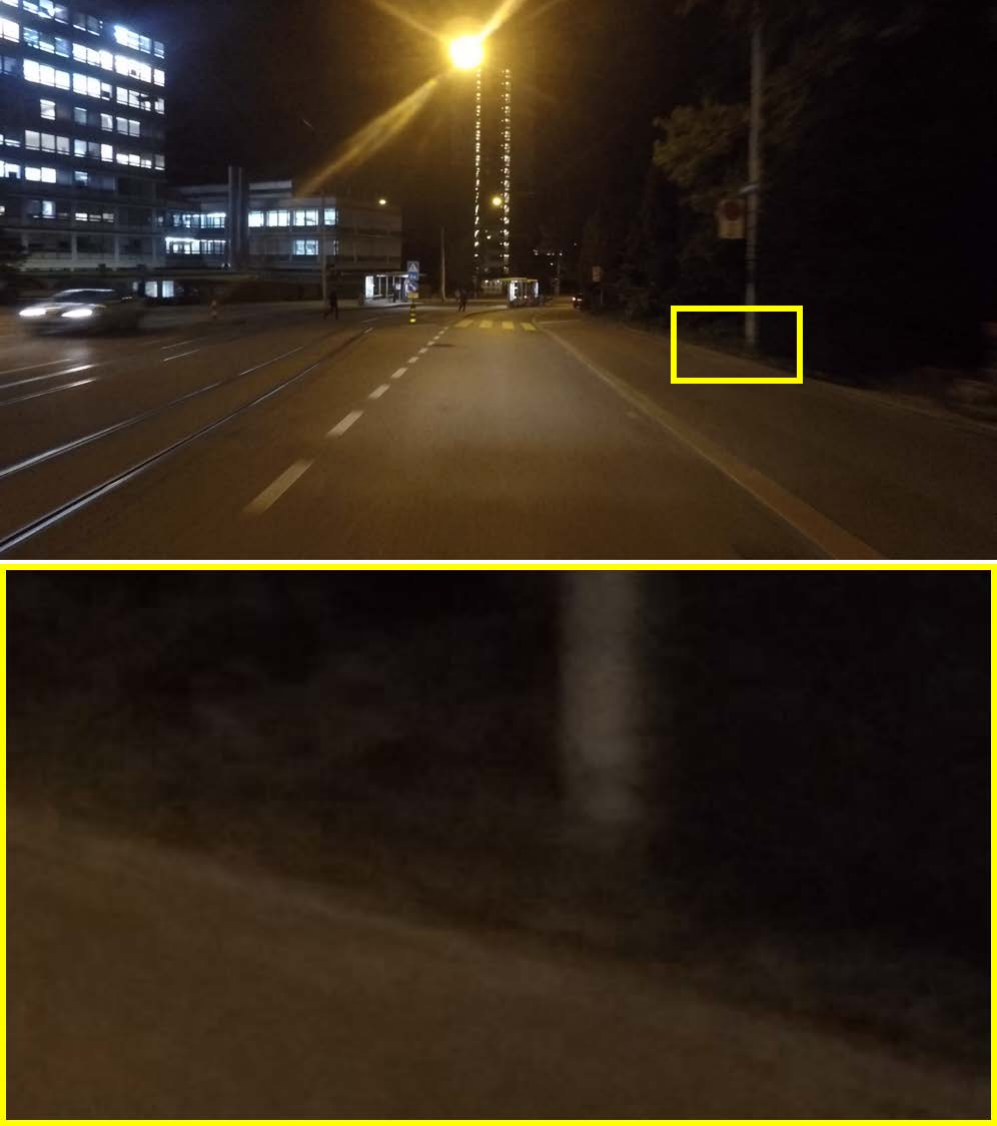}&
		\includegraphics[width=0.102\textwidth]{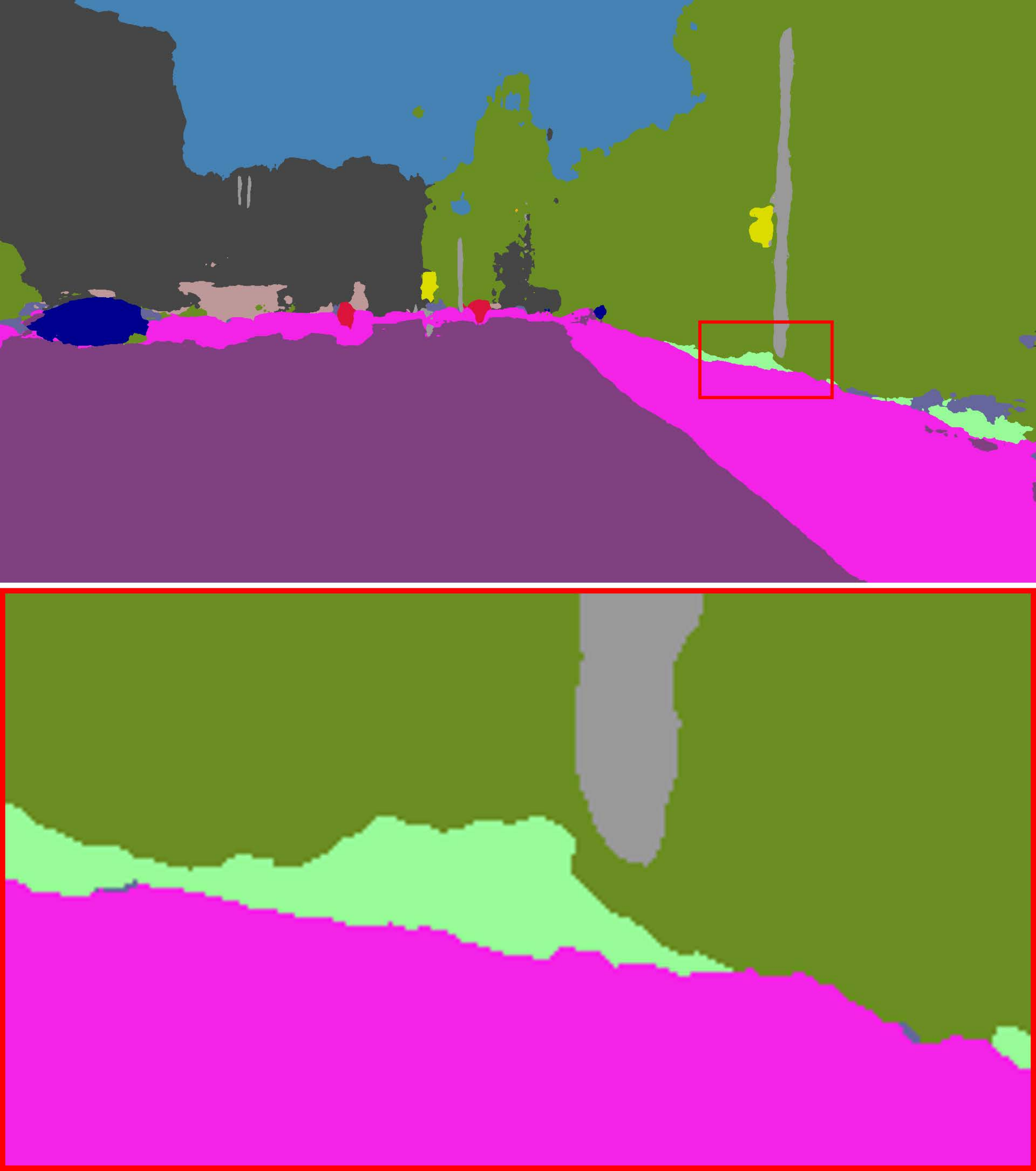}&
		\includegraphics[width=0.102\textwidth]{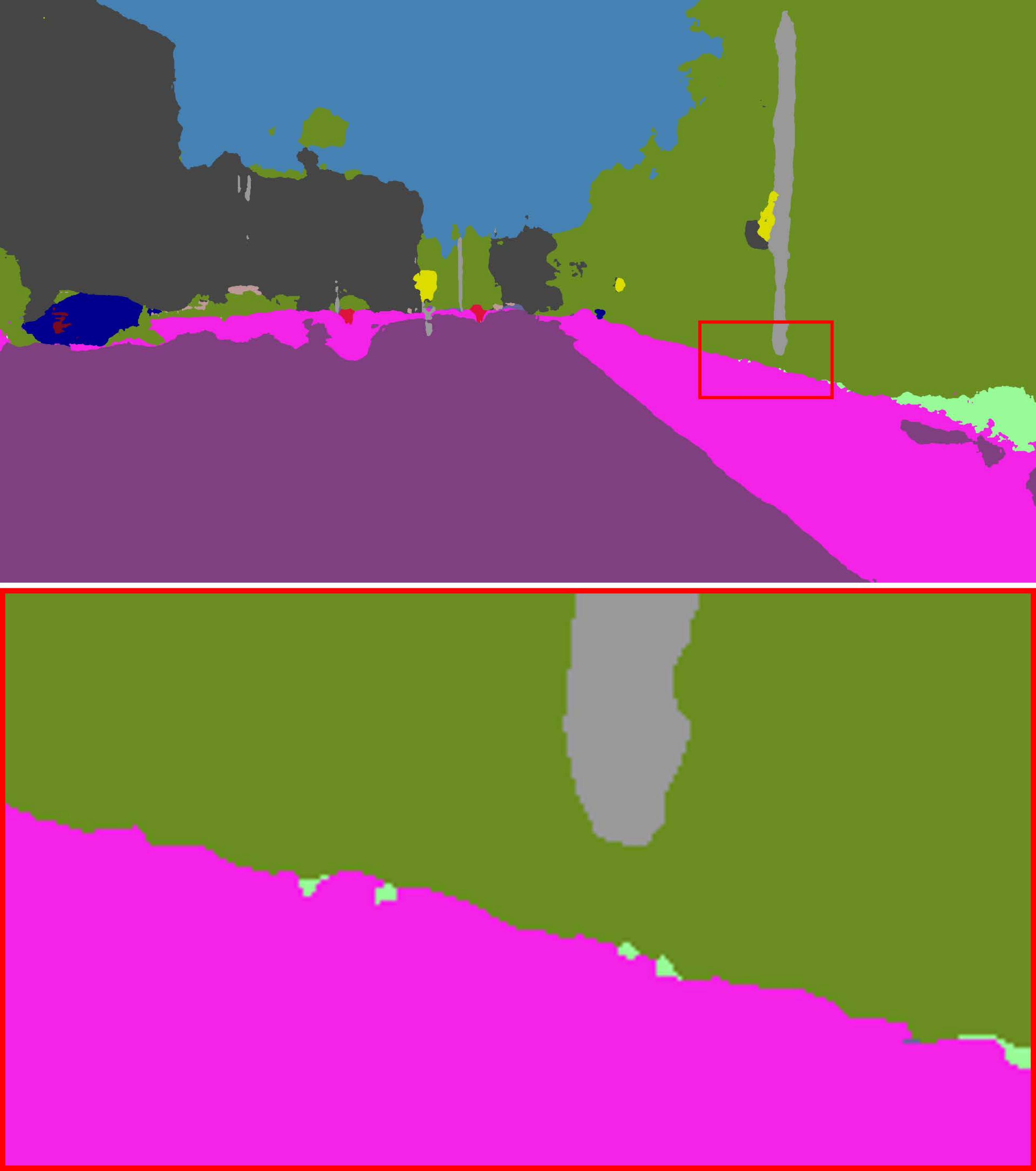}&
		\includegraphics[width=0.102\textwidth]{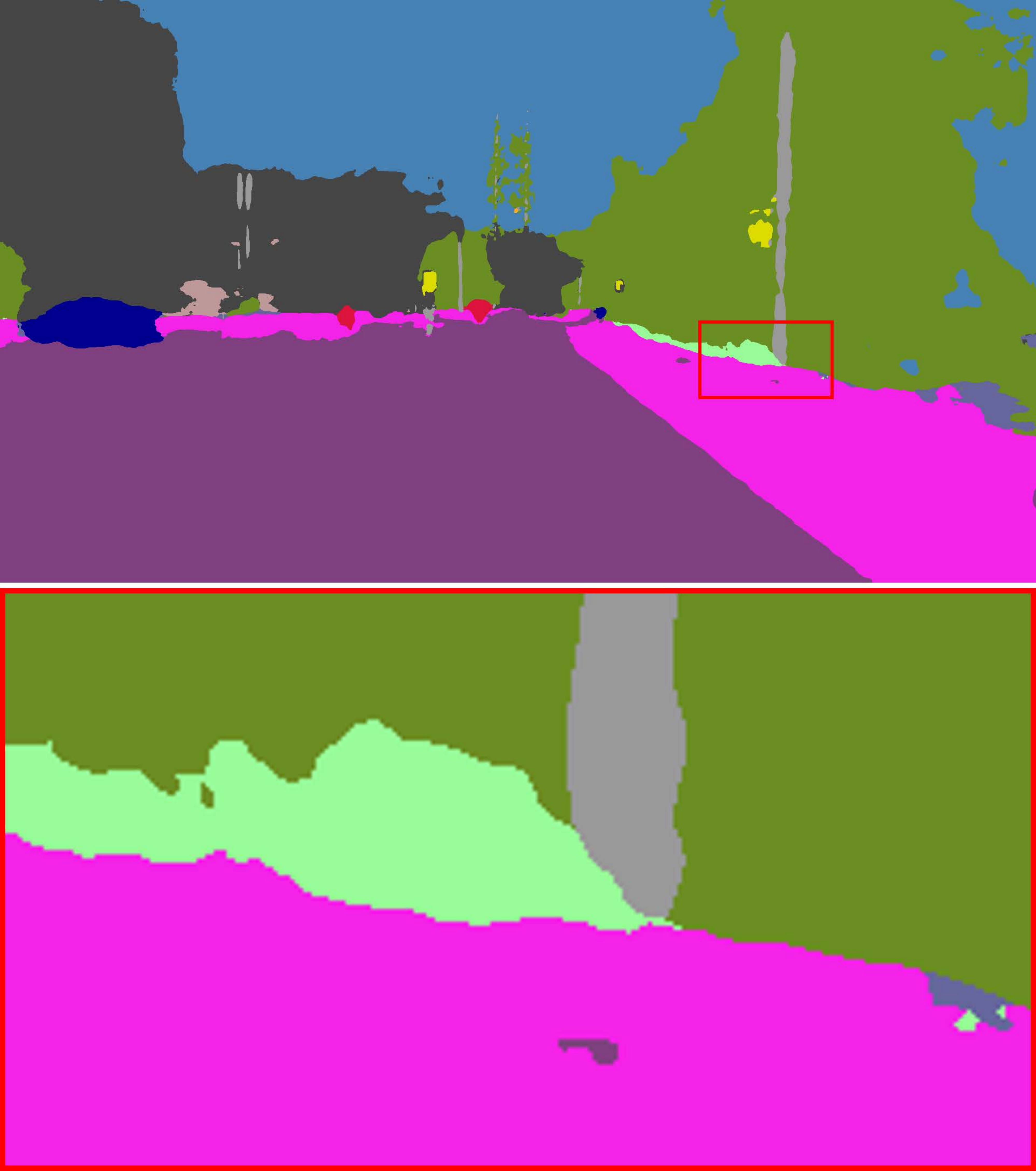}&
		\includegraphics[width=0.102\textwidth]{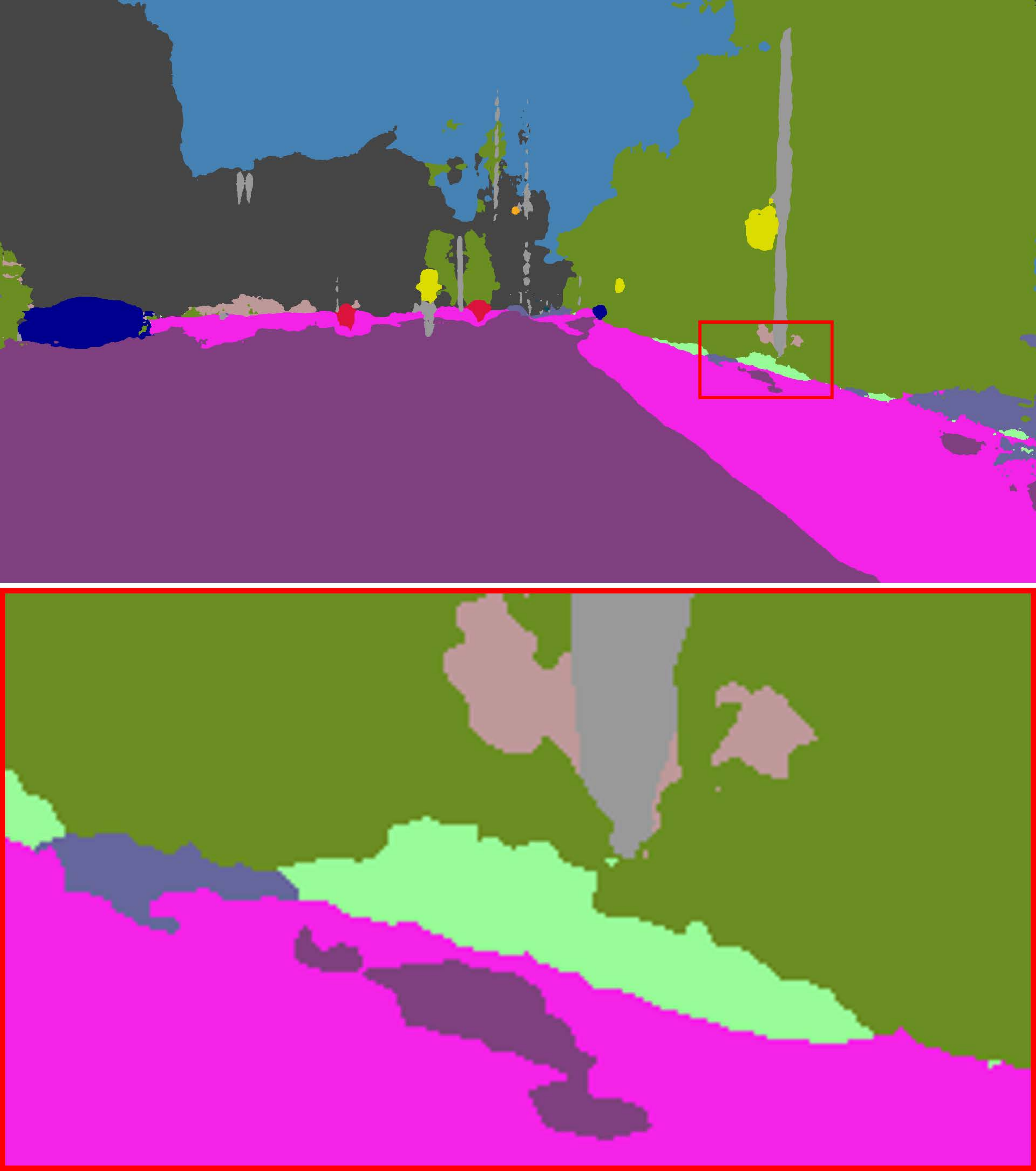}&
		\includegraphics[width=0.102\textwidth]{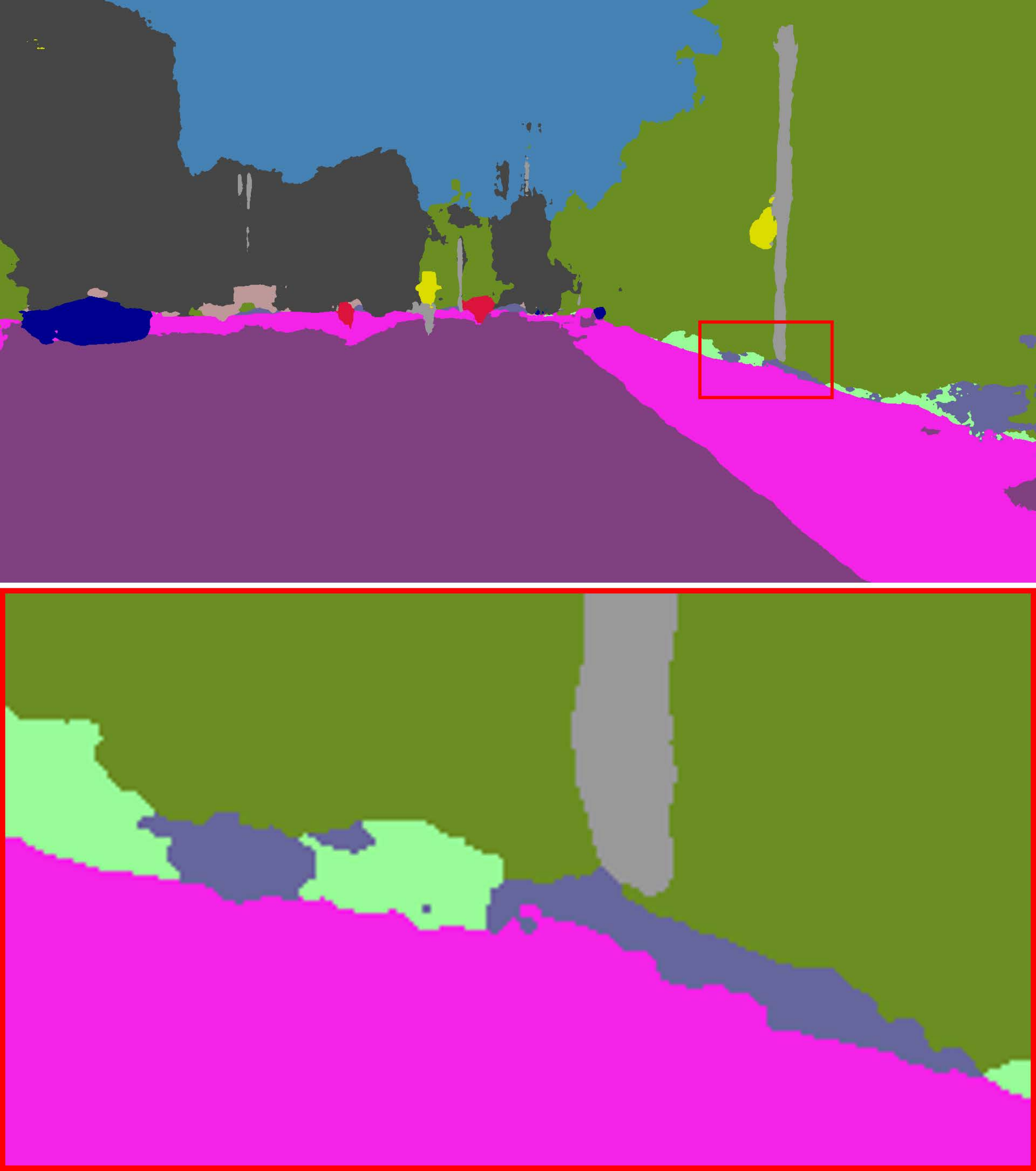}&
		\includegraphics[width=0.102\textwidth]{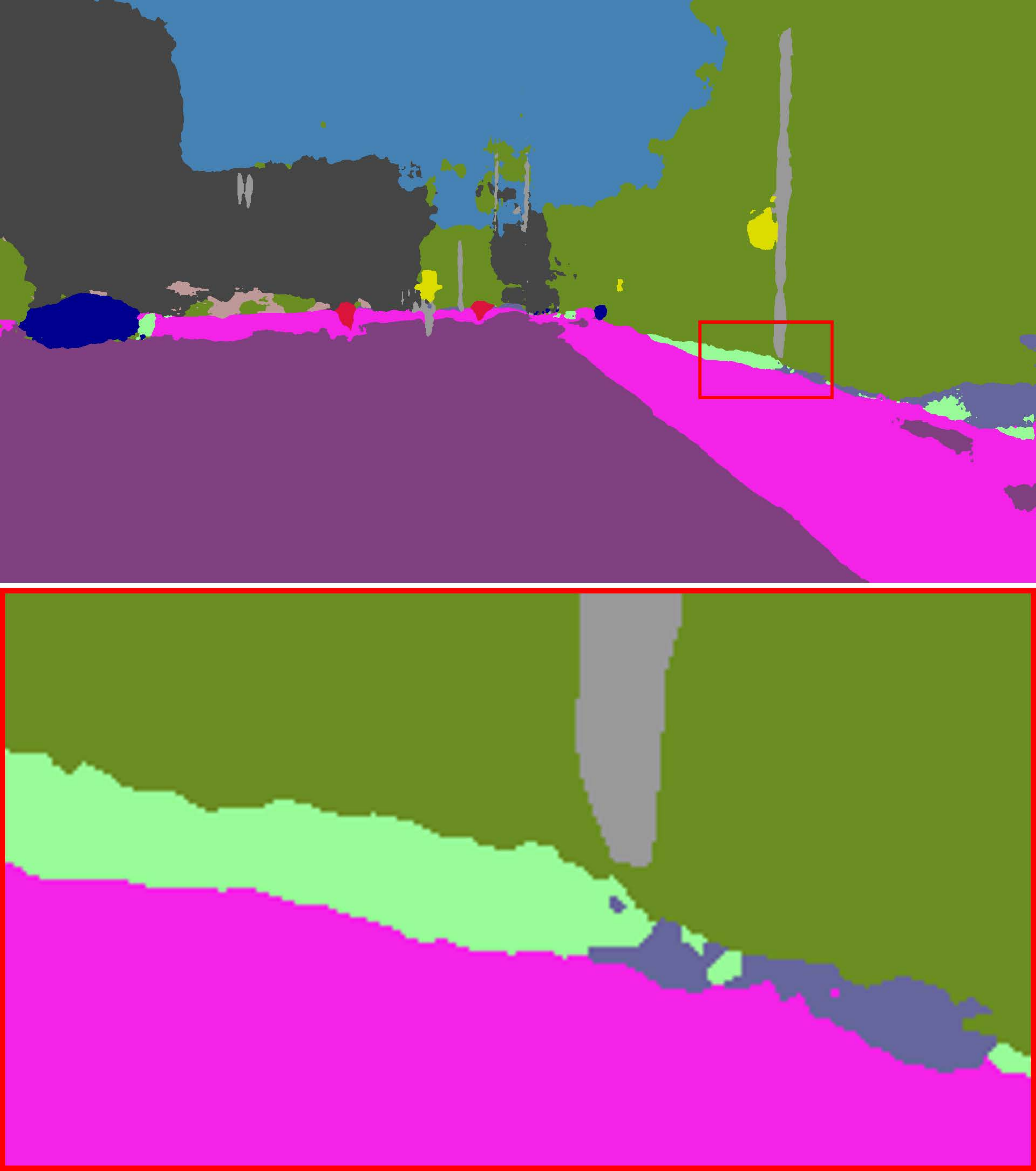}&
		\includegraphics[width=0.102\textwidth]{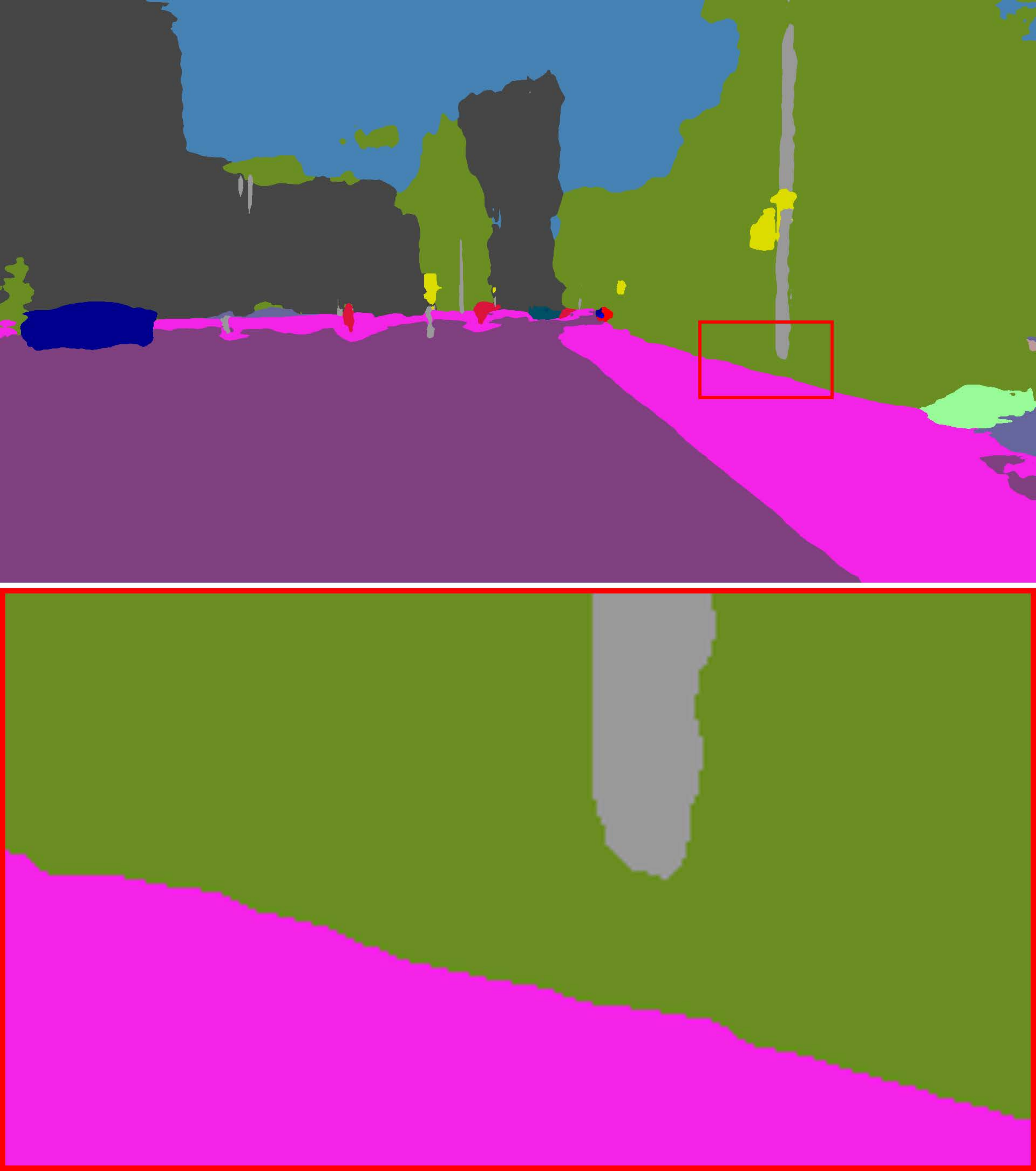}&
		\includegraphics[width=0.102\textwidth]{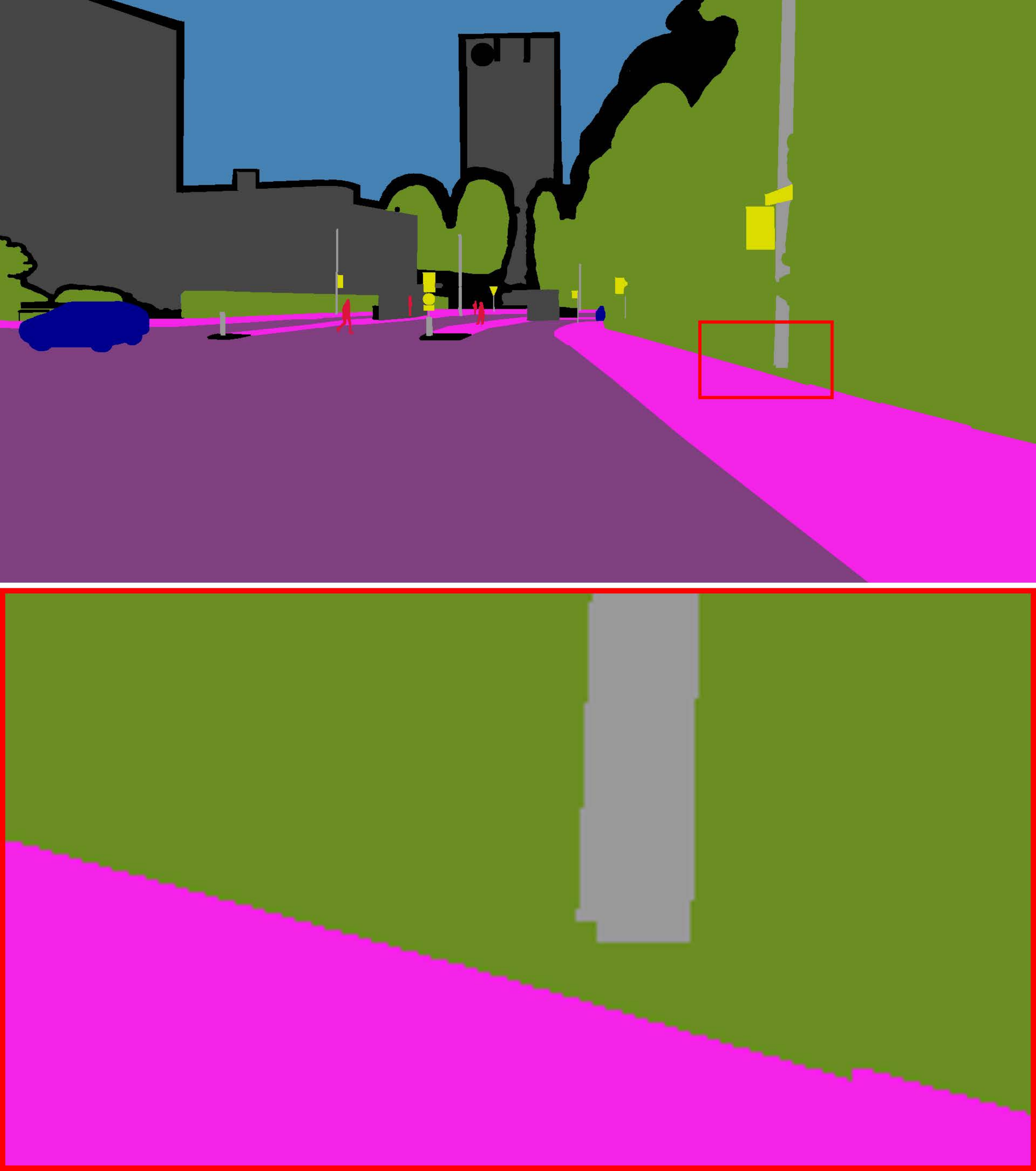}\\
		
		\includegraphics[width=0.102\textwidth]{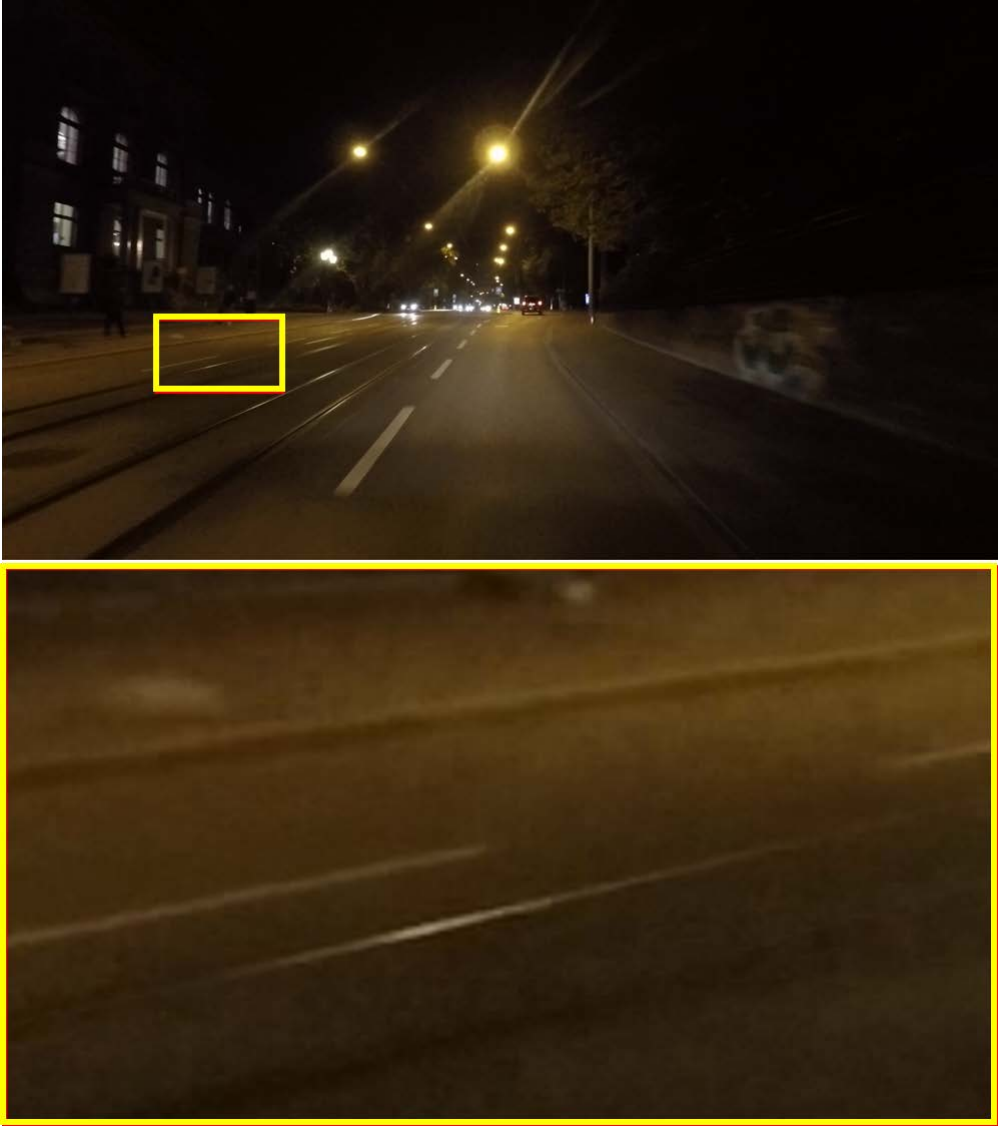}&
		\includegraphics[width=0.102\textwidth]{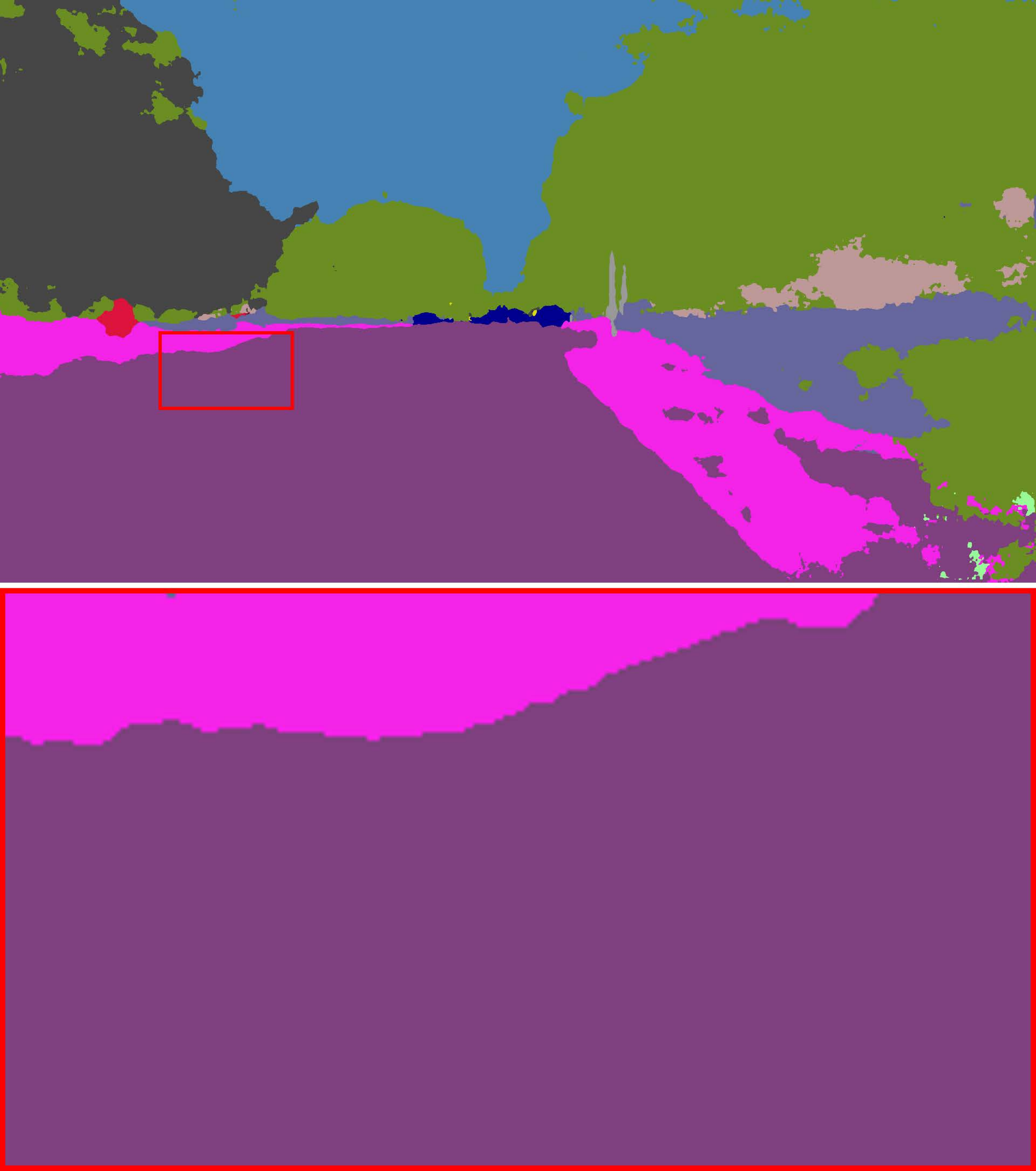}&
		\includegraphics[width=0.102\textwidth]{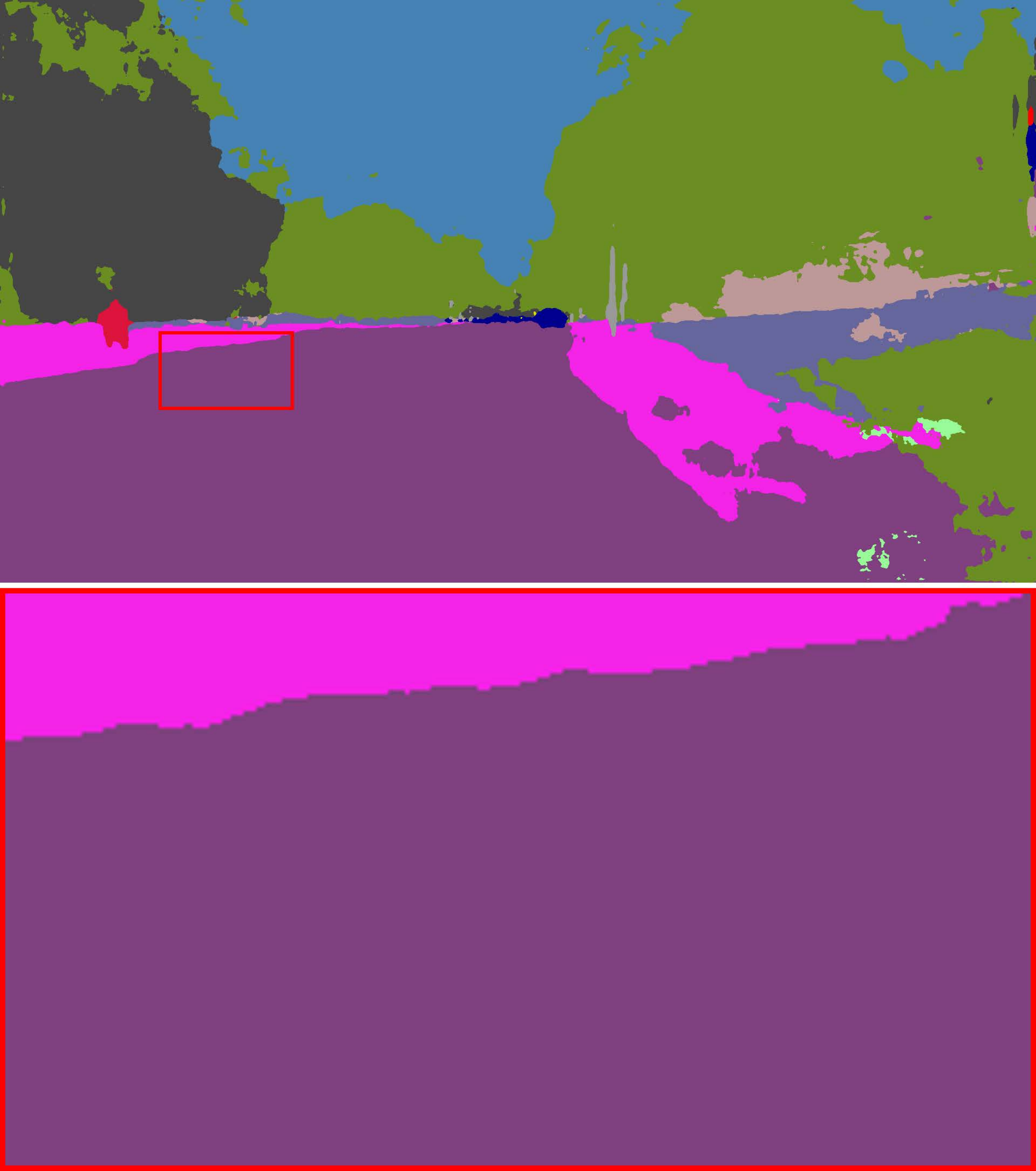}&
		\includegraphics[width=0.102\textwidth]{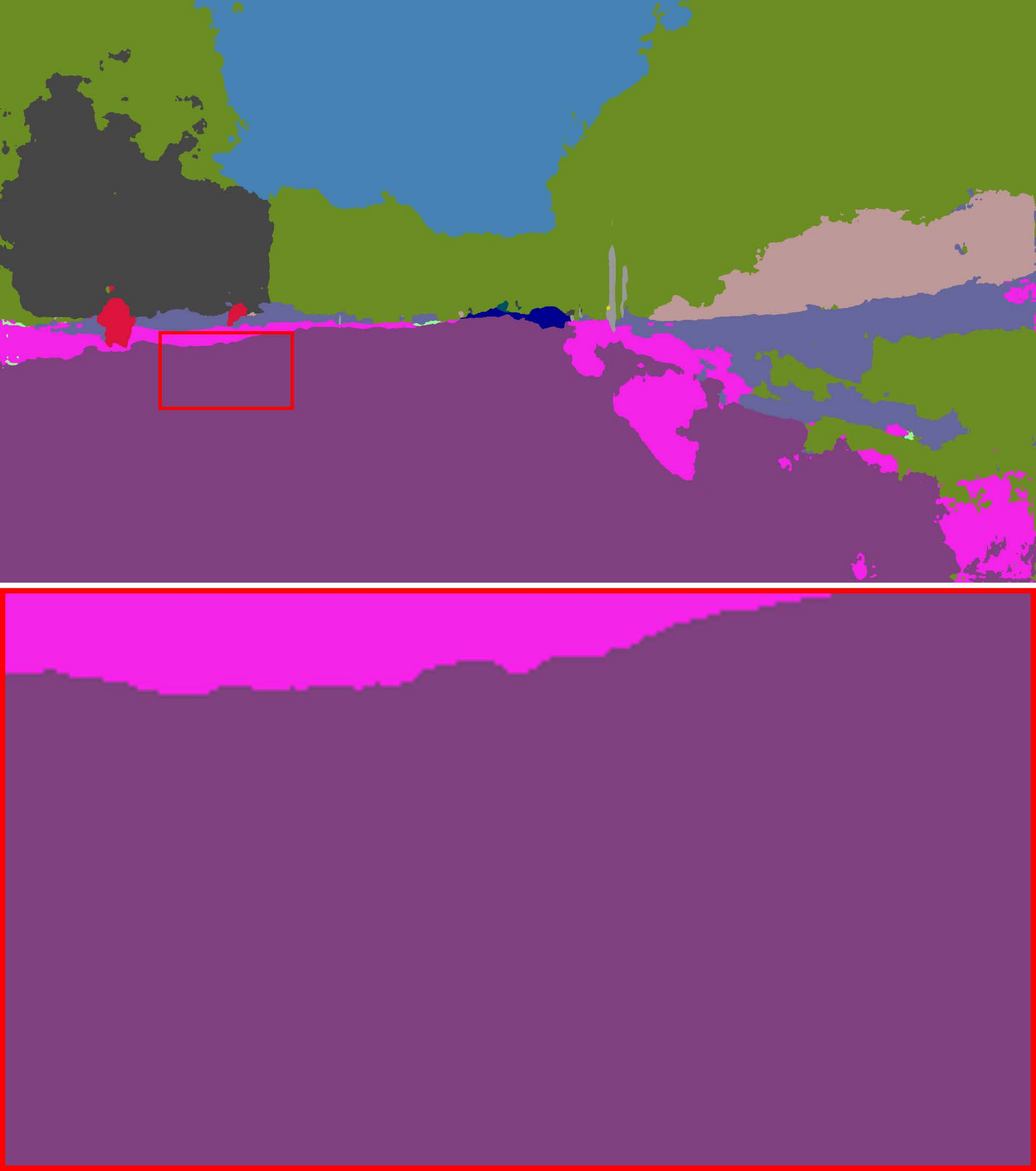}&
		\includegraphics[width=0.102\textwidth]{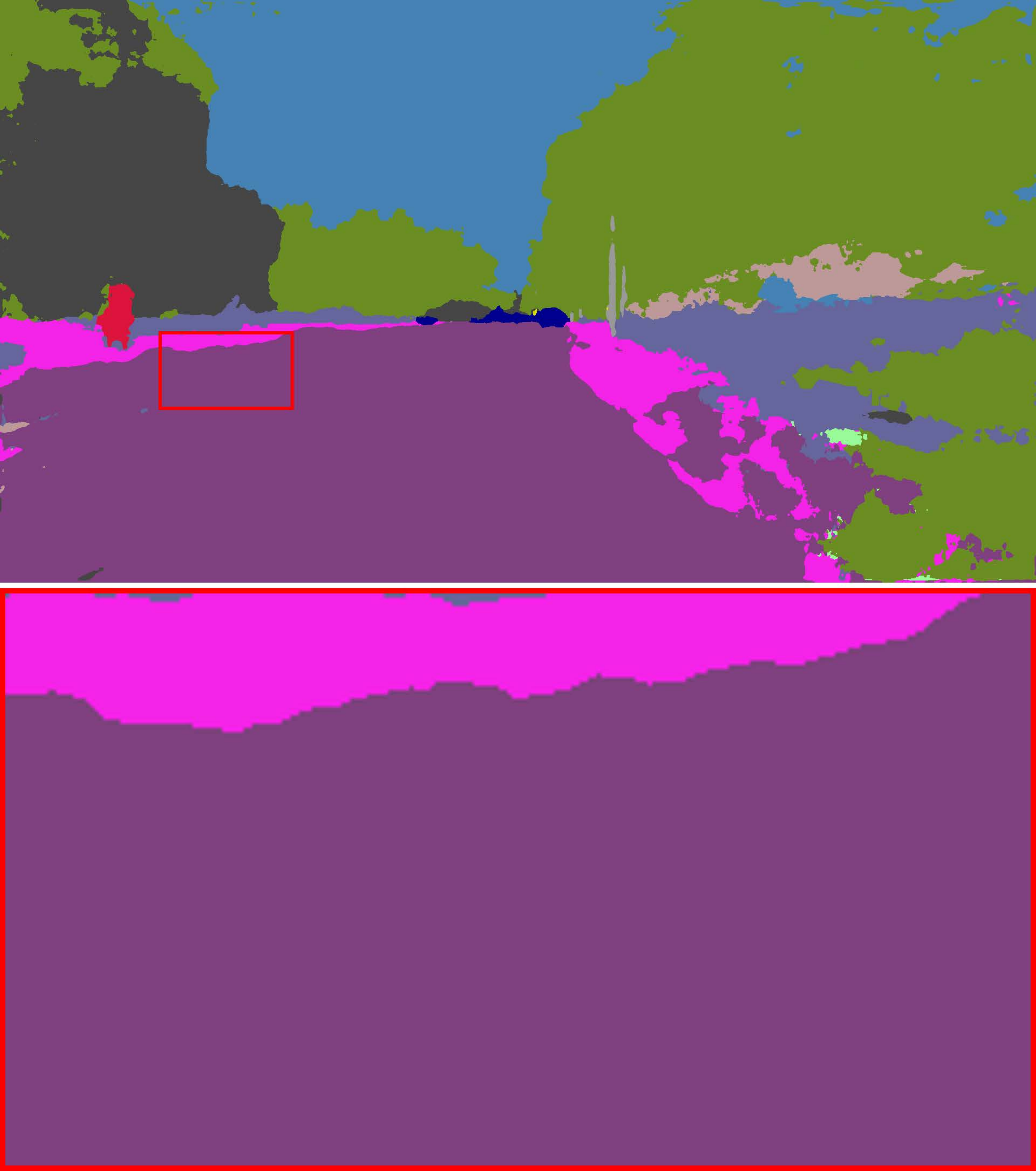}&
		\includegraphics[width=0.102\textwidth]{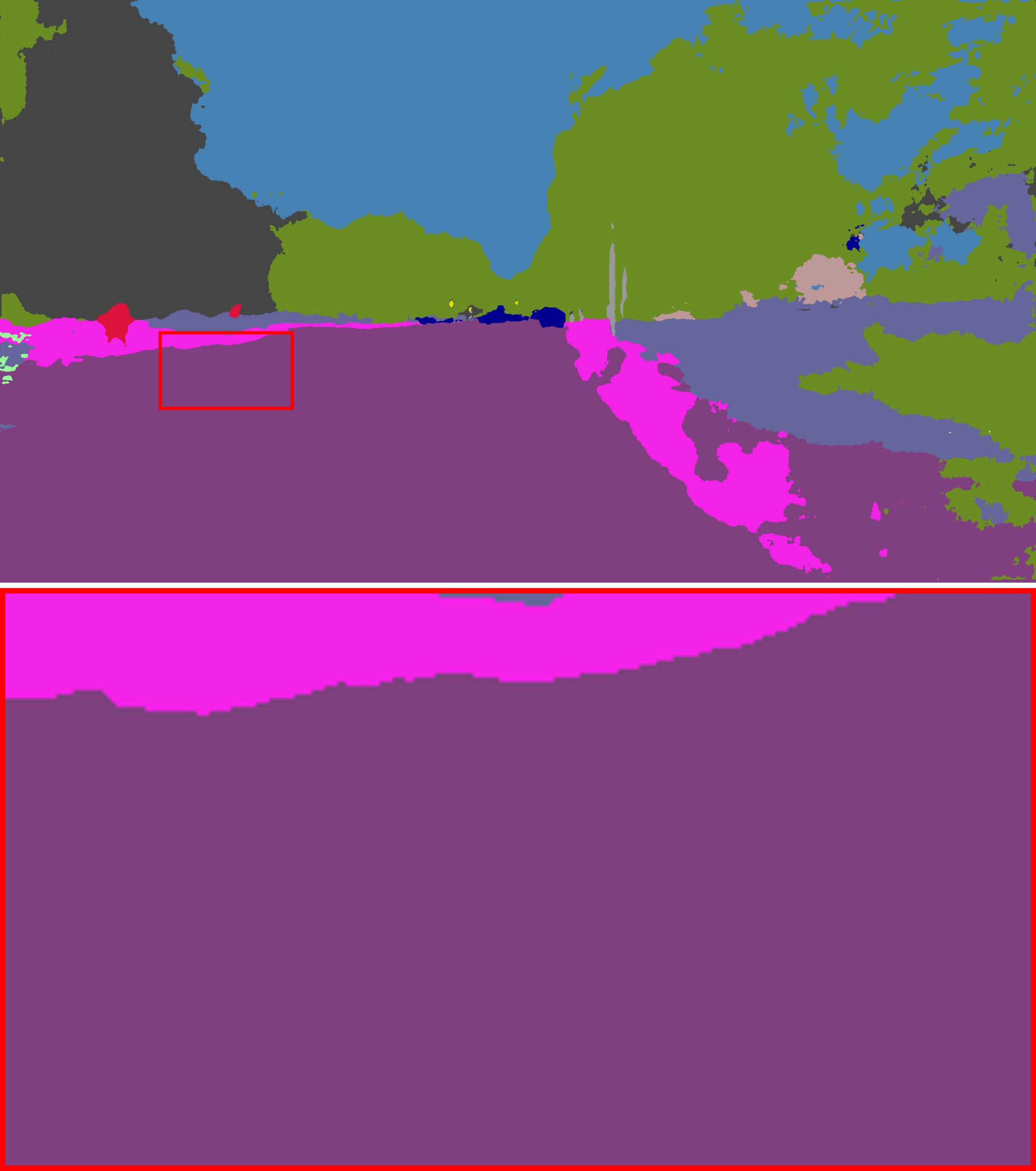}&
		\includegraphics[width=0.102\textwidth]{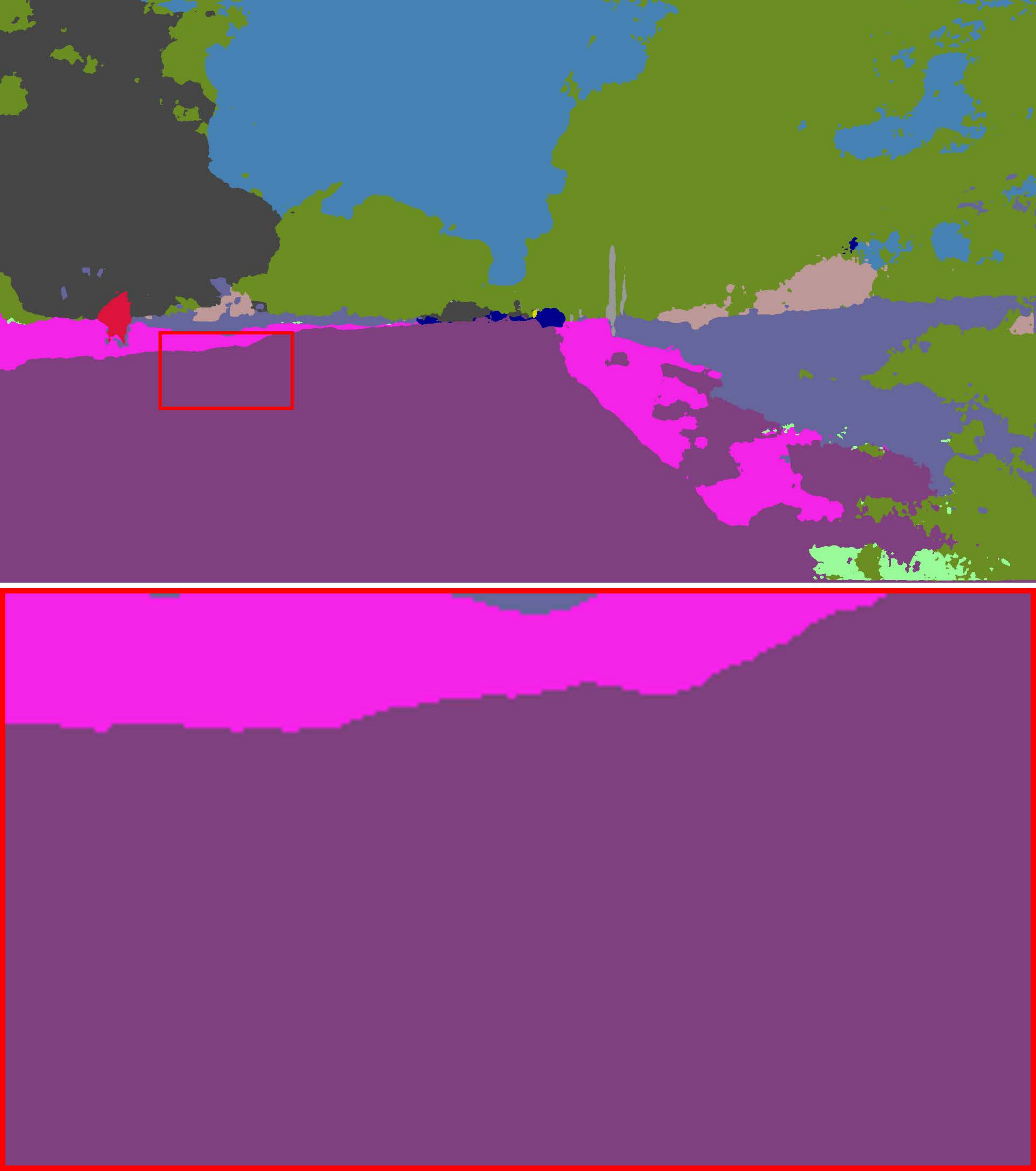}&
		\includegraphics[width=0.102\textwidth]{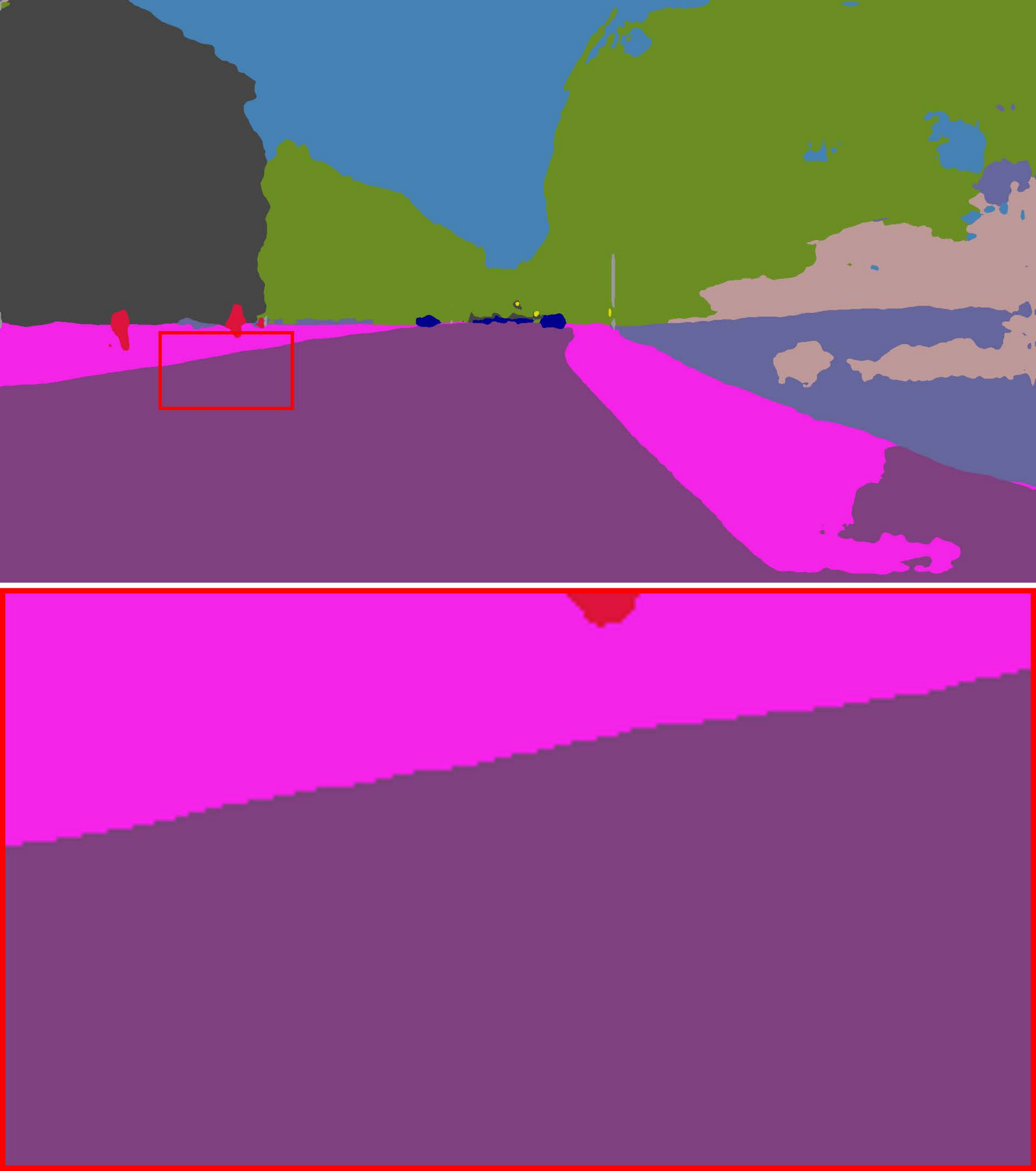}&
		\includegraphics[width=0.102\textwidth]{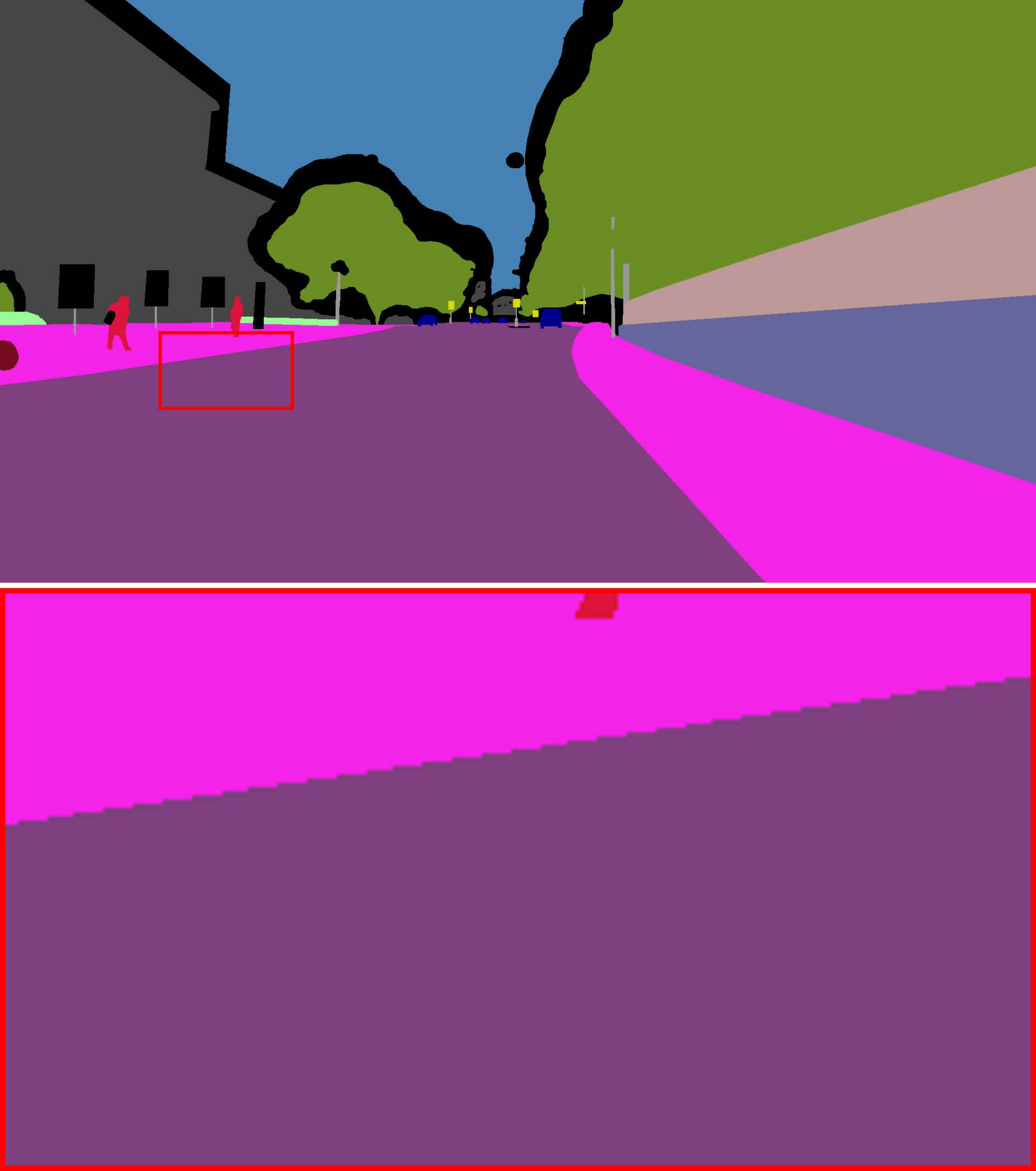}\\

		Input & DRBN & FIDE& ZeroDCE& SUC& SCI&URetinex & Ours & Ground Truth\\
	\end{tabular}
	
	\caption{Visual comparison results  against state-of-the-art low-light image enhancement approaches on the ACDC dataset. }
	\label{fig:seg}
\end{figure*}

\textit{Qualitative comparison.} We also compared our method with other enhanced methods. DeepLabV3+~\cite{chen2018encoder} serves as the baseline model for segmentation task, and the outcomes of all comparative methods are derived in the same manner with the dark face detection.  As shown in Fig.~\ref{fig:seg}, compared with other methods, our scheme can accurately segment the edges of different objects, such as roads , sidewalks, and street light signs. In the first row of images, our visuals appear more natural and accurate, thanks to the mitigation of severe artifacts and color distortion issues that exist in other results, such as SCI and URetinex.  It is obviously that our scheme can obtain more accurate segmentation results. 
\begin{figure}[thb]
	\centering
	\begin{tabular}{c@{\extracolsep{0.25em}}c@{\extracolsep{0.25em}}c} 
	
		\includegraphics[width=0.15\textwidth]{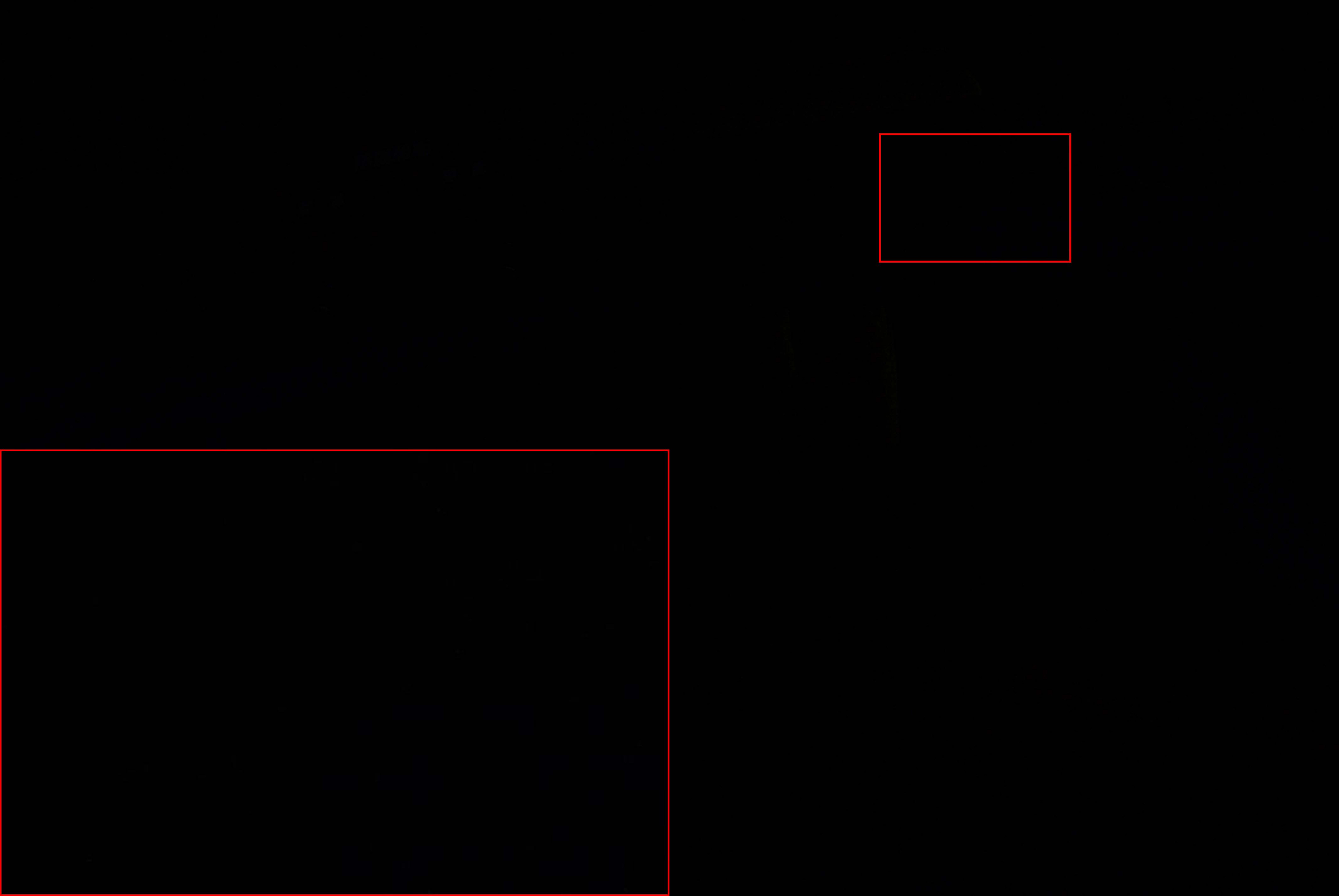}&
		\includegraphics[width=0.15\textwidth]{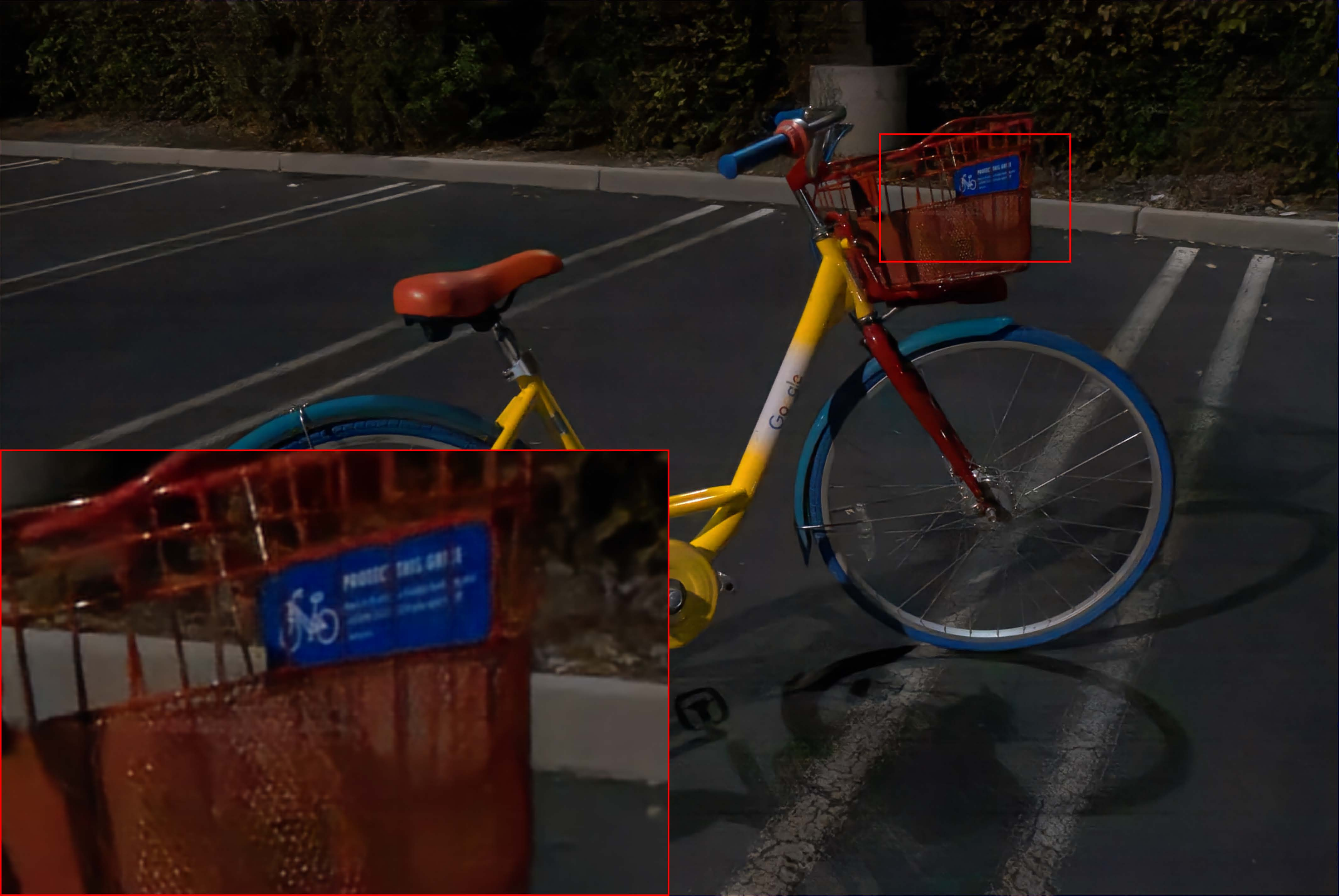}&
		\includegraphics[width=0.15\textwidth]{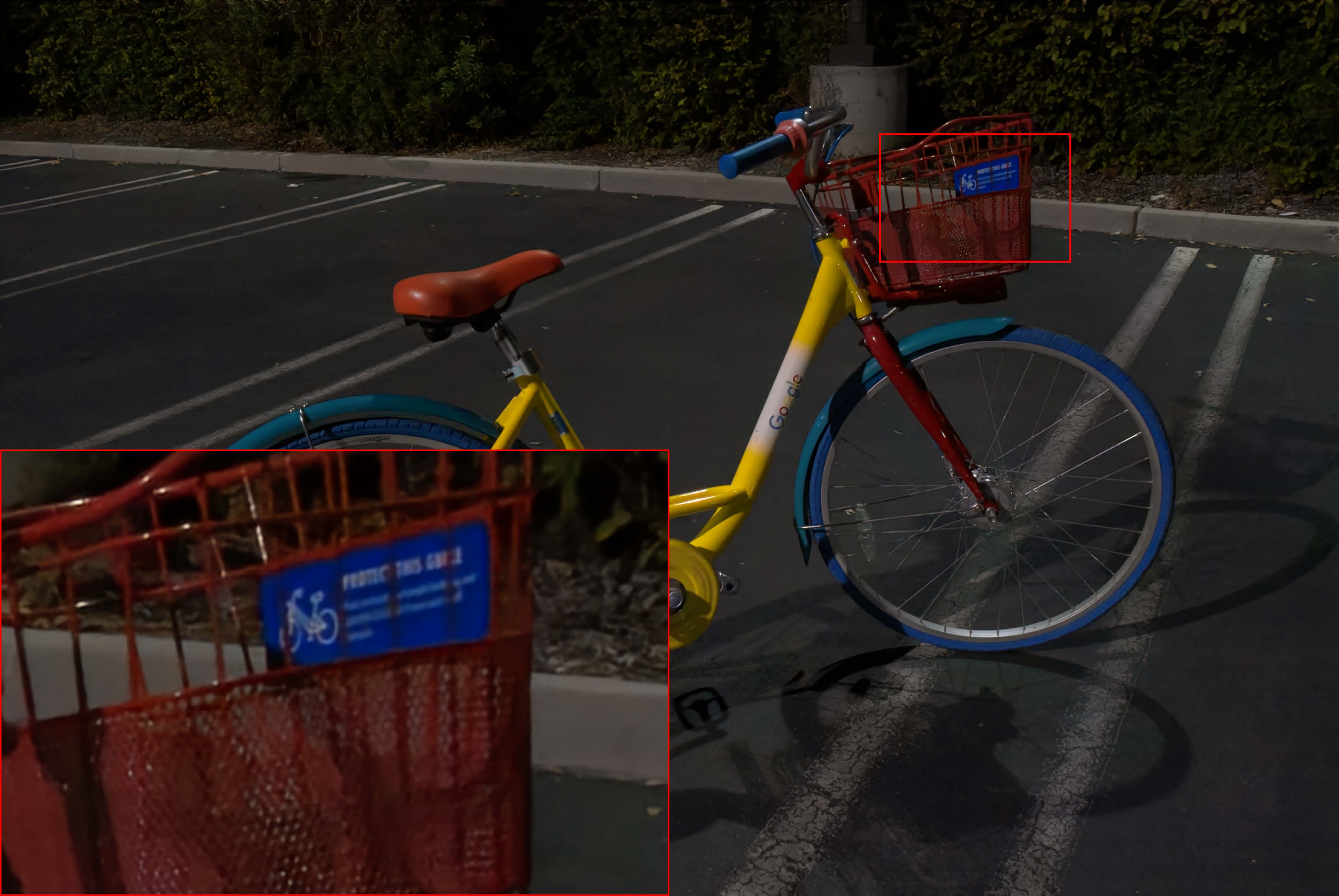}\\
		--/--&31.739/0.848&\textbf{34.344/0.900} \\	
		
		Input & TBGL& IBGL \\	
		
	\end{tabular}
	
	\caption{Visual enhancement qualitative results based on SID dataset among various training strategy.}
	\label{fig:ab_en_tb_ib}
	
\end{figure}

\textit{Quantitative comparison.} Table.~\ref{tab:ACDC_tab} demonstrates the quantitative results on the nighttime semantic segmentation task. Obviously, our results exhibits a 1.93\%  mIOU improvement over SCI method. Additionally, our segmentation results exhibit greater accuracy for specific categories, including building, wall, and fence. For example, our method achieves a 17.40\% mIOU higher score than SUC in segmenting fence, and a 0.5 mIOU higher score than URetinex in segmenting building. In a word, our results are more superior from the statistical perspective.

\begin{figure}[htb]
	\centering
	\begin{tabular}{c@{\extracolsep{0.1em}}c} 
		\includegraphics[width=0.23\textwidth]{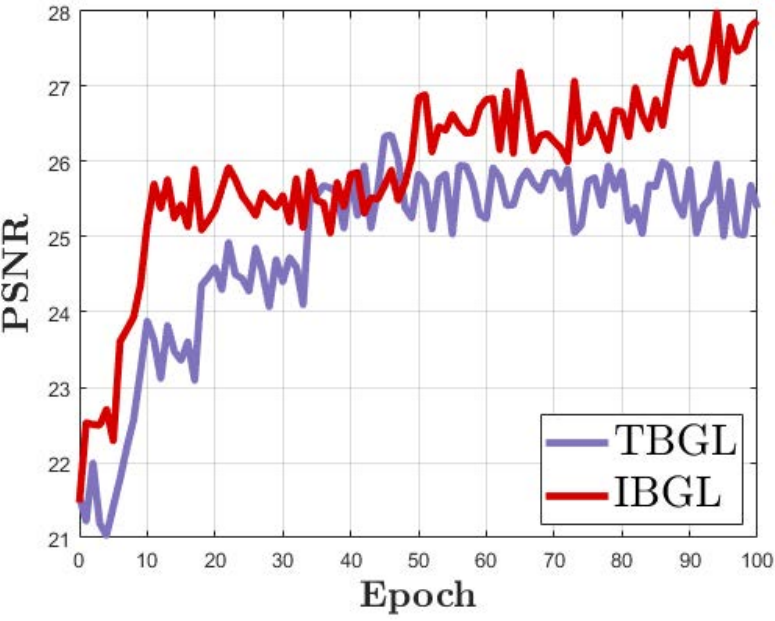}&
		
		\includegraphics[width=0.24\textwidth]{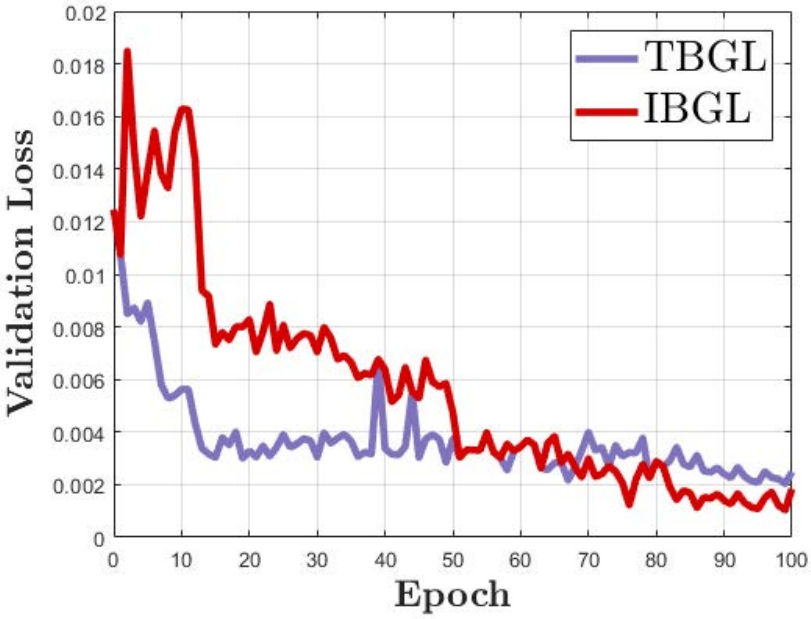}\\

	\end{tabular}
	
	\caption{Comparison results of the TBGL and IBGL by analysing the PSNR and validation loss during the training.}
	\label{fig:ablation_EN_tb_ib}
	
\end{figure}
\subsection{Framework Evaluation}

\begin{figure}[thb]
	\centering
	\begin{tabular}{c@{\extracolsep{0.25em}}c@{\extracolsep{0.25em}}c} 
		\includegraphics[width=0.15\textwidth]{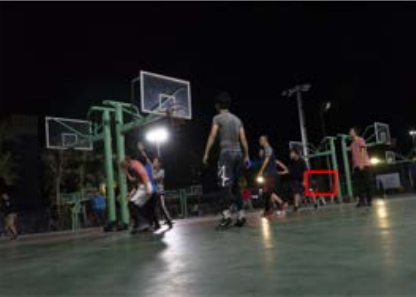}&
		\includegraphics[width=0.15\textwidth]{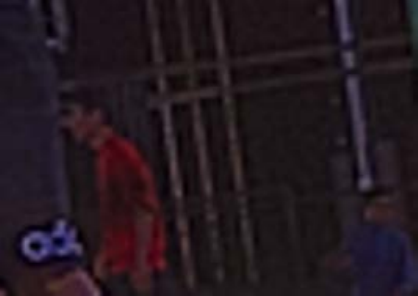}&
		\includegraphics[width=0.15\textwidth]{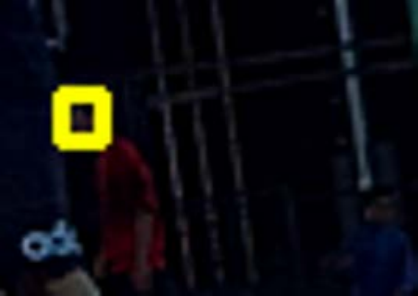}\\
		Input&M$_\mathtt{a}$& M$_\mathtt{b}$ \\	
		\includegraphics[width=0.15\textwidth]{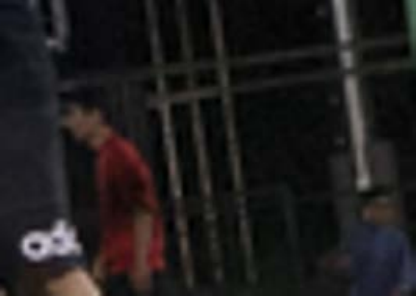}&
		\includegraphics[width=0.15\textwidth]{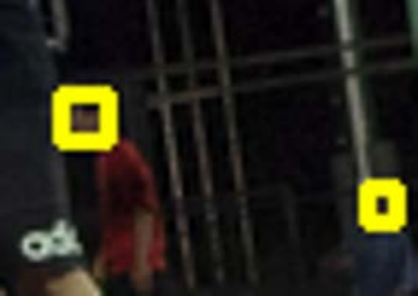}&
		\includegraphics[width=0.15\textwidth]{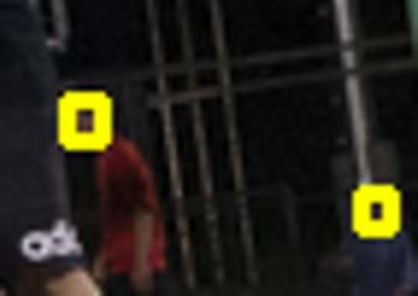}\\
		
		M$_\mathtt{c}$& M$_\mathtt{d}$& Ground Truth \\			
	\end{tabular}
	\caption{Comparative analysis of dark face detection results using various training strategies and vision tasks on the DARK FACE dataset .}
	\label{fig:cross_experiment_de}
	
\end{figure}

\begin{figure}[thb]
	\centering
	\begin{tabular}{c@{\extracolsep{0.25em}}c@{\extracolsep{0.25em}}c} 
		
		\includegraphics[width=0.15\textwidth]{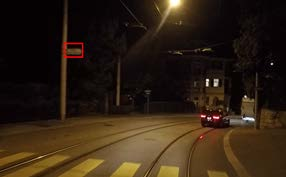}&
		\includegraphics[width=0.15\textwidth]{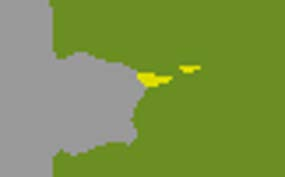}&
		\includegraphics[width=0.15\textwidth]{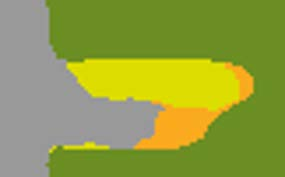}\\
		Input&M$_\mathtt{a}$& M$_\mathtt{b}$ \\	
		\includegraphics[width=0.15\textwidth]{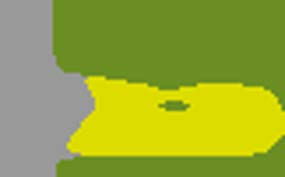}&
		\includegraphics[width=0.15\textwidth]{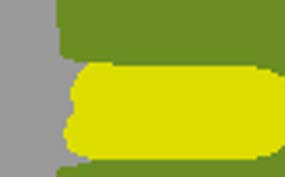}&
		\includegraphics[width=0.15\textwidth]{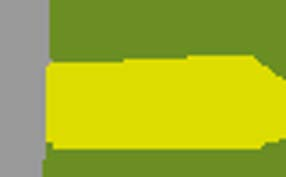}\\
		
		M$_\mathtt{c}$& M$_\mathtt{d}$& Ground Truth \\		
	\end{tabular}
	
	\caption{Comparative analysis of low-light semantic segmentation results  using various training strategies and vision tasks on the ACDC dataset.}
	\label{fig:cross_experiment_seg}
	
\end{figure}
\textit{Evaluation of the generative block.} We first analyzed the effectiveness of our proposed generative block approach (referred to as GB). For visual enhancement, as shown in first figure, the PSNR with GB improved significantly by 0.87 dB and restored more English alphabet compared with the framework without GB.  Both the numerical and visual comparisons clearly demonstrate the crucial role of the GB in providing task-friendly observations, underscoring the effectiveness of our proposed  GB.

\begin{table}[htb]
	\centering
	\caption{Verifying the generalization of GB trained on different basic tasks. ``$\mathtt{GB}$", ``$\mathtt{EN}$"
		,  ``$\mathtt{DE}$", ``$\mathtt{SE}$" represent the generative block, enhancement task, detection task and segmentation task, respectively.}
	\label{tab:cross_de_se}
	\renewcommand{\arraystretch}{1.1}
	
	\setlength{\tabcolsep}{0.15mm}{
		\begin{tabular}{|c|cc|cc|c|cc|ccc|}
			\hline
			\multirow{2}{*}{Model}&\multicolumn{2}{c|}{Basic Task} &\multicolumn{2}{c|}{Learning Strategy}& \multicolumn{2}{c|}{$\mathtt{EN}$}                                                                                                                         & \multicolumn{1}{c|}{$\mathtt{DE}$}                                                                                                                                               & \multicolumn{1}{c|}{$\mathtt{SE}$}                                                                                                                      \\ \cline{2-9} 
			&\multicolumn{1}{c|}{$\mathtt{EN}$}&\multicolumn{1}{c|}{$\mathtt{DE}$}&\multicolumn{1}{c|}{Naive}&\multicolumn{1}{c|}{IBGL}& \multicolumn{1}{c|}{NIQE $\downarrow$} & \multicolumn{1}{c|}{LOE $\downarrow$}  & \multicolumn{1}{c|}{mAP$\uparrow$ }   & \multicolumn{1}{c|}{mIOU$\uparrow$ }     
			\\ \hline
			
			\multirow{1}{*}{M$_\mathtt{a}$}	&\multicolumn{1}{c|}{\ding{55}} &\multicolumn{1}{c|}{\ding{51}}&\multicolumn{1}{c|}{\ding{51}} &\multicolumn{1}{c|}{\ding{55}}	& \multicolumn{1}{c|}{4.52}    & \multicolumn{1}{c|}{103.72}           & \multicolumn{1}{c|}{0.664}           
			&  \multicolumn{1}{c|}{42.31}                             
			\\ \hline
			
			\multirow{1}{*}{M$_\mathtt{b}$}&\multicolumn{1}{c|}{\ding{55}} &\multicolumn{1}{c|}{\ding{51 }}&\multicolumn{1}{c|}{\ding{55}}&\multicolumn{1}{c|}{\ding{51}}	& \multicolumn{1}{c|}{4.35}    & \multicolumn{1}{c|}{{94.75}}          & \multicolumn{1}{c|}{0.671}          
			& \multicolumn{1}{c|}{43.31}                           
			\\ \hline 
			
			\multirow{1}{*}{M$_\mathtt{c}$}&\multicolumn{1}{c|}{\ding{51}} &\multicolumn{1}{c|}{\ding{55}}&\multicolumn{1}{c|}{\ding{51}}&\multicolumn{1}{c|}{\ding{55}}& \multicolumn{1}{c|}{4.30}    & \multicolumn{1}{c|}{114.62}       & \multicolumn{1}{c|}{0.683}     
			&  \multicolumn{1}{c|}{{44.15}}                          
			\\ \hline 
			\multirow{1}{*}{M$_\mathtt{d}$}&\multicolumn{1}{c|}{\ding{51}} &\multicolumn{1}{c|}{\ding{55}}&\multicolumn{1}{c|}{\ding{55}}&\multicolumn{1}{c|}{\ding{51}}	& \multicolumn{1}{c|}{{\textbf{3.93}}}    & \multicolumn{1}{c|}{{\textbf{76.96}}}       & \multicolumn{1}{c|}{\textbf{0.690}}    
			&  \multicolumn{1}{c|}{\textbf{44.91} }               
			\\ \hline
	\end{tabular}}
\end{table}

\textit{Generalization of  bilevel learning.} We examine the inherent characteristics of both components of the generative block and bilevel learning. When GB are trained on various tasks, we embed them directly into other tasks while keeping their parameters frozen. It is clear that the GB trained on the enhancement task achieves excellent results on other vision tasks. As shown in  Fig.~\ref{fig:cross_experiment_de} and Fig.~\ref{fig:cross_experiment_seg}, our method outperforms M$_\mathtt{b}$ in terms of face detection task, as it can detect a greater number of faces. Additionally, our scheme exhibits higher accuracy in the segmentation task compared to M$_\mathtt{c}$. According to the Table.~\ref{tab:cross_de_se}, our method delivers a significant improvement in mAP and mIOU, boosting them by 3.6\% and 4.5\% respectively in comparison to M$_\mathtt{b}$. Our method demonstrates superior performance to the naive approach, with the  4.3\% mAP and  2.1\% mIOU increase  for both the detection and segmentation tasks. Through our analysis, we have discovered that training on enhancement task can facilitate the learning of scene information that is transferable to a range of vision tasks. Our bilevel training strategy is particularly effective in highlighting this generalization.

\textit{Analysis of the training strategies.} 
In this section, we evaluated our proposed demand-oriented learning strategy with two variants, TBGL and IBGL. Fig.~\ref{fig:ab_en_tb_ib} demonstates that IBGL outperforms the  TBGL methods in restoring texture details. For example, the text of IBGL is clearer, while the effect of  TBGL is more blurred in the zoomed-in area in the first row. From the numerical results as shown in Table.~\ref{tab:ablation_Enhancement_tb_ib}, IBGL achieved an improvement of almost 0.639 dB in PSNR than TBGL.  The last row in the table shows that  IBGL demands more accurate computations and utilizes a greater amount of memory space.  Fig.~\ref{fig:ablation_EN_tb_ib} depicts the different training trends for PSNR and validation loss after approximately 500 epochs during the warm start phase. IBGL is more convergent than TBGL, it demonstrated better accuracy.

\begin{table}[htb]
	\centering
	\renewcommand{\arraystretch}{1.1}
	\caption{Comparison of quantitative results of IBGL and TBGL on the SID dataset. Three metrics (\textit{i.e.}, PSNR, SSIM, LPIPS ) are reported, Mb for GPU Usage.}
	\setlength{\tabcolsep}{3.1mm}{
		\begin{tabular}{|c| c | c| c| c|}
			\hline
			Strategy & PSNR$\uparrow$  &SSIM$\uparrow$&LPIPS$\downarrow$&GPU Usage$\downarrow$   \\\hline

			TBGL&28.483&0.753&0.667&\textbf{5781}\\\hline
			
			IBGL&\textbf{29.122}&\textbf{0.769}&0.432&12605\\
			\hline
		\end{tabular}	
	}
	\label{tab:ablation_Enhancement_tb_ib}
\end{table}

\section{CONCLUSION}
In this paper, we have proposed a unified framework to address diverse low-light vision tasks. By introducing GB from the RAW domain space, we have constructed a common low-light observation that connects different vision tasks. The inner relationship of them is seldom considered. Through imposing learnable data generation, we have formulated a bilevel generative modeling approach. Additionally, we have proposed two practical strategies to effectively address real-world demands. The comprehensive ability of our framework has been demonstrated through two representative low-light vision tasks, namely detction and segmentation, achieving state-of-the-art results. 

\section*{ACKNOWLEDGEMENTS}
This work is partially supported by the National Key R\&D Program of China (No. 2022YFA1004101), the National Natural Science Foundation of China (No. U22B2052).

\bibliographystyle{ACM-Reference-Format}
\bibliography{sample-base}

\appendix

\end{document}